\documentclass[smallextended,twocolumn]{svjour3}
\smartqed

\usepackage[utf8]{inputenc}
\usepackage[T1]{fontenc}
\usepackage{newtxtext}
\usepackage{newtxmath}
\usepackage{graphicx}
\usepackage{microtype}
\usepackage{booktabs}
\usepackage{multirow}
\usepackage{makecell}
\usepackage{xcolor}     
\usepackage{siunitx}
\usepackage{subcaption}
\usepackage{enumitem}
\usepackage{mathtools}  
\usepackage[numbers,sort&compress]{natbib}
\usepackage[colorlinks,citecolor=blue,urlcolor=blue,linkcolor=blue]{hyperref}

\usepackage{xspace}
\newcommand{\xwhu}[1]{\textcolor{black}{#1}} 
\newcommand{\ie}{\emph{i.e.}\xspace}

\journalname{International Journal of Computer Vision}
\usepackage{fancyhdr}
\begin{document}
	
	\title{Unveiling Deep Shadows: A Survey and Benchmark on Image and Video Shadow Detection, Removal, and Generation in the Deep Learning Era}
	
\author{
	Xiaowei Hu$^{1}$ \and 
	Zhenghao Xing$^{2}$ \and 
	Tianyu Wang$^{3}$ \and 
	Chi-Wing Fu$^{2}$ \and 
	Pheng-Ann Heng$^{2}$
}

\institute{
	$^{1}$School of Future Technology, South China University of Technology, Guangzhou, China\\
	$^{2}$Department of Computer Science and Engineering, The Chinese University of Hong Kong, Hong Kong SAR, China\\
	$^{3}$Adobe Research, San Francisco, CA, USA\\
	Corresponding author: X. Hu (huxiaowei@scut.edu.cn).\\
}

	\date{Received: date / Accepted: date}
	
	\maketitle
	
	\begin{abstract}

	Shadows, formed by the occlusion of light, play an essential role in visual perception and directly influence scene understanding, image quality, and visual realism. 
	This paper presents a unified survey and benchmark of deep-learning-based shadow detection, removal, and generation across images and videos. 
	We introduce consistent taxonomies for architectures, supervision strategies, and learning paradigms; review major datasets and evaluation protocols; and re-train representative methods under standardized settings to enable fair comparison.
	Our benchmark reveals key findings, including inconsistencies in prior reports, strong dependence on model design and resolution, and limited cross-dataset generalization due to dataset bias. 
	By synthesizing insights across the three tasks, we highlight shared illumination cues and priors that connect detection, removal, and generation. 
	We further outline future directions involving unified all-in-one frameworks, semantics- and geometry-aware reasoning, shadow-based AIGC authenticity analysis, and the integration of physics-guided priors into multimodal foundation models. 
	Corrected datasets, trained models, and evaluation tools are released to support reproducible research.
	\end{abstract}
	
	\section{Introduction} \label{sec:intro}

Shadows arise when light is partially or fully occluded by objects, producing regions of reduced illumination whose appearance reflects the underlying light intensity, scene geometry, and object-surface relationships. 
In computer vision and multimedia processing, shadow analysis is essential for both understanding and manipulating visual content. 
Shadow detection provides cues about illumination direction, scene structure, and hidden light-object interactions; shadow removal improves the fidelity of downstream vision tasks and is widely used in photography, image enhancement, and visual communication; and shadow generation supports realistic rendering and plays a central role in virtual content creation, including graphics, AR/VR, and image/video editing systems.

With the advent of deep learning, the performance of shadow detection, removal, and generation has progressed rapidly. 
However, the proliferation of models, datasets, and task formulations makes it increasingly difficult to understand and compare the underlying principles of state-of-the-art approaches. 
Despite this growth, the past decade has lacked a unified survey that jointly examines deep-learning-based shadow detection, removal, and generation across both images and videos, motivating the need for a systematic consolidation of this field.

The earliest survey~\cite{woo1990survey} reviews the types of shadows and shadow generation algorithms in computer graphics.
\cite{prati2001analysis,prati2001comparative} review shadow detection methods in videos, encompassing deterministic model and non-model-based, and statistical parametric and nonparametric methods.
Surveys on shadow detection and removal in the 2010s are summarized as follows~\cite{al2012survey,sanin2012shadow,shahtahmassebi2013review,sasi2015shadow,mahajan2015survey,tiwari2016survey,mostafa2017review,murali2018survey}.
\cite{al2012survey} surveys shadow detection methods, organized based on object/environment dependency and implementation domain.
\cite{sanin2012shadow} reviews shadow detection methods in a feature-based taxonomy.
\cite{shahtahmassebi2013review,mostafa2017review} review shadow detection and removal methods in remote sensing.
\cite{sasi2015shadow} reviews image-based shadow detection and removal methods in real images.
\cite{mahajan2015survey} reviews shadow detection methods using difference index and succeeding thresholding.
\cite{tiwari2016survey} analyzes the performance of shadow detection techniques for images and videos in various scenarios, including indoor and outdoor scenes, fixed or moving cameras, and detection of umbra and penumbra shadows.
\cite{murali2018survey} categorizes shadow detection methods into five categories: invariant-based detection, feature-based detection, region-based detection, color model-based detection, and interactive shadow detection.

Recently, deep learning methods have been reviewed for shadow detection in remote sensing~\cite{dong2024review} and satellite images~\cite{lei2023shadow}. 
\cite{zhu2024darkness} reviews image shadow removal methods from 2017 to 2024 but does not compare these methods under a consistent experimental environment.
The concurrent work~\cite{guo2024single} surveys deep models for image shadow removal but neglects methods for video, facial, and document shadow removal, and other shadow-related tasks. Additionally, it does not incorporate the latest datasets, shadow masks, and evaluation metrics that scale up data samples or address errors from previous works. Crucially, the study fails to re-train deep models under unified settings for experimental comparisons.

To date, there is no comprehensive survey and benchmark covering deep-learning-based shadow detection, removal, and generation across both images and videos. 
Addressing this gap, our paper presents a unified and in-depth examination of modern shadow analysis. We provide a taxonomy-driven review of methods and datasets, conduct benchmark experiments under standardized settings for fair comparison, report cross-dataset evaluations to reveal model generalization behavior, analyze size-speed-accuracy trade-offs, and synthesize how detection, removal, and generation interact through shared illumination cues and priors.
The survey concludes with a discussion of emerging trends in AIGC and large multimodal models and outlines future research opportunities.

In this paper, our contributions are summarized as follows:

\begin{itemize}[leftmargin=*]
	
	\item \textbf{A Comprehensive Survey of Deep Shadow Analysis.}
	We provide the first unified survey that concurrently examines deep-learning-based shadow detection, removal, and generation in both images and videos.
	The paper presents a structured taxonomy that synthesizes task formulations, supervision strategies, architectural paradigms, and the semantic and geometric cues exploited by modern methods.
	
	\vspace{2mm}
	
	\item \textbf{Fair and Reproducible Benchmarking under Standardized Settings.}
	We unify input resolution, training configuration, and evaluation metrics, and we retrain representative methods on refined datasets with corrected annotations.
	This facilitates fair comparison across approaches and reveals performance behaviors that are not apparent in previously reported results.
	
	\vspace{2mm}
	
	\item \textbf{Analysis of the Relationship among Model Size, Inference Speed, and Accuracy.}
	We conduct a systematic investigation of computational efficiency and accuracy, providing practical insights into the trade-offs among model scale, runtime, and performance.
	
	\vspace{2mm}
	
	\item \textbf{Cross-Dataset Generalization Study.}
	To assess robustness beyond dataset-specific biases, we evaluate representative models across multiple datasets and analyze their transferability, highlighting generalization limitations and potential remedies.
	
	\vspace{2mm}
	
	\item \textbf{Synthesized Trends, Insights, and Open Challenges.}
	We distill common trends across detection, removal, and generation, identify current obstacles in semantic and geometric reasoning, illumination modeling, and hybrid supervision, and articulate research gaps that remain unaddressed by existing studies.
	
	\vspace{2mm}
	
	\item \textbf{Future Directions Enabled by AIGC and Large Multimodal Models.}
	We discuss emerging opportunities in AIGC authenticity assessment, multimodal large-model reasoning, and real-world applications in 3D reconstruction, robotics, and visual content creation.
	
	\vspace{2mm}
	
	\item \textbf{Public Release of Models, Results, and Protocols.}
	We provide trained models, evaluation results, and standardized protocols at Github\footnote{\url{https://github.com/xw-hu/Unveiling-Deep-Shadows}} to support transparent and reproducible research.
	
\end{itemize}
\vspace{2mm}

Together, these contributions provide a comprehensive survey and a fair evaluation benchmark, setting it apart from earlier review papers. The subsequent sections of the paper are organized as follows. Sections \ref{sec:shadow_physics}\&\ref{sec:1.2} show the shadow physics and history. Sections \ref{sec:detection}-\ref{sec:generation} present a comprehensive survey on shadow detection, instance shadow detection, shadow removal, and shadow generation, respectively. 
Each section contains the introductions of deep models, datasets, evaluation metrics, and experimental results.  
Sections~\ref{sec:discussion_unified}\&\ref{sec:future} delve into the discussion on shadow analysis and highlight open issues and research challenges in the field. Finally, Section~\ref{sec:conclusion} presents the conclusions of the paper.

\begin{figure*}[tp]
	\centering
	\includegraphics[width=\linewidth]{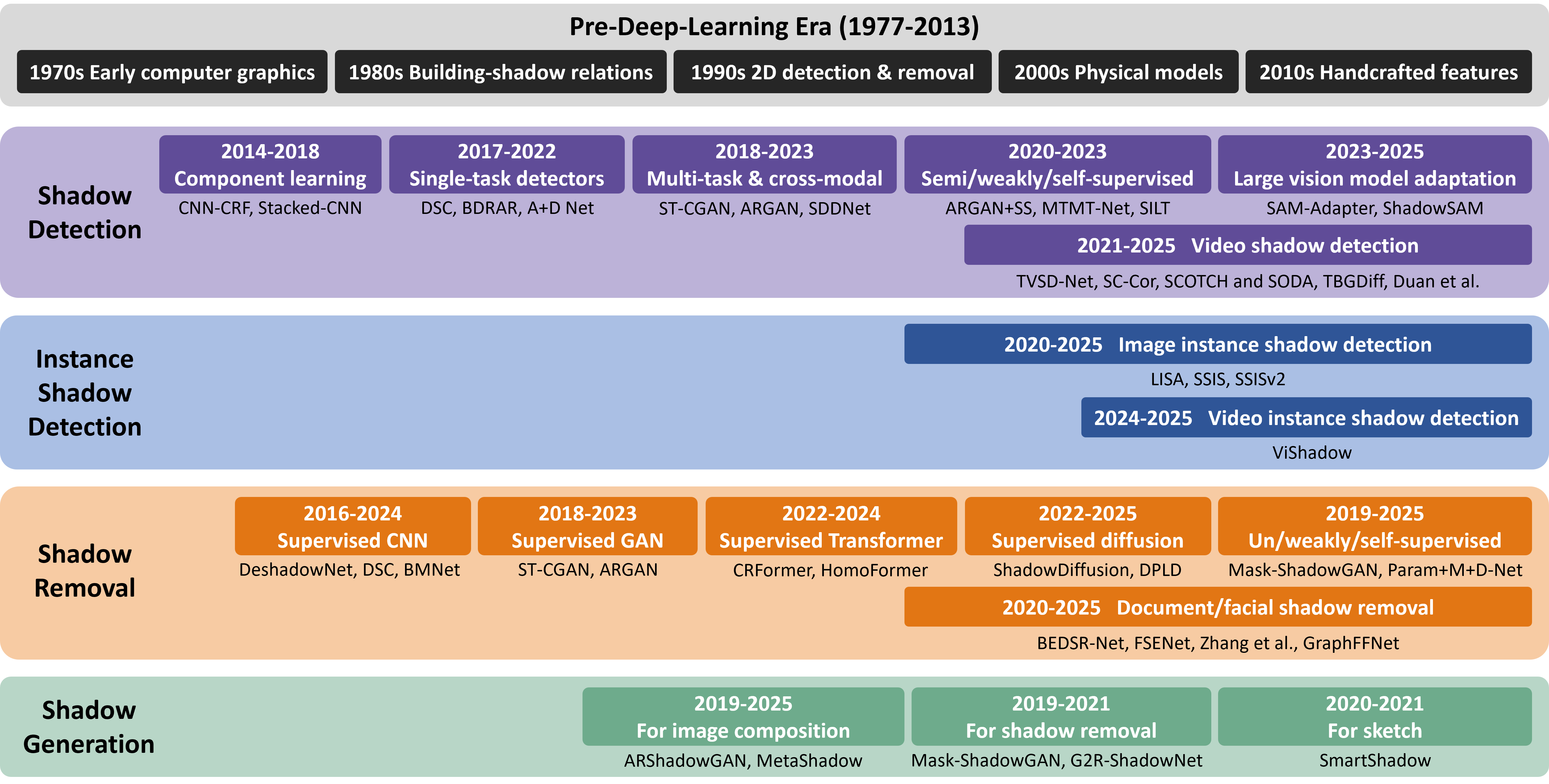}
	\caption{Chronological overview of the evolution of image and video shadow analysis from the pre-deep-learning era (1977-2013) to the deep learning era (2014-2025). The diagram highlights representative milestones across four task families: shadow detection, instance shadow detection, shadow removal, and shadow generation. For clarity, only a subset of representative methods is included in each category.}
	\label{fig:timeline}
\end{figure*}

\section{Shadow Physics and Formation}
\label{sec:shadow_physics}
{Shadows arise from the physical interaction of light, objects, and receiver surfaces, governed by precise geometric and photometric principles. Geometrically, \emph{attached shadows} occur on surface regions facing away from the light, while \emph{cast shadows} appear when an occluder blocks light from reaching a different surface. Within a cast shadow, the \emph{umbra} is fully occluded by direct illumination and the \emph{penumbra} exhibits a graded transition due to partial visibility, with appearance modulated by light-source size/extent and surface geometry.}

{A simple photometric model writes the observed intensity as a product of surface reflectance and illumination,}
\begin{equation}
	I(x,y) \;=\; R(x,y)\,\cdot\,L(x,y),
\end{equation}
\xwhu{where $R$ denotes (approximately) view- and illumination-invariant reflectance (albedo) and $L$ captures shading and incident illumination. In shadowed regions, $L$ is reduced primarily through diminished direct visibility, whereas ambient/indirect components remain. Under a Lambertian assumption, an equivalent additive shading form is}
\begin{equation}
	I(x,y) \;=\; \rho(x,y)\,\big(S_{\text{dir}}(x,y) + S_{\text{amb}}(x,y)\big),
\end{equation}
\xwhu{where $\rho$ is diffuse albedo and $S_{\text{dir}}$, $S_{\text{amb}}$ denote direct and ambient shading terms driven by light intensity, direction, softness, and visibility. The smooth variation of visibility across penumbra explains the characteristic soft boundary transitions.}

\xwhu{These principles motivate learning signals and objectives widely used in modern approaches. For detection, intensity/ratio contrasts and boundary-aware features help localize umbra/penumbra. For removal, \emph{attenuation ratios} and \emph{shadow mattes} encourage physically consistent relighting while preserving albedo. For generation, light direction, extent (softness), and receiver geometry guide the placement and spread of synthetic shadows.}
\xwhu{While deep learning models for shadow detection, removal, and generation are data-driven, their effectiveness often stems from implicitly learning features consistent with these physical principles, such as sharp illumination transitions at shadow boundaries or the behavior of diffuse reflection, providing a foundation for their architectural designs and loss functions.}

\section{History and Scope}
\label{sec:1.2}
The analysis of shadow images remains a foundational challenge in computer vision, with a longstanding research emphasis.
Fig.~\ref{fig:timeline} traces the evolution of shadow-related research prior to the deep learning era, covering early computer graphics methods, geometric reasoning studies, 2D shadow detection and removal, and traditional handcrafted or physically based models.
Exploring shadows in computer graphics~\cite{crow1977shadow} has a history spanning half a century, primarily aimed at enhancing the realism of computer-synthesized images.
In the 1980s, specific attention was directed towards studying the relationship between objects (buildings) and their shadows~\cite{irvin1989methods}.
In the 1990s, research ventured into shadow detection and removal in 2D images, with contributions from various studies~\cite{scanlan1990shadow,jiang1992shadow,rosin1995image,funka1995combining}. This line of inquiry expanded in the 2000s, encompassing both images and videos, as demonstrated by works such as~\cite{salvador2002cast,finlayson2002removing,prati2003detecting,salvador2003spatio,nadimi2004physical,salvador2004cast,finlayson2006removal,wu2007natural,liu2008texture,finlayson2009entropy}.
Later on, machine learning algorithms with hand-crafted features were extensively studied for shadow detection and removal~\cite{lalonde2010detecting,zhu2010learning,guo2011single,huang2011characterizes,guo2013paired,gong2014interactive,vicente2015leave,gryka2015learning}.
Since 2014~\cite{khan2014automatic}, algorithms based on deep learning~\cite{hu2020shadow} have exhibited promising performance, solidifying their status as the primary approaches for shadow analysis.
This paper surveys the landscape of image and video shadow detection, removal, and generation over the past decade in the era of deep learning, with an overview summarized in Fig.~\ref{fig:timeline}.
%


We clarify the scope and selection criteria by stating that our survey considers deep learning-based methods published between 2014 and 2025 in major computer vision and graphics venues (e.g., CVPR, ICCV, ECCV, SIGGRAPH, SIGGRAPH Asia, AAAI, ACM MM, TPAMI, IJCV, TCSVT), selecting only works that (i) use modern learnable architectures (CNN, Transformer, diffusion, etc.), (ii) report quantitative results on public datasets, and (iii) directly address shadow detection, removal, or generation.

This paper does not cover shadow analysis in remote sensing, but it is worth briefly outlining how that domain differs from standard image and video shadow analysis. 
Remote-sensing-based shadow analysis often relies on multi-modal inputs, including radar, visible-spectrum, and infrared imagery, and focuses on large-scale aerial or satellite perspectives rather than ground-level views. Such settings introduce unique challenges, such as the integration of heterogeneous data types, geometric distortions from varying sensor altitudes and viewing angles, and illumination variations caused by atmospheric conditions and surface roughness. 
The shadows captured from a top-down viewpoint differ fundamentally in shape and appearance from those observed in natural perspective images.
These differences make shadow analysis in remote sensing a distinct subfield with its own objectives and methodologies, which are comprehensively reviewed in~\cite{alavipanah2022shadow}.

\begin{table*}[tp]
\caption{Deep models for image shadow detection. 
\xwhu{Methods are grouped by learning paradigm and architectural design. 
*: denotes real-time detector; \#: denotes additional supervision; \$: denotes extra training data.}}
\vspace{-1mm}
\label{tab:image_shadow_detection}
\centering
\setlength{\tabcolsep}{2.2pt}  
\resizebox{\textwidth}{!}{
\begin{tabular}{c|c|c|c|c|c|c|c|c}
\hline
\xwhu{Years} & \xwhu{Refs.} & \xwhu{Methods} & \xwhu{Publications} & \xwhu{Backbones} &
\xwhu{Architecture Type} & \xwhu{Supervision} & \xwhu{Key Innovation / Contribution} & \xwhu{Learning Paradigm} \\ \hline

\multicolumn{9}{c}{\textbf{Component Learning (Patch/Feature-based CNN + CRF/Optimization)}} \\ \hline
2014 & \cite{khan2014automatic,khan2016automatic} & CNN-CRF & CVPR & 7-layer CNN & CNN + CRF & Full & Boundary + smoothness & Component \\
2015 & \cite{shen2015shadow} & SCNN-LinearOpt & CVPR & 7-layer CNN & CNN + LinearOpt & Full & Structural consistency & Component \\
2016 & \cite{vicente2016large,hou2021large} & Stacked-CNN & ECCV & VGG16 & FCN + Patch CNN & Full & Intensity prior & Component \\
2018 & \cite{hosseinzadeh2018fast} & Patched-CNN & IROS & 7-layer CNN & Patch CNN & Full & Statistical prior & Component \\ \hline

\multicolumn{9}{c}{\textbf{Single-Task End-to-End Detection (CNN / Transformer / Attention)}} \\ \hline
2017 & \cite{nguyen2017shadow} & scGAN & ICCV & U-Net & CNN + GAN & Full & Adversarial + mask ratio & Single-task \\
2018 & \cite{Hu_2018_CVPR,hu2020direction} & DSC & CVPR/TPAMI & VGG16 & CNN + Direction-Aware & Full & Directional context & Single-task \\
2018 & \cite{wang2018densely} & DC-DSPF & IJCAI & VGG16 & Multi-branch CNN & Full & Deep supervision & Single-task \\
2018 & \cite{mohajerani2018cpnet} & CPNet & MMSP & U-Net & CNN & Full & Reconstruction & Single-task \\
2018 & \cite{le2018a+d} & A+D Net* & ECCV & U-Net & CNN & Full & Data augment (A-Net) & Single-task \\
2018 & \cite{zhu2018bidirectional} & BDRAR & ECCV & ResNeXt101 & CNN & Full & Context refinement & Single-task \\
2019 & \cite{zheng2019distraction} & DSDNet\# & CVPR & ResNeXt101 & CNN & Full & FP/FN regularization & Single-task \\
2019 & \cite{mohajerani2019shadow} & CPAdv-Net & TIP & U-Net & CNN & Full & Adversarial robustness & Single-task \\
2020 & \cite{luo2020deeply} & DSSDNet & P\&RS & – & Encoder–Decoder & Full & Progressive fusion & Single-task \\
2021 & \cite{hu2021revisiting} & FSDNet* & TIP & MobileNetV2 & Lightweight CNN & Full & Detail enhancement & Single-task \\
2021 & \cite{fang2021robust} & ECA & ACM MM & ResNet101 & Multi-kernel CNN & Full & Scale-aware & Single-task \\
2021 & \cite{liao2021shadow} & RCMPNet\# & ACM MM & ResNet & CNN + LSTM & Full & Confidence regression & Single-task \\
2022 & \cite{zhu2022single} & SDCM & ACM MM & EfficientNet-B3 & Dual-Branch CNN & Full & Discriminative / identity & Single-task \\
2022 & \cite{jie2022fast} & TransShadow & ICASSP & EfficientNet-B1 & Transformer Hybrid & Full & Multi-Scale Consistency & Single-task \\ \hline

\multicolumn{9}{c}{\textbf{Multi-Task Detection \& Removal (Joint or Cascaded Learning)}} \\ \hline
2018 & \cite{wang2018stacked} & ST-CGAN & CVPR & U-Net & CNN + GAN Cascade & Full & Adversarial / reconstruction & Multi-task \\
2019 & \cite{ding2019argan} & ARGAN / ARGAN+SS\$ & ICCV & CNN + LSTM & CNN + Attn. Recurrent & Full/Semi & Adversarial + consistency & Multi-task \\
2023 & \cite{valanarasu2023fine} & R2D\# & WACV & ResNeXt101 & CNN + Detector Block & Full & Context refinement & Multi-task \\
2023 & \cite{yucel2023lra} & LRA\&LDRA\$ & WACV & – & Residual Stack & Full & Reconstruction + color blend & Multi-task \\
2023 & \cite{cong2023sddnet} & SDDNet & ACM MM & EfficientNet-B3 & Dual-Branch CNN & Full & Style / Gram constraint & Multi-task \\
2023 & \cite{sun2023adaptive} & AIM (Sun et al.) & ICCV & VGG16 + ConvNeXt & CNN + Illum. Mapping & Full & Illumination consistency & Multi-task \\ \hline

\multicolumn{9}{c}{\textbf{Semi- / Weakly- / Self-Supervised Learning}} \\ \hline
2020 & \cite{chen2020multi} & MTMT-Net\$ & CVPR & ResNeXt101 & CNN + Mean Teacher & Semi & Consistency / multi-Task & Multi-task \\
2023 & \cite{wu2023many} & SDTR / SDTR+\$* & TCSVT & MiT-B2 & Transformer & Semi/Weak & Reliability selection & Single-task \\
2021 & \cite{zhu2021mitigating} & FDRNet & ICCV & EfficientNet-B3 & CNN + Self-Supervised & Self & Intensity invariance & Multi-task \\
2023 & \cite{yang2023silt} & SILT\$ & ICCV & U-Net + PVTv2 & Hybrid CNN–Transformer & Self & Label tuning / aug. & Single-task \\ \hline

\multicolumn{9}{c}{\textbf{Large Vision Model Adaptation (Prompt-based Fine-tuning)}} \\ \hline
2023 & \cite{chen2023sam} & SAM-Adapter & ICCVW & SAM (ViT-H) & Adapter-based Transformer & Full & Fine-tuned decoder & Single-task \\
2023 & \cite{chen2023make} & ShadowSAM & TGRS & SAM (ViT-B) & Prompt + MLP & Un/Fully & Illumination-guided & Single-task \\
2023 & \cite{jie2023adaptershadow} & AdapterShadow & arXiv & SAM (ViT-B)+Eff.B1 & Adapter Hybrid & Full & Grid prompting & Single-task \\ \hline
\end{tabular}}
\vspace{-2.5mm}
\end{table*}

\section{Shadow Detection}
\label{sec:detection}

\xwhu{Shadow detection aims to predict binary or instance-level masks that delineate shadow regions in images or videos. Accurate shadow localization serves as the foundation for high-level vision tasks such as shadow removal, generation, and relighting, and further benefits object detection, scene understanding, and video analysis by improving illumination awareness and geometry reasoning.}

\xwhu{This section provides a reorganized and analytical overview of deep-learning-based shadow detection methods, categorized by architectural paradigm, supervision strategy, and learning objective. It also summarizes datasets, metrics, and empirical comparisons, revealing trends and challenges that guide future research.}

\subsection{Deep Models for Image Shadow Detection}
\xwhu{Table~\ref{tab:image_shadow_detection} presents representative deep-learning methods for image shadow detection, organized by architectural design (CNN-based, multi-branch, transformer-based, large vision models), supervision strategy (fully-, semi-, and self-supervised), and learning objective (boundary-aware, context-aware, or illumination-consistent losses).}

\subsubsection{Component Learning (Feature-level and Patch-based Models)}
\xwhu{Early works relied on CNNs combined with hand-crafted post-processing (e.g., CRF or optimization) to infer shadows at superpixel or patch level, marking the transition from classical to deep paradigms.}
\begin{itemize}[leftmargin=*]
\item \textbf{CNN-CRF}~\cite{khan2014automatic,khan2016automatic} adopts multiple CNNs to extract superpixel-level and boundary-level features, followed by CRF refinement to generate smooth and contiguous masks.  
\item \textbf{SCNN-LinearOpt}~\cite{shen2015shadow} uses a CNN to detect local shadow edges and applies least-squares optimization to enforce structural consistency.  
\item \textbf{Stacked-CNN}~\cite{vicente2016large,hou2021large} combines a global FCN-based prior map and local patch CNNs for region-wise prediction, later fused by weighted averaging.  
\item \textbf{Patched-CNN}~\cite{hosseinzadeh2018fast} integrates statistical priors with CNN-based patch prediction to accelerate inference on small images.  
\end{itemize}
\xwhu{Although influential, these approaches lacked end-to-end training and global illumination understanding, motivating later architectures that directly regress full-resolution masks.}

\subsubsection{Single-Task End-to-End Learning}
\xwhu{With the rise of encoder–decoder architectures and adversarial training, shadow detection evolved toward end-to-end systems that predict shadow masks directly from RGB inputs. These models exploit hierarchical context, attention, and direction-aware priors to capture both global illumination and fine boundary cues.}
\begin{itemize}[leftmargin=*]
\item \textbf{scGAN}~\cite{nguyen2017shadow} introduces a conditional GAN that controls mask density via a sensitivity parameter, highlighting the trade-off between recall and precision.  
\item \textbf{DSC}~\cite{Hu_2018_CVPR,hu2020direction} proposes direction-aware spatial context modules that encode illumination orientation and edge continuity, a design later reused in removal models.  
\item \textbf{DC-DSPF}~\cite{wang2018densely} employs densely cascaded fusion with deep supervision to progressively refine mask boundaries.  
\item \textbf{CPNet}~\cite{mohajerani2018cpnet} integrates residual U-Net connections to maintain texture fidelity while suppressing artifacts.  
\item \textbf{A+D Net}~\cite{le2018a+d} introduces a two-stage training strategy, which is the attenuation-based data augmentation (A-Net) followed by detection (D-Net) to achieve real-time inference.  
\item \textbf{BDRAR}~\cite{zhu2018bidirectional} introduces recurrent attention and bidirectional feature pyramids to propagate contextual cues between scales.  
\item \textbf{DSDNet}~\cite{zheng2019distraction} models false-positive and false-negative distributions explicitly, making the network aware of detection uncertainty.  
\item \textbf{CPAdv-Net}~\cite{mohajerani2019shadow} strengthens robustness by introducing adversarial perturbations and feature remapping within skip connections.  
\item \textbf{DSSDNet}~\cite{luo2020deeply} extends detection to aerial imagery via residual encoder–decoder structures and progressive fusion.  
\item \textbf{FSDNet}~\cite{hu2021revisiting} reuses DSC modules within a lightweight MobileNetV2 backbone for real-time, mobile-efficient detection.  
\item \textbf{ECA}~\cite{fang2021robust} employs multi-kernel convolutions for scale-adaptive feature extraction.  
\item \textbf{RCMPNet}~\cite{liao2021shadow} predicts confidence maps via attention-based LSTMs, estimating per-pixel reliability of shadow detection.  
\item \textbf{SDCM}~\cite{zhu2022single} decouples shadow/non-shadow streams with identity reconstruction and discriminative losses to enhance contrast.  
\item \textbf{TransShadow}~\cite{jie2022fast} leverages transformer-based multi-scale feature fusion to distinguish subtle penumbra transitions with reduced latency.  
\end{itemize}

\xwhu{Across this group, the field has transitioned from local patch-based networks to globally context-aware and transformer-augmented architectures, progressively improving robustness in complex illumination.}

\subsubsection{Multi-Task and Cross-Modal Learning}
\xwhu{Multi-task frameworks jointly optimize shadow detection and removal, leveraging shared representations to improve consistency and generalization.}
\begin{itemize}[leftmargin=*]
\item \textbf{ST-CGAN}~\cite{wang2018stacked} employs stacked conditional GANs: one predicts masks, another reconstructs shadow-free images, establishing an early link between detection and removal.  
\item \textbf{ARGAN}~\cite{ding2019argan} adopts recurrent attention generators for coarse-to-fine refinement of both attention maps and de-shadowed images, operating even with unlabeled data.  
\item \textbf{R2D}~\cite{valanarasu2023fine} reuses feature discriminators from removal to enhance fine-grained detection, showing explicit inter-task transfer.  
\item \textbf{LRA\&LDRA}~\cite{yucel2023lra} introduces residual stack optimization for simultaneous detection and reconstruction, improving color blending and fidelity.  
\item \textbf{SDDNet}~\cite{cong2023sddnet} disentangles style and illumination layers via dual supervision and Gram-based style constraints, offering interpretable feature separation.  
\item \textbf{AIM}~\cite{sun2023adaptive} integrates adaptive illumination mapping for shadow-aware tone transformation, coupled with contrast-based detection feedback.  
\end{itemize}
\xwhu{These methods underscore the bidirectional benefits between detection and removal, where masks inform reconstruction and relighting cues improve mask precision, foreshadowing unified multi-task pipelines.}

\subsubsection{Semi- and Weakly-Supervised Learning}
\xwhu{Due to costly mask annotation, semi- and weakly-supervised learning has become crucial for scaling detection to diverse domains.}
\begin{itemize}[leftmargin=*]
\item \textbf{ARGAN+SS}~\cite{ding2019argan} extends ARGAN to utilize unlabeled data with adversarial consistency constraints.  
\item \textbf{MTMT-Net}~\cite{chen2020multi} introduces a mean-teacher scheme that aligns student–teacher predictions across tasks (mask, edge, and count), ensuring stability under limited supervision.  
\item \textbf{SDTR / SDTR+}~\cite{wu2023many} leverage reliable-sample selection and flexible annotations (boxes, scribbles, points) to construct pseudo masks, achieving real-time operation via MiT-B2 backbones~\cite{xie2021segformer}.  
\end{itemize}
\xwhu{Such semi-supervised frameworks expand scalability while maintaining generalization, demonstrating that consistency and pseudo-label refinement effectively replace dense data annotation.}

\subsubsection{Self-Supervised Learning}
\xwhu{Self-supervised learning further relaxes supervision by constructing intrinsic pretext tasks, enabling networks to learn illumination-invariant representations.}
\begin{itemize}[leftmargin=*]
\item \textbf{FDRNet}~\cite{zhu2021mitigating} decomposes features into intensity-variant and invariant components using brightness-adjusted self-supervision, reducing over-reliance on luminance cues.  
\item \textbf{SILT}~\cite{yang2023silt} employs shadow-aware label tuning and data augmentation with shadow-free or dark-object Internet images, leveraging U-Net backbones such as ResNeXt101~\cite{xie2017aggregated}, EfficientNet~\cite{tan2019efficientnet}, and PVTv2~\cite{wang2022pvt}.  
\end{itemize}
\xwhu{These designs highlight that data-driven self-regularization can mitigate overfitting to luminance and enhance discrimination between shadows and dark objects.}

\subsubsection{Large Vision Models and Prompt-based Adaptation}
\xwhu{Recent advances in foundation segmentation models have extended to shadow detection. Although SAM~\cite{kirillov2023segment} offers strong zero-shot segmentation, its global priors struggle with subtle, context-dependent shadows, motivating task-specific adaptation.}
\begin{itemize}[leftmargin=*]
\item \textbf{SAM-Adapter}~\cite{chen2023sam} integrates trainable adapters into the SAM encoder and fine-tunes the decoder for improved context integration.  
\item \textbf{ShadowSAM}~\cite{chen2023make} adds illumination-guided prompters and mask diversity regularization for curriculum adaptation, trainable in both unsupervised and supervised modes.  
\item \textbf{AdapterShadow}~\cite{jie2023adaptershadow} embeds adapters within the frozen SAM ViT-H encoder~\cite{dosovitskiy2020image}, using grid-based point prompting and EfficientNet-B1 guidance.  
\end{itemize}
\xwhu{These large vision model adaptations demonstrate the transferability of segmentation priors to illumination-aware tasks, marking a new research frontier for generalizable shadow detection.}

\vspace{2.5mm}
\hspace{-6.5mm} \textbf{Trends and Insights.}
(i) Recent detectors increasingly rely on \emph{shadow-specific cues} such as direction context (DSC~\cite{hu2020direction}) and error-aware boundary refinement (DSDNet~\cite{zheng2019distraction}), showing that modeling light flow and penumbra uncertainty is more effective than simply deepening architectures. These complementary strategies are now unified in transformer-based architectures (e.g., ShadowFormer~\cite{guo2023shadowformer}) that combine long-range attention with fine-grained spatial refinement. 
(ii) Semi and self-supervised methods leverage \emph{illumination ratios, color invariants, and pseudo-masks} to improve transferability, but remain constrained by noise in penumbra regions.
(iii) Boundary, penumbra, and attenuation consistency emerge as the \emph{most stable inductive biases} for cross-scene robustness.
(iv) Detection modules (e.g., DSC, ARGAN, ST-CGAN) are reused in removal and generation, indicating that shadow localization and light–surface reasoning form a \emph{shared core} for all shadow tasks.

\begin{table*}[tp]
\caption{Deep models for video shadow detection. 
\xwhu{Methods are grouped by temporal modeling strategy and architectural design. 
*: denotes real-time detector; \$: denotes additional training data.}}
\vspace{-1mm}
\label{tab:video_shadow_detection}
\centering
\setlength{\tabcolsep}{2.2pt}
\resizebox{\textwidth}{!}{
\begin{tabular}{c|c|c|c|c|c|c|c|c}
\hline
\xwhu{Years} & \xwhu{Refs.} & \xwhu{Methods} & \xwhu{Publications} & \xwhu{Backbones} &
\xwhu{Architecture Type} & \xwhu{Temporal Modeling Strategy} & \xwhu{Supervision} & \xwhu{Learning Paradigm} \\ \hline

\multicolumn{9}{c}{\textbf{Early Frame Alignment and Optical Flow–Based Modeling}} \\ \hline
2021 & \cite{chen2021triple} & TVSD-Net & CVPR & ResNeXt101 & CNN Triple-Branch & Dual Gated Co-Attention + Inter-video Similarity & Full & Single-task \\
2021 & \cite{hu2021temporal} & Hu \textit{et al.} & arXiv & MobileNetV2 & CNN + Flow Warping & Optical Flow–Guided Temporal Alignment & Full & Single-task \\ \hline

\multicolumn{9}{c}{\textbf{Temporal Consistency Learning}} \\ \hline
2022 & \cite{lu2022video} & STICT\$* & CVPR & ResNet50 & CNN + Mean Teacher & Spatio-Temporal Interpolation Consistency & Semi & Single-task \\
2022 & \cite{ding2022learning} & SC-Cor & ECCV & – & Plug-in Correspondence Module & Pixel-to-Set Correspondence Learning & Full & Single-task \\ \hline

\multicolumn{9}{c}{\textbf{Real-Time Spatial–Temporal Fusion Networks}} \\ \hline
2022 & \cite{lin2022spatial} & STF-Net* & VRCAI & Res2Net50 & Lightweight CNN & Spatial–Temporal Fusion Block & Full & Single-task \\ \hline

\multicolumn{9}{c}{\textbf{Contrastive / Aggregation-Based Video Detectors}} \\ \hline
2023 & \cite{liu2023scotch} & SCOTCH \& SODA & CVPR & MiT-B3 & Transformer + CNN & Contrastive + Spatial–Temporal Aggregation & Full & Single-task \\ \hline

\multicolumn{9}{c}{\textbf{Large Vision and Foundation-Model Adaptations}} \\ \hline
2023 & \cite{wang2023detect} & ShadowSAM & TCSVT & SAM(ViT-B)+MobileNetV2 & Foundation Model + LSTM & Long–Short-Term Attention Propagation & Full & Single-task \\ \hline

\multicolumn{9}{c}{\textbf{Multimodal / Diffusion-Based and Domain-Adaptive Frameworks}} \\ \hline
2024 & \cite{Wang2024language} & RSM-Net & ACM MM & ResNet50 + RoBERTa & CNN + Language Fusion & Twin-Track Synergistic Memory (Referring) & Full & Multi-modal (Referring) \\
2024 & \cite{Zhou2024timeline} & TBGDiff & ACM MM & MiT-B3 & Diffusion Transformer & Dual-Scale Temporal Guidance + Boundary Head & Full & Multi-task \\
2024 & \cite{duan2024twostage} & Duan \textit{et al.} & ECCV & SegFormer & Transformer + ControlNet & Temporal/Spatial Adaptation Blocks & Full & Single-task \\ \hline
\end{tabular}}
\vspace{-2.5mm}
\end{table*}

\subsection{Deep Models for Video Shadow Detection}
\label{sec:video_shadow_detection}

\xwhu{Video shadow detection extends single-image detection to dynamic scenes, requiring temporal coherence and robustness to illumination fluctuation, motion blur, and camera jitter. Unlike image detectors that process frames independently, video models must learn spatio-temporal correlations to maintain consistent mask boundaries across time.}
\xwhu{This subsection summarizes representative deep-learning approaches for video shadow detection, categorized by architectural paradigm (CNN-, recurrent-, and transformer-based), supervision strategy (fully vs. semi-supervised), and temporal modeling mechanism (optical flow, memory, contrastive, or diffusion). Table~\ref{tab:video_shadow_detection} shows the key features of methods.}

\begin{itemize}[leftmargin=*] 

\item \textbf{TVSD-Net}~\cite{chen2021triple}, the pioneer in deep-learning-based video shadow detection, employs triple parallel networks collaboratively to obtain discriminative representations at intra-video and inter-video levels. The network includes a dual gated co-attention module to constrain features from neighboring frames in the same video, along with an auxiliary similarity loss for capturing semantic information between different videos. 

\item \textbf{Hu \textit{et al.}}~\cite{hu2021temporal} employs an optical-flow-based warping module to align and combine features between frames, applying it across multiple deep-network layers to extract information from neighboring frames, and encompassing both local details and high-level semantic information. 

\item \textbf{STICT} \cite{lu2022video} uses mean-teacher learning to combine labeled images and unlabeled video frames for real-time shadow detection. It introduces spatio-temporal interpolation consistency training for better generalization and temporal consistency. 

\item \textbf{SC-Cor}~\cite{ding2022learning} employs correspondence learning to improve fine-grained pixel-wise similarity in a pixel-to-set manner, refining pixel alignment within shadow regions across frames. It enhances temporal consistency and seamlessly serves as a plug-and-play module in existing shadow detectors with no computational cost. 

\item \textbf{STF-Net}~\cite{lin2022spatial} efficiently detects shadows in videos in real-time using Res2Net50~\cite{gao2019res2net} as its backbone, introducing a straightforward yet effective spatial-temporal fusion block to leverage both temporal and spatial information. 

\item \textbf{SCOTCH and SODA}~\cite{liu2023scotch} form a video shadow detection framework. SCOTCH uses supervised contrastive loss to enhance shadow feature discrimination, while SODA applies a spatial-temporal aggregation mechanism to manage shadow deformations. This combination improves feature learning and spatial-temporal dynamics. 

\item \textbf{ShadowSAM}~\cite{wang2023detect} fine-tunes SAM~\cite{kirillov2023segment} to detect shadows in the first frame using bounding boxes as prompts and employs a long-short-term network with MobileNetV2 as the backbone to propagate the mask across the video, using long-short-term attention to enhance performance. 

\item \textbf{RSM-Net}~\cite{Wang2024language} introduces the referring video shadow detection task and proposes a referring shadow-track memory network that utilizes a twin-track synergistic memory and mixed-prior shadow attention to segment specific shadows in videos based on descriptive natural language prompts. 

\item \textbf{TBGDiff}~\cite{Zhou2024timeline} is the first diffusion model for video shadow detection by extracting temporal guidance and boundary information, using dual scale aggregation for temporal signal and an auxiliary head for boundary context extraction and timeline temporal guidance via space-time encoded embedding. 

\item \textbf{Duan et al.}~\cite{duan2024twostage} uses a two-stage training paradigm, starting with a pre-trained image-domain model that is adapted to the video domain using a temporal-adaption block for temporal consistency and a spatial-adaption block for integrating high-resolution local patches with global context features. ControlNet~\cite{Zhang_2023_ICCV}-like structure is used in these two blocks. 

\end{itemize}

\xwhu{Overall, video shadow detection research is moving from frame-level supervision toward self- and semi-supervised temporal consistency learning, and from handcrafted motion modeling to implicit temporal reasoning via memory and diffusion.}

\vspace{2.5mm}
\hspace{-6.5mm} \textbf{Trends and Insights.}
(i) Lightweight temporal aggregation proves sufficient for local illumination continuity, and heavy recurrent structures offer limited additional value.
(ii) Emerging cross-modal approaches integrate \emph{referring language, diffusion priors, or temporal geometry cues} to resolve ambiguities that appearance-based models alone cannot.
(iii) Temporal consistency and multimodal grounding are becoming central, suggesting a shift toward \emph{physically and semantically informed video shadow understanding} aligned with broader video segmentation trends.

\subsection{Shadow Detection Datasets}
Next, we exclusively discuss  widely-used datasets for model training and evaluation, omitting other data for additional semi/weakly-supervised training.

\subsubsection{Image Datasets for Shadow Detection}

Earlier datasets, ~\ie, UCF~\cite{zhu2010learning} and UIUC~\cite{guo2011single}, are prepared to train the traditional machine learning methods with hand-crafted features. 
UCF consists of 245 images, featuring 117 captured in diverse outdoor environments, encompassing campus and downtown areas. The remaining images are sourced from existing datasets. For each image, shadows have been meticulously hand-labeled at the pixel level, with validation performed by two individuals.
UIUC has 108 shadow images with the labeled shadow masks and shadow-free images, which is the first to enable quantitative evaluation of shadow removal on dozens of images. 

Later, datasets with thousands of shadow images are collected to train the deep-learning models.
\begin{itemize}[leftmargin=*] 
\item  \textbf{SBU}
\cite{hou2021large, vicente2016noisy, vicente2016large} \& \textbf{SBU-Refine} \cite{yang2023silt}: 
SBU is a large-scale shadow dataset with 4,087 training and 638 testing images, using a lazy labeling approach where users initially coarsely label shadow and non-shadow regions. An optimization algorithm refines these labels.
SBU-Refine relabels the test set manually and refines the noise labels in training set by algorithm. 

\item  \textbf{ISTD}~\cite{wang2018stacked} is a dataset with shadow images, shadow-free images, and shadow masks, designed for shadow detection and removal. It includes 1,330 training images, 540 testing images, and 135 distinct background scenes. See \textit{ISTD+} in Sec.~\ref{sec:shadow_removal_data}.

\textbf{CUHK-Shadow}~\cite{hu2021revisiting} is a large dataset with 10,500 shadow images, including 7,350 for training, 1,050 for validation, and 2,100 for testing. It features five categories: (i) Shadow-ADE: 1,132 ADE20K images (building shadows), (ii) Shadow-KITTI: 2,773 KITTI images (vehicle, tree, roadside shadows), (iii) Shadow-MAP: 1,595 Google Maps photos, (iv) Shadow-USR: 2,445 USR images (people and object shadows), and (v) Shadow-WEB: 2,555 Internet images from Flickr.

\item \textbf{SynShadow}~\cite{inoue2021learning} is a synthetic dataset of 10,000 shadow/ shadow-free/matte image triplets, generated using a shadow illumination model and 3D models. It assumes occluders outside the camera view and flat surfaces for shadow projection, with shadow-free images from USR~\cite{hu2019mask}, supporting pre-training or zero-shot learning.

\item 
\textbf{SARA}~\cite{sun2023adaptive} includes 7,019 raw images with shadow masks, split into 6,143 for training and 876 for testing, featuring shadows from 17 categories across 11 backgrounds.

\end{itemize}

\subsubsection{Video Datasets for Shadow Detection}

\begin{itemize}[leftmargin=*] 
\item  \textbf{ViSha}~\cite{chen2021triple}
features 120 diverse videos with pixel-level shadow annotations  using binary masks. It contains 11,685 frames across 390 seconds, standardized to 30 fps, and is divided into a 5:7 training-testing ratio.


\item \textbf{RVSD}~\cite{Wang2024language} selects 86 videos from ViSha, re-annotating them with separate shadow instances and descriptive natural language prompts, ensuring quality through validation.

\item  \textbf{CVSD}~\cite{duan2024twostage} is a video shadow dataset, containing 196 video clips across 149 categories with diverse shadow patterns. It includes 278,504 annotated shadow areas and 19,757 frames with shadow masks in complex scenarios.

\end{itemize}

\subsection{Evaluation Metrics}

\subsubsection{Evaluation Metrics for Image Shadow Detection}

\begin{itemize}[leftmargin=*] 
\item  \textbf{BER}~\cite{vicente2016large} (Balanced Error Rate) serves as a common metric for assessing shadow detection performance. In this evaluation, shadow and non-shadow regions contribute equally, regardless of their relative areas. The BER is computed using the formula:
\begin{equation}
BER \ = \ (1-\frac{1}{2}(\frac{TP}{TP+FN}+\frac{TN}{TN+FP}))\times 100 \ ,
\end{equation}
where $TP$, $TN$, $FP$, and $FN$ represent true positives, true negatives, false positives, and false negatives, respectively.
To calculate these values, the predicted shadow mask is first quantized into a binary mask. Pixels are set to one if the values exceed $0.5$ and zero otherwise. This binary mask is then compared with the ground-truth mask. A lower BER value indicates a more effective detection result. 
Occasionally, BER values for both shadow and non-shadow regions are also provided.

\item  \textbf{$F_\beta^\omega$-measure}~\cite{margolin2014evaluate,hu2021revisiting} is proposed for evaluating non-binary prediction values in shadow masks. This metric calculates precision and recall in a weighted manner, with a higher $F_\beta^\omega$ indicating a superior result.

\end{itemize}

\subsubsection{Evaluation Metrics for Video Shadow Detection}

The first paper~\cite{chen2021triple} in video shadow detection with deep learning uses the Mean Absolute Error (MAE), F-measure ($F_\beta$), Intersection over Union (IoU), and Balance Error Rate (BER) to evaluate the performance. However, the evaluation  is only  on individual image (frame-level) without capturing the temporal stability. Ding~\emph{et al.}~\cite{ding2022learning} introduces the temporal stability metric.

\begin{itemize}[leftmargin=*] 
\item  \textbf{Temporal Stability (TS)}~\cite{ding2022learning} calculates the optical flow between the ground-truth labels of two adjacent frames, denoted as $Y_t$ and $Y_{t+1}$.  While ARFlow~\cite{liu2020learning} was originally used for optical flow calculation, this paper adopts RAFT~\cite{teed2020raft}. This approach is employed because the motions of shadows are difficult to capture in RGB frames.
Define $I_{t \longrightarrow t+1}$ as the optical flow between $Y_t$ and $Y_{t+1}$. Then, the reconstructed result, which warps $\hat{Y}_{t+1}$ by the optical flow $I_{t \longrightarrow t+1}$, is denoted as $\mathbf{Y}_t$. $T$ is the number of video frames. Next, the temporal stability of video shadow detection is measured based on the flow warping Intersection over Union (IoU) between the adjacent frames:
\begin{equation}
    \text{TS} \ = \ \frac{1}{T-1}\sum^{T-1}_{t=1} \text{IoU}(\hat{Y}_{t}, \mathbf{Y}_t) \ .
\end{equation}
\vspace{-3mm}
\end{itemize}

\begin{table*}[tp]
\centering
\caption{Comparing image shadow detection methods on an NVIDIA GeForce RTX 4090 GPU. \$: additional training data; *: real-time shadow detector; \#: extra supervision from other methods. Note that for the results shown in the rightmost columns, we report the cross-dataset generalization evaluation, where the models were trained on SBU-Refine and tested on SRD. }
\vspace{-1mm}
\label{table:sbu_CUHK-Shadow_evaluation}
\resizebox{1.0\linewidth}{!}{ %
\setlength{\tabcolsep}{4pt}
\begin{tabular}{c|c|c|c|c|c|c|c|c|c||c|c|c}
\hline
\multirow{2}{*}{Input Size} & \multirow{2}{*}{Methods} & \multicolumn{3}{c|}{SBU-Refine} & \multicolumn{3}{c|}{CUHK-Shadow} & \multirow{2}{*}{Param.(M)} & \multirow{2}{*}{Infer.(images/s)} & \multicolumn{3}{c}{\textit{SRD (cross)}} \\ \cline{3-8} \cline{11-13} 
                      &                          & BER $\downarrow$ & BER$_S$ $\downarrow$ & BER$_{NS}$ $\downarrow$ & BER $\downarrow$ & BER$_S$ $\downarrow$ & BER$_{NS}$ $\downarrow$ &                             &                                    & BER $\downarrow$ & BER$_S$ $\downarrow$ & BER$_{NS}$ $\downarrow$ \\ \hline            
\multirow{9}{*}{256 $\times$ 256} 
                        & DSC~\cite{Hu_2018_CVPR,hu2020direction}   &6.79 &9.36 &4.23     &10.97 &7.49 &14.45     &122.49 &26.86   &11.10 &15.82 &6.39     \\ 
                      & BDRAR~\cite{zhu2018bidirectional}              &6.27 &8.21 &4.34          &10.09 &7.30 &12.88         &42.46 &39.76     &9.13 &\textbf{11.42} &6.84     \\ 
                    & DSDNet\#~\cite{zheng2019distraction} & \textbf{5.37} &\textbf{6.65} &4.09         &\textbf{8.56} &\textbf{6.27} &10.84          &58.16 &37.53     &10.29  &14.63  &5.94    \\ 
                    & MTMT-Net\$~\cite{chen2020multi} &6.32 &9.77 &2.86         &8.90 &8.70 &9.10         &44.13 &34.04     &9.97 &14.90 &5.04    \\ 
                    & FDRNet~\cite{zhu2021mitigating}                &5.64 &7.85 &3.43          &14.39 &17.87 &10.91          &10.77 &41.39     &11.82 &17.03 &6.62    \\ 
                    & FSDNet*~\cite{hu2021revisiting}                &7.16 &11.67 &2.64          &9.93 &11.35 &8.51          &\textbf{4.39} &\textbf{150.99}     &12.13 &19.40 &4.87    \\ 
                      & ECA~\cite{fang2021robust}                & 7.08 & 12.51 & \textbf{1.64}     & 8.58 & 11.25 & \textbf{5.91}     &157.76 &27.55     &11.97  &20.38  &\textbf{3.57}    \\ 
                      & SDDNet~\cite{cong2023sddnet}               & 5.39 & 7.17 & 3.61     & 8.66 & 7.85 & 9.47     & 15.02 & 36.73   &\textbf{8.64}  &11.53  &5.74   \\ \hline 
\multirow{9}{*}{512 $\times$ 512} & DSC~\cite{Hu_2018_CVPR,hu2020direction}                &6.34 &8.24 &4.45          &9.53 &6.87 &12.19          &122.49 &22.59     &11.62 &17.06 &6.18    \\ 
                      & BDRAR~\cite{zhu2018bidirectional}              &5.62 &6.50 &4.73          &8.79 &7.71 &9.88          &42.46 &31.34     &8.53 &10.10 &6.97   \\ 
                    & DSDNet\#~\cite{zheng2019distraction} & 5.04 &\textbf{5.47} &4.60          &7.79 &\textbf{6.44} &9.14          &58.16 &32.69     &8.92  &10.58  &7.27    \\ 
                    & MTMT-Net\$~\cite{chen2020multi} &5.79 &8.74 &2.85          &8.32 &10.03 &6.60         &44.13 &28.75     &9.19 &12.86 &5.53   \\ 
                    & FDRNet~\cite{zhu2021mitigating}                &5.39 &7.35 &3.43          &\textbf{6.58} &7.56 &5.59          &10.77 &35.00     &8.81 &12.17 &5.46    \\ 
                    & FSDNet*~\cite{hu2021revisiting}                &6.80 &11.47 &2.13          &8.84 &10.29 &7.39          &\textbf{4.39} &\textbf{134.47}     &11.94 &20.10 &3.79    \\ 
                      & ECA~\cite{fang2021robust}                &7.52 &13.43 &\textbf{1.61}          &7.99 &9.50 &\textbf{5.25}          &157.76 &22.41     &12.71 &22.45 &\textbf{2.97}    \\ 
                      & SDDNet~\cite{cong2023sddnet}               & \textbf{4.86} & 6.42 & 3.31     & 7.65 & 6.57 & 8.74     & 15.02 & 37.65     &\textbf{7.65}  &\textbf{10.04}  &5.27   \\ \hline
 \end{tabular} }
\vspace{-1mm}
\end{table*}

\subsection{Experimental Results}

The reported comparison results among existing methods in their original papers suffer from inconsistencies in input sizes, evaluation metrics, datasets, and implementation platforms. 
Hence, we standardize experimental setting and perform experiments on the same platform across various methods to
ensure a fair comparison.
Besides, we further compare the methods in various aspects, including models' sizes and speeds, and perform cross-dataset evaluation for generalization capability evaluation. 

\subsubsection{Image Shadow Detection}

\begin{figure}[tp]
    \centering
    \begin{subfigure}[b]{0.9\linewidth}
        \centering
        \includegraphics[width=\linewidth]{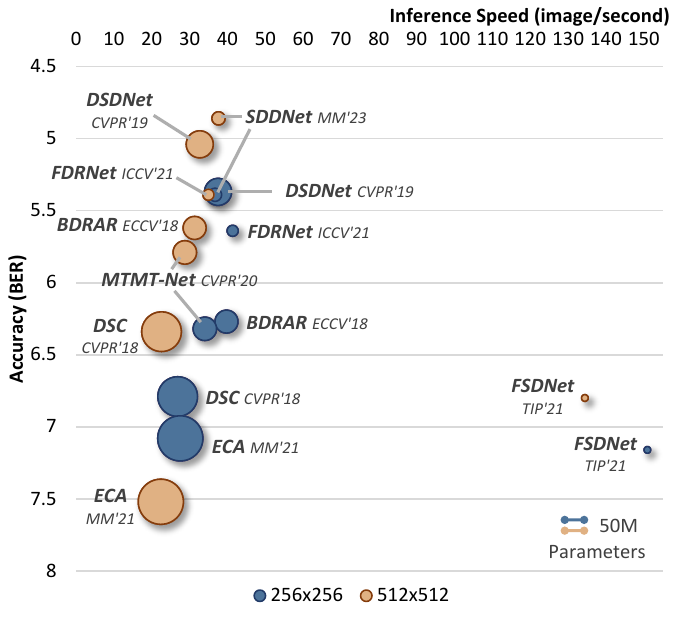}
        \vspace{-4mm}
    \end{subfigure}
    \\[2mm]
    \begin{subfigure}[b]{0.9\linewidth}
        \centering
        \includegraphics[width=\linewidth]{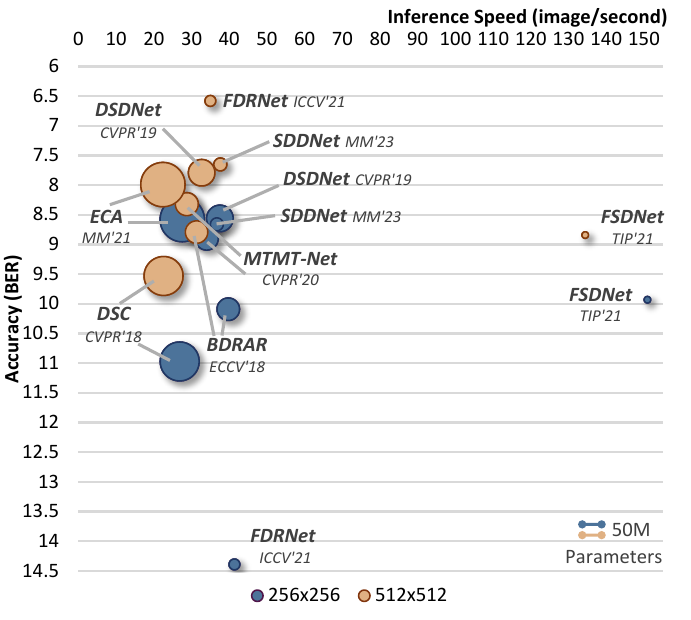}
        \vspace{-4mm}
    \end{subfigure}
    \vspace{-2mm}
    \caption{Shadow detection methods on the SBU-Refine (top) and CUHK-Shadow (bottom) datasets: accuracy, parameters (indicated by the area of the bubbles), and speed.}
    \label{fig:shadow_detection}
    \vspace{-4mm}
\end{figure}

\begin{table*}[tp]
\caption{Comparison of video shadow detection methods on the ViSha dataset using an NVIDIA GeForce RTX 4090 GPU. AVG represents the average score of IoU and TS, reflecting both frame-level and temporal-level IoUs. ShadowSAM uses the size of 1024x1024, while others adopt the size of 512x512.}
\vspace{-1mm}
\centering
\label{table:video_shadow_compare}
\resizebox{0.7\linewidth}{!}{
\begin{tabular}{c|c|c|c|c|c|c}
\hline
Methods & BER $\downarrow$ & IoU [$\%$] $\uparrow$ & TS [$\%$] $\uparrow$ & AVG $\uparrow$ & Param. (M) & Infer. (frames/s) \\
\hline
TVSD-Net \cite{chen2021triple} & 14.21 & 56.36 & 22.69 & 39.53 & 60.83 & 15.50 \\
STICT\$* \cite{lu2022video} & 13.05 & 43.75 & \textbf{39.10} & 41.43 & \textbf{26.17} & \textbf{91.34} \\
SC-Cor \cite{ding2022learning} & 12.80 & 55.56 & 23.68 & 39.62 & 58.16 & 27.91 \\
SCOTCH and SODA \cite{liu2023scotch} & \textbf{10.36} & 61.24 & 25.76 & \textbf{43.50} & 53.11 & 16.16 \\ \hline
ShadowSAM \cite{wang2023detect} & 13.38 & \textbf{61.72} & 23.77 & 42.75 & 93.74 & 15.53 \\

\hline
\end{tabular}
}
\vspace{-2.5mm}
\end{table*}

\hspace{-1.5mm} \textbf{Overall Performance Benchmark Results.}
SBU-Refine \cite{yang2023silt} and CUHK-Shadow \cite{hu2021revisiting} are utilized to assess the performance of various methods. SBU-Refine improves the evaluation accuracy by correcting erroneously labeled masks, thereby reducing overfitting issues in comparison methods. CUHK-Shadow, the largest real dataset, offers a diverse range of scenarios for comprehensive testing.
The methods compared are listed in Table \ref{table:sbu_CUHK-Shadow_evaluation}, and we excluded those for which code is not available.
We retrained the methods using the original source code, except for DSC, which was implemented in PyTorch with a ResNeXt101 backbone. 
\xwhu{All models were trained on the training set of SBU-Refine or CUHK-Shadow.}
\emph{Post-processing, such as CRF, is omitted for all compared methods.}
%
Previous methods adopted various input sizes. In this paper, we set the input sizes to $256 \times 256$ and $512 \times 512$ to present results at two resolutions. 
We take BER as the evaluation metric, calculated using Python code. 
BERs for both shadow ($BER_S$) and non-shadow ($BER_{NS}$) regions are reported.  
Results are resized to the ground-truth resolution in evaluation for fair comparison.

Table \ref{table:sbu_CUHK-Shadow_evaluation} and Fig.~\ref{fig:shadow_detection} illustrate the accuracy, running time, and parameters of each method. We can observe that (i) some relatively older methods perform better than recent ones, indicating an over-fitting issue on the original SBU dataset; (ii) FSDNet~\cite{hu2021revisiting} is the only open-source (both training and testing code available) real-time shadow detector with a few parameters and fast inference speed; (iii) DSDNet~\cite{zheng2019distraction} incorporates the results from DSC~\cite{Hu_2018_CVPR,hu2020direction} and BDRAR~\cite{zhu2018bidirectional} in its training process and achieves comparable performance with the recent method SDDNet~\cite{cong2023sddnet}; (iv) a larger input size usually brings performance gains but also requires more time; and (v) CUHK-Shadow is more challenging than SBU-Refine. FDRNet~\cite{zhu2021mitigating} is particularly sensitive to input resolution when detecting shadows in the CUHK-Shadow, which contains complex shadows or finer details that benefit from higher resolution inputs ($512\times 512$). 
\emph{See the visual comparisons in the appendix.}

\vspace{2.5mm}
\hspace{-6.5mm} \textbf{Cross-Dataset Generalization Evaluation.}
To evaluate the generalization capability of shadow detection methods, we perform cross-dataset evaluation by using the trained models on SBU-Refine training set to detect shadows on the SRD testing set; see Sec.~\ref{sec:shadow_removal_data}. 
SRD is used due to its similar complexity in background features to SBU.
Note that this is the first time to evaluate the generalization capability on a large-scale dataset. 

The three rightmost columns in Table~\ref{table:sbu_CUHK-Shadow_evaluation} show the results, where the performance degrades a lot, especially on the shadow region.
This highlights the importance of cross-dataset evaluation for robust shadow detection. The performance drop in the shadow regions suggests that the methods struggle with varying lighting conditions and complex background textures present in SRD. Future work should focus on improving the robustness of shadow detection models to better generalize across different datasets.

\vspace{2.5mm}
\hspace{-6.5mm} \textbf{Summary.} As demonstrated by the experimental results, \emph{how to develop an efficient and robust model with high detection accuracy for image shadow detection}, especially under complex scenarios, remains a challenging problem.

\begin{table*}[tp]
\caption{Deep models for instance shadow detection. 
}
\label{tab:instance_shadow_detection}
\vspace{-1mm}
\centering
\setlength{\tabcolsep}{3pt} 
\resizebox{\textwidth}{!}{
\begin{tabular}{c|c|c|c|c|c|c|c}
\hline
\xwhu{Data Type} & \xwhu{Years} & \xwhu{Refs.} & \xwhu{Methods} & \xwhu{Publications} & \xwhu{Backbones} & \xwhu{Architecture Type / Key Modules} & \xwhu{Supervision Levels} \\ 
\hline
\multicolumn{8}{c}{\textbf{Image-level Instance Shadow Detection}} \\ \hline
Image & 2020 & \cite{wang2020instance} & LISA & CVPR & ResNeXt101-FPN & \xwhu{Two-stage Mask R-CNN + Association Head + Light Dir. Est.} & Fully supervised \\ 
Image & 2021 & \cite{wang2021single} & SSIS & CVPR & ResNet101-BiFPN & \xwhu{Single-stage FCN + Bidirectional Relation Learning} & Fully supervised \\ 
Image & 2023 & \cite{wang2023instance} & SSISv2 & TPAMI & ResNet101-BiFPN & \xwhu{Enhanced SSIS + Deformable MaskIoU + Shadow-aware Copy-Paste} & Fully supervised \\ 
\hline
\multicolumn{8}{c}{\textbf{Video-level Instance Shadow Detection}} \\ \hline
Video & 2024 & \cite{xing2024video} & ViShadow & TIP & ResNet101-BiFPN & \xwhu{Semi-supervised Temporal Association + Contrastive / Cycle Consistency Loss} & Semi-supervised \\ 
\hline
\end{tabular}}
\vspace{-2mm}
\end{table*}

\subsubsection{Video Shadow Detection}

ViSha~\cite{chen2021triple} is used to evaluate video shadow detection methods with an input size of $512\times 512$, following~\cite{hu2021temporal,liu2023scotch}. ShadowSAM uses $1024\times 1024$ due to the SAM pre-trained model's positional embeddings. 
SC-Cor \cite{ding2022learning} uses the DSDNet~\cite{zheng2019distraction} as the basic network.
STICT \cite{lu2022video} uses additional SBU dataset images for training. 
Except for commonly-used metrics BER and IoU for image-level evaluation, we also adopt Temporal Stability (TS), often ignored by the compared methods. 
Results are resized to $512\times 512$ for optical flow in TS and to ground-truth resolution for other metrics.

%

Table \ref{table:video_shadow_compare} shows the results, showing distinct advantages and trade-offs among video shadow detection methods. SCOTCH and SODA exhibit the best overall performance with the lowest BER and highest AVG, while ShadowSAM achieves the highest IoU but with a larger model size. STICT stands out for its fastest inference speed, making it ideal for real-time applications despite a lower IoU. SC-Cor and TVSD-Net show balanced performances with moderate BER, IoU, and TS scores. 

\vspace{2.5mm}
\hspace{-6.5mm} \textbf{Summary.} As demonstrated by the experimental results,  \emph{how to achieve an optimal balance between frame-level accuracy, temporal stability, model complexity, and inference speed in video shadow detection} remains a challenging problem.

\begin{table*}[tp]
\caption{Comparisons of image instance shadow detection methods. Speed evaluated on an NVIDIA GeForce RTX 4090 GPU.}
\label{table:soba_comprehensive_eval}
\vspace{-1mm}
\centering
\resizebox{0.99\linewidth}{!}{%
\begin{tabular}{c}
\hline
{Evaluation on the \textit{SOBA-testing} Set} \\
\hline
\begin{tabular}{c|c|c|c|c|c|c|c|c}
Methods & $SOAP_{segm}$ $\uparrow$ & $SOAP_{bbox}$ $\uparrow$ & Asso. $AP_{segm}$ $\uparrow$ & Asso. $AP_{bbox}$ $\uparrow$ & Ins. $AP_{segm}$ $\uparrow$ & Ins. $AP_{bbox}$ $\uparrow$ & Param. (M) & Infer. (images/s) \\ \hline 
LISA~\cite{wang2020instance}   & 23.5        & 21.9        & 42.7                  & 50.4                  & 39.7               & 38.2               & 91.26 & \textbf{8.16} \\
SSIS~\cite{wang2021single}   & 29.9        & 26.8        & 52.3                  & 59.2                  & 43.5               & 41.5               & \textbf{57.87} & 5.83 \\
SSISv2~\cite{wang2023instance} & \textbf{35.3}        & \textbf{29.0}        & \textbf{59.2}                  & \textbf{63.0}                 & \textbf{50.2}               & \textbf{44.4}               & 76.77 & 5.17 \\    
\end{tabular} \\
\hline
{Evaluation on the \textit{SOBA-challenge} Set} \\
\hline
\begin{tabular}{c|c|c|c|c|c|c|c|c}
Methods & $SOAP_{segm}$ $\uparrow$ & $SOAP_{bbox}$ $\uparrow$ & Asso. $AP_{segm}$ $\uparrow$ & Asso. $AP_{bbox}$ $\uparrow$ & Ins. $AP_{segm}$ $\uparrow$ & Ins. $AP_{bbox}$ $\uparrow$ & Param. (M) & Infer. (images/s) \\ \hline   
LISA~\cite{wang2020instance}   & 10.2        & 9.8         & 21.6                  & 26.0                  & 23.9               & 24.7               & 91.26 & \textbf{4.52} \\
SSIS~\cite{wang2021single}   & 12.8        & 12.9        & 28.4                  & 32.5                  & 25.7               & 26.5               & \textbf{57.87} & 2.26 \\
SSISv2~\cite{wang2023instance} & \textbf{17.7}        & \textbf{15.0}        & \textbf{34.5}                  & \textbf{37.2}                  & \textbf{31.0}               & \textbf{28.4}               & 76.77 & 1.91 \\    
\end{tabular} \\
\hline
\end{tabular} }
\vspace{-4mm}
\end{table*}

\section{Instance Shadow Detection}
\label{sec:instance}

This section introduces another task, instance shadow detection, which aims to find shadows together with their associated objects.
Knowing the relations between the objects and their shadows benefits lots of image/video editing applications, since it is easy to manipulate objects with their associated shadows simultaneously.
This task is first formulated by~\cite{wang2020instance} at the image level, and then extended to videos by~\cite{xing2024video}.
Table~\ref{tab:instance_shadow_detection} encapsulates the essential properties of the surveyed methods.
\xwhu{Compared with pixel-wise detection, instance-level modeling establishes a physical and semantic linkage between casting objects and shadow regions, bridging geometry, illumination, and semantics.}

\subsection{Deep Models for Image Instance Shadow Detection}

Instance shadow detection aims to detect shadow instances and the associated object instances that cast each shadow.
\xwhu{This task lies between object detection and shadow segmentation, requiring networks to reason jointly over geometry (light direction, projection) and semantics (object identity).}

\begin{itemize}[leftmargin=*] 
\item  \textbf{LISA}~\cite{wang2020instance} initiates by generating region proposals likely to contain shadow/object instances and their associations. 
For each proposal, it predicts bounding boxes and masks for individual shadow/object instances, generates bounding boxes for shadow–object associations (pairs), and estimates the light direction for each shadow–object association. 
\xwhu{It formalized the first explicit association head that jointly predicts bounding boxes and geometric cues, creating a foundation for physically consistent shadow–object reasoning.}

\item \textbf{SSIS}~\cite{wang2021single} introduces a single-stage fully convolutional network with a bidirectional relation-learning module for end-to-end learning of relations between shadow and object instances. 
\xwhu{By directly learning offset vectors between paired centers, it eliminates the need for proposal-level pairing, improving efficiency and structural compactness.}
This module learns offset vectors from the center of each shadow instance to its associated object instance, and vice versa.

\item \textbf{SSISv2}~\cite{wang2023instance} extends SSIS with several improvements, including a deformable MaskIoU head, a shadow-aware copy–paste data augmentation strategy, and a boundary loss to enhance segmentation of both shadow/object instances and their associations. 
\xwhu{These upgrades yield more robust detection under cluttered illumination and fine-grained penumbra boundaries, marking a shift from geometric pairing to context-aware relational learning.}

\end{itemize}

\xwhu{In summary, the evolution from LISA to SSISv2 reflects a transition from proposal-driven multi-branch frameworks to unified relational detectors that jointly encode geometry, semantics, and boundary consistency.}

\subsection{Deep Models for Video Instance Shadow Detection}

Video instance shadow detection entails not just identifying shadows and their associated objects in video frames, but also continuously tracking each shadow, object, and their associations throughout the entire video sequence, accommodating even temporary disappearance of shadow or object parts within associations.
\xwhu{This introduces the additional dimension of temporal coherence, where association reasoning must persist across frames despite motion, occlusion, or illumination variation.}

\begin{itemize}[leftmargin=*] 
\item  \textbf{ViShadow}~\cite{xing2024video} is a semi-supervised framework trained on labeled image data and unlabeled video sequences. 
Initial training involves pairing shadows and objects across different images using center-contrastive learning. 
Subsequently, unlabeled videos are leveraged with a cycle-consistency loss to enhance temporal association tracking. 
It also addresses the challenge of temporary disappearance of object or shadow instances by a retrieval mechanism.
\xwhu{ViShadow effectively bridges static and temporal domains, demonstrating how contrastive pairing and cross-frame consistency can extend spatial associations into long-range temporal reasoning.}
\end{itemize}

\vspace{2.5mm}
\hspace{-6.5mm} \textbf{Trends and Insights.}
(i) The field moves from heuristic pairing (e.g., LISA) toward \emph{learned relational reasoning} (e.g., SSIS/SSISv2), where detectors jointly infer object–shadow associations rather than pairing post hoc.
(ii) Newer models enforce structural coherence by coupling \emph{object geometry, light direction, and shadow boundaries}, improving plausibility in cluttered scenes.
(iii) Early temporal extensions show that instance reasoning naturally benefits from \emph{multi-frame or multi-view} cues, especially for disambiguating overlapping or interacting shadows.
(iv) Instance-level outputs provide structured priors that directly condition removal or generation, reinforcing their role as a \emph{bridge task} in unified pipelines.

\subsection{Instance Shadow Detection Datasets}

\xwhu{Benchmark datasets play a pivotal role in standardizing association reasoning and evaluating generalization across illumination and motion conditions.}

\begin{itemize}[leftmargin=*] 
\item \textbf{SOBA}~\cite{wang2020instance,wang2023instance} is the first dataset for image instance shadow detection, comprising 1,100 images with 4,293 annotated shadow–object associations. 
Initially, 1,000 images were collected by~\cite{wang2020instance}, and~\cite{wang2023instance} added 100 more challenging shadow–object pairs for exclusive testing. 
The training set includes 840 images with 2,999 pairs.

\item  \textbf{SOBA-VID}~\cite{xing2024video} is a dataset crafted for video instance shadow detection, comprising 292 videos with 7,045 frames, divided into 232 training videos (5,863 frames) and 60 testing videos (1,182 frames). 
\xwhu{SOBA-VID provides frame-level and cross-frame annotations for each object–shadow pair, enabling consistent temporal evaluation and retrieval-based benchmarking.} 
The test set includes dense annotations, while the training set provides labels for every fourth frame to facilitate semi-supervised learning.
\end{itemize}

\xwhu{Together, SOBA and SOBA-VID establish standardized benchmarks for evaluating both static and dynamic association reasoning, enabling quantitative comparison across single-frame and sequence-level tasks.}

\subsection{Evaluation Metrics}

\begin{itemize}[leftmargin=*] 
\item \textbf{SOAP}~\cite{wang2020instance,wang2023instance} (Shadow-Object Average Precision) assesses image instance shadow detection performance by computing average precision (AP) with intersection over union (IoU). It extends the criteria for true positives, requiring IoU thresholds for predicted and ground-truth shadow instances, object instances, and shadow-object associations to be greater than or equal to $\tau$. Evaluation is conducted with a specific $\tau$ value of 0.5 (SOAP50) or 0.75 (SOAP75), and an average is computed across a range of $\tau$ values from 0.5 to 0.95 in increments of 0.05 (SOAP).

\item  \textbf{SOAP-VID}~\cite{xing2024video}  assesses video instance shadow detection by substituting the IoU in SOAP with the spatio-temporal IoU.

\end{itemize}

\subsection{Experimental Results}

\subsubsection{Evaluation of Image Instance Shadow Detection}

\begin{table*}[tp]
\caption{Cross-dataset generalization evaluation. Models were trained on SOBA and tested on SOBA-VID.}
\vspace{-1mm}
\label{table:soba_cross_sobavid}
\centering
\begin{tabular}{c}
\hline
\begin{tabular}{c|c|c|c|c|c|c}
Methods & $SOAP_{segm}$ $\uparrow$ & $SOAP_{bbox}$ $\uparrow$ & Asso. $AP_{segm}$ $\uparrow$ & Asso. $AP_{bbox}$ $\uparrow$ & Ins. $AP_{segm}$ $\uparrow$ & Ins. $AP_{bbox}$ $\uparrow$ \\ \hline 
LISA~\cite{wang2020instance}   & 22.6      & 21.1        & 44.2       & 53.6     & 39.0     & 37.3               \\
SSIS~\cite{wang2021single}     & 32.1      & 26.6        & 58.6       & 64.0     & 46.4     & 41.0               \\
SSISv2~\cite{wang2023instance} & \textbf{37.0}      & \textbf{26.7}        & \textbf{63.6}       & \textbf{67.5}     & \textbf{51.8}     & \textbf{42.8}               \\    
\end{tabular} \\
\hline
\end{tabular}
\vspace{-4mm}
\end{table*}

\hspace{-1.5mm} \textbf{Overall Performance Benchmark Results.}
SOAP~\cite{wang2020instance,wang2023instance} is used as the dataset and SOBA is the evaluation metric. 
The methods compared are listed in Table \ref{table:soba_comprehensive_eval}. 
We re-train the methods using their original code, resizing the shorter side of input images during training to one of six sizes: 640, 672, 704, 736, 768, or 800. During inference, we resize the shorter side to 800, ensuring the longer side does not exceed 1333.

Table \ref{table:soba_comprehensive_eval} shows the accuracy, running time, and parameters of each method, where we observe that (i) SSISv2 achieves the best performance but with the slowest speed; (ii) all have limited performance to deal with complex scenarios; and (iii) more instances in complex scenarios significantly reduce the inference speed. 
\emph{See the visual comparisons in the appendix.}

\vspace{2.5mm}
\hspace{-6.5mm} \textbf{Cross-Dataset Generalization Evaluation.}
To assess generalization capability, we conducted a cross-dataset evaluation by applying models trained on the SOBA training set to detect image instance shadows/objects in video frames of the SOBA-VID~\cite{xing2024video} testing set. 
Note that there are no temporal consistency evaluation.

Table~\ref{table:soba_cross_sobavid} provides the results, where (i) the trend of the compared methods is consistent with the trend observed on the SOBA testing set, and (ii) the performance does not degrade significantly, demonstrating the powerful generalization capability of the instance shadow detection methods.

\vspace{2.5mm}
\hspace{-6.5mm} \textbf{Summary.} As demonstrated by the experimental results, \emph{how to develop an efficient model for accurate segmentation of both shadow and object instances} remains challenging.

\subsubsection{Evaluation of Video Instance Shadow Detection}
Here, we present the performance metrics of ViShadow~\cite{xing2024video} on the SOBA-VID test set: SOAP-VID at 39.6, Association AP at 61.5, and Instance AP at 50.9.  The total inference time for 20 frames is 93.63 seconds, processing about 0.21 frames per second, with 66.26M model parameters. 


\begin{table*}[t]
\caption{Deep models for image shadow removal. \$ denotes using additional training data.}
\vspace{-1mm}
\label{tab:image_shadow_removal}
\centering
\setlength{\tabcolsep}{3.6pt}
\renewcommand{\arraystretch}{1.1}
\resizebox{\textwidth}{!}{%
\begin{tabular}{c|c|c|c|c|c|c}
\hline
Years & Refs. & Methods & Publications & Architecture Type & Supervision & Key Innovation / Contribution \\ 
\hline

\multicolumn{7}{c}{\textbf{Supervised - CNN-based}} \\ 
\hline
2016 & \cite{khan2016automatic} & CNN-CRF & TPAMI & CNN + CRF & Full (paired+mask) & {CRF smoothness, reconstruction} \\
2017 & \cite{qu2017deshadownet} & DeshadowNet & CVPR & CNN (encoder–decoder) & Full (paired) & {Reconstruction (L1/SSIM)} \\
2019 & \cite{le2019shadow} & SP+M-Net & ICCV & CNN (dual-branch) & Full (paired+mask) & {Shadow matte estimation, smoothness constraint} \\
2020 & \cite{hu2020direction} & DSC & TPAMI & CNN (direction-aware) & Full (paired) & {Directional context, residual reconstruction} \\
2020 & \cite{cun2020towards} & DHAN+DA & AAAI & CNN (attention-based) & Full (paired+mask) & {Attention loss, shadow mask synthesis} \\
2021 & \cite{le2021physics} & SP+M+I-Net & TPAMI & CNN (triple-branch) & Full (paired+mask) & {Penumbra reconstruction, matte smoothness, inpainting} \\
2021 & \cite{fu2021auto} & Auto & CVPR & CNN (fusion-based) & Full (paired+mask) & {Boundary consistency, adaptive fusion} \\
2021 & \cite{chen2021canet} & CANet & ICCV & CNN (context-aware) & Full (paired) & {Patch correspondence, reconstruction} \\
2022 & \cite{zhu2022efficient} & EMDNet & AAAI & CNN (model-driven) & Full (paired+mask) & {Iterative fidelity and regularization losses} \\
2022 & \cite{zhu2022bijective} & BMNet & CVPR & Invertible CNN & Full (paired+mask) & {Bijective mapping, color-invariant guidance} \\
2022 & \cite{gao2022towards} & G2C-DeshadowNet & CVPRW & CNN (two-stage) & Full (paired+mask) & {Grayscale-to-color transfer, attention fusion} \\
2022 & \cite{wan2022style} & SG-ShadowNet & ECCV & CNN (style-guided) & Full (paired+mask) & {Style transfer, prototypical normalization} \\
2023 & \cite{liu2023structure} & MStructNet & TIP & CNN (structure prior) & Full (paired+mask) & {Structure consistency loss} \\
2023 & \cite{Vasluianu_2023_CVPR} & DNSR & CVPRW & CNN (dynamic conv.) & Full (paired+mask) & {Exposure adjustment, distillation refinement} \\
2023 & \cite{Cui_2023_CVPR} & PES & CVPRW & CNN (pyramid input) & Full (paired) & {Hybrid perceptual and reconstruction objectives} \\
2023 & \cite{li2023leveraging} & Inpaint4shadow\$ & ICCV & CNN (inpainting) & Full (paired+mask) & {Feature fusion, inpainting-guided reconstruction} \\
2023 & \cite{sen2023shards} & SHARDS & WACV & CNN (two-stage HR) & Full (paired+mask) & {LR-to-HR refinement, CBAM attention} \\
2024 & \cite{wang2024progressive} & PRNet & CVIU & CNN + RNN & Full (paired+mask) & {Progressive refinement, ConvGRU update} \\
\hline

\multicolumn{7}{c}{\textbf{Supervised - GAN-based}} \\ 
\hline
2018 & \cite{wang2018stacked} & ST-CGAN & CVPR & Stacked cGANs & Full (paired+mask) & {Adversarial + consistency (detection + removal)} \\
2019 & \cite{sidorov2019conditional} & AngularGAN & CVPRW & GAN & Full (paired) & {Adversarial reconstruction on synthetic pairs} \\
2019 & \cite{ding2019argan} & ARGAN/ARGAN+SS\$ & ICCV & Attentive recurrent GAN & Full/Semi & {Adversarial + attention, semi-supervised consistency} \\
2020 & \cite{zhang2020ris} & RIS-GAN & AAAI & Multi-gen/disc GAN & Full (paired) & {Residual illumination estimation, adversarial loss} \\
2023 & \cite{liu2023shadow} & TBRNet & TNNLS & Multi-branch GAN & Full (paired) & {Matte + reconstruction + adversarial objectives} \\
\hline

\multicolumn{7}{c}{\textbf{Supervised - Transformer / Hybrid}} \\ 
\hline
2022 & \cite{wan2022crformer} & CRFormer & arXiv & CNN+Transformer & Full (paired+mask) & {Region-aware cross-attention, reconstruction} \\
2022 & \cite{yu2022cnsnet} & CNSNet & ECCVW & Hybrid CNN+Transformer & Full (paired+mask) & {Normalization alignment, cross-aggregation} \\
2023 & \cite{guo2023shadowformer} & ShadowFormer & AAAI & Transformer & Full (paired+mask) & {Channel–context interaction attention} \\
2023 & \cite{10191081} & SpA-Former & IJCNN & Hybrid Transformer & Full (paired+mask) & {Fourier residual blocks, DSC-like attention} \\
2023 & \cite{Chang_2023_CVPR} & TSRFormer & CVPRW & Two-stage Transformer & Full (paired) & {Residual suppression, content refinement} \\
2024 & \cite{li2024shadowmaskformer} & ShadowMaskFormer & arXiv & Transformer & Full (paired+mask) & {Mask-embedded tokens, binarization prior} \\
2024 & \cite{dong2024shadowrefiner} & ShadowRefiner & CVPRW & Hybrid (ConvNeXt+FFT) & Full (paired) & {Fourier attention, color/structure consistency} \\
2024 & \cite{xiao2024homoformer} & HomoFormer & CVPR & Local Transformer & Full (paired+mask) & {Homogenization, local self-attention} \\
\hline

\multicolumn{7}{c}{\textbf{Supervised - Diffusion-based}} \\ 
\hline
2022 & \cite{jin2022shadowdiffusion} & ShadowDiffusion(J) & arXiv & Diffusion & Full (paired) & {Classifier-driven attention, chromaticity consistency} \\
2023 & \cite{guo2023shadowdiffusion} & ShadowDiffusion(G) & CVPR & Diffusion & Full (paired+mask) & {Degradation priors, auxiliary mask guidance} \\
2024 & \cite{jin2024des3} & DeS3 & AAAI & Diffusion & Full (paired) & {Adaptive attention, ViT similarity guidance} \\
2024 & \cite{liu2024recasting} & Recasting & AAAI & Diffusion (2-stage) & Full (paired+mask) & {Reflectance–illumination decomposition, bilateral correction} \\
2024 & \cite{mei2024latent} & LFG-Diffusion & WACV & Diffusion (latent) & Full (paired+mask) & {Latent prior learning, invariant loss} \\
2024 & \cite{luo2024diffshadow} & Diff-Shadow & arXiv & Diffusion (global-guided) & Full (paired+mask) & {Re-weight cross-attention, global sampling guidance} \\
2025 & \cite{xu2025detail} & DPLD & CVPR & Diffusion (latent + VAE DI) & Full (paired) &
{Latent stable diffusion + detail injection} \\
\hline

\multicolumn{7}{c}{\textbf{Unsupervised}} \\ 
\hline
2019 & \cite{hu2019mask} & Mask-ShadowGAN & ICCV & GAN & Unpaired & {Cycle-consistency, mask-guided generation} \\
2021 & \cite{Vasluianu_2021_CVPR} & PUL & CVPRW & GAN (loss-aug.) & Unpaired & {Mask, color, content, and style losses} \\
2021 & \cite{jin2021dc} & DC-ShadowNet & ICCV & CNN + domain classifier & Unpaired & {Chromaticity entropy, perceptual, boundary loss} \\
2021 & \cite{liu2021shadow_lightness} & LG-ShadowNet & TIP & CNN (lightness-guided) & Unpaired & {Lab-lightness dual-stream architecture} \\
2024 & \cite{zeng2024semantic} & SG-GAN+DBRM & arXiv & GAN + Diffusion & Unpaired & {CLIP-guided semantics, diffusion refinement} \\
\hline

\multicolumn{7}{c}{\textbf{Weakly Supervised}} \\ 
\hline
2020 & \cite{le2020shadow} & Param+M+D-Net & ECCV & Physics-guided CNN & Weak (shadow+mask) & {Physical constraints, patch mapping} \\
2021 & \cite{liu2021shadow} & G2R-ShadowNet & CVPR & GAN (gen.+rem.) & Weak (shadow+mask) & {Generation–removal co-training, illumination consistency} \\
2023 & \cite{guo2023boundary} & BCDiff & ICCV & Diffusion (conditional) & Weak (shadow+mask) & {Intrinsic reflectance, illumination consistency} \\
\hline

\multicolumn{7}{c}{\textbf{Self-Supervised (Single Image)}} \\ 
\hline
2023 & \cite{jiang2023learning} & Self-ShadowGAN & IJCV & GAN (relighting) & Self (single image+mask) & {Relighting coefficients, histogram and patch discriminators} \\
\hline
\end{tabular}}
\vspace{-3mm}
\end{table*}

\section{Shadow Removal}
\label{sec:removal}

Shadow removal aims to generate shadow-free images or video frames by recovering the colors under the shadows.
Besides general scenes, document and facial shadow removal are important specific applications.
This subsection presents a comprehensive overview of deep models on shadow removal and summarizes commonly-used datasets and metrics for evaluating shadow removal methods. Further, to assess the effectiveness of various methods, we conduct experiments and present comparative results.

\subsection{Deep Models for Image Shadow Removal}
\label{sec:general_shadow_removal}

Table~\ref{tab:image_shadow_removal} summarizes the surveyed papers on image shadow removal. We categorize the methods by supervision levels. \xwhu{Across categories, two recurring priors emerge: (i) \emph{attenuation/matte} estimation that aligns illumination across shadow/non-shadow regions, and (ii) \emph{intrinsic-like} reflectance and shading separation, both also power detection and generation, explaining why several methods are cross-listed across tasks.}

\vspace{2.5mm}
\hspace{-6.5mm} \textbf{Supervised Learning.} Here, the supervision is usually based on either (i) the shadow-free images or (ii) the shadow-free images and shadow masks. \xwhu{Supervised pipelines tend to trade scalability for accuracy, but provide the clearest testbed to compare architectures and losses.}

\vspace{1.5mm}
\hspace{-6.5mm}
\textit{(i) CNN-based} methods: \xwhu{These approaches rely on locality-aware encoders/decoders plus boundary handling; many explicitly incorporate mask/matte priors or physics-guided constraints to stabilize color restoration near penumbrae.}

\begin{itemize}[leftmargin=*]
\item \textbf{CNN-CRF} \cite{khan2016automatic} utilizes multiple CNNs to learn shadow detection and builds a Bayesian model to eliminate image shadows. The deep networks are employed solely for shadow detection. \xwhu{This early pipeline foreshadows today’s mask-conditioned removal.}

\item \textbf{DeshadowNet} \cite{qu2017deshadownet} is an end-to-end network with three subnetworks to extract features from a global view of images. \xwhu{It establishes the encoder–decoder template later reused by many works.}

\item \textbf{SP+M-Net} \cite{le2019shadow} models the shadow image as a combination of a shadow-free image, shadow parameters, and a shadow matte, and then predicts the shadow parameters and shadow matte using two separate deep networks. In testing, it uses the shadow mask predicted from~\cite{zhu2018bidirectional} as an additional input.  

\item \textbf{DSC} \cite{hu2020direction} introduces a direction-aware spatial context (DSC) module to analyze image context with directional awareness. A CNN with multiple DSC modules~\cite{Hu_2018_CVPR} generates residuals that are combined with the inputs to produce shadow-free images. \xwhu{Notably cross-listed with detection, showing module reuse across tasks.}

\item \textbf{DHAN+DA} \cite{cun2020towards} presents the hierarchical aggregation attention model with multi-contexts and the attention loss from shadow masks, and synthesizes shadow images from various shadow masks and shadow-free images using the network of \textbf{Shadow Matting GAN}. 

\item \textbf{SP+M+I-Net} \cite{le2021physics}  extends \cite{le2019shadow} by constraining SP-Net and M-Net's search spaces, adding a penumbra reconstruction loss to help M-Net attend to shadow penumbra regions, utilizing I-Net for inpainting, and introducing a smoothness loss to regulate the matte layer. It can be extended for patch-based weakly-supervised shadow removal~\cite{le2020shadow}. 

\item \textbf{Auto} \cite{fu2021auto} matches shadow regions with non-shadowed areas in color to generate overexposed images, which are merged with the input via a shadow-aware FusionNet to produce an adaptive kernel weight map. Last, a boundary-aware RefineNet reduces remaining penumbra effects along shadow boundaries. 

\item \textbf{CANet} \cite{chen2021canet} uses a two-stage context-aware approach: first adopts a contextual patch matching module to find potential shadow and non-shadow patch pairs, facilitating information transfer from non-shadow to shadow areas across different scales, and employs an encoder-decoder to refine and finalize. 

\item  \textbf{EMDNet} \cite{zhu2022efficient} proposes a model-driven network for shadow removal iterative optimization. Each stage updates the transformation map
and shadow-free image. 

\item  \textbf{BMNet} \cite{zhu2022bijective}  is a bijective mapping network that integrates shadow removal and shadow generation sharing parameters. It features invertible blocks for affine transformations and includes a shadow-invariant color guidance module that leverages U-Net-derived shadow-invariant colors for color restoration. 

\item \textbf{G2C-DeshadowNet} \cite{gao2022towards} is a two-stage shadow removal framework that first removes shadows from grayscale images and colorizes them utilizing modified self-attention blocks to optimize global image information. 

\item \textbf{SG-ShadowNet} \cite{wan2022style} is a two-part style-guided shadow removal network: a U-Net-based coarse deshadow network for initial shadow processing and a style-guided re-deshadow network for refining outcomes, employing a spatially region-aware prototypical normalization layer to render the non-shadow region style to the shadow region. 

\item \textbf{MStructNet}~\cite{liu2023structure} reconstructs the structural information of input images to remove shadows, harnessing a shadow-free structural prior for image-level shadow eradication and engaging multi-level structural insights.  

\item \textbf{DNSR} \cite{Vasluianu_2023_CVPR} is a U-Net-based architecture, featuring dynamic convolution, exposure adjustment, and a distillation phase to enhance feature maps. It integrates channel attention and fused pooling for improved feature blending. 

\item \textbf{PES} \cite{Cui_2023_CVPR} uses pyramid inputs for various shadow sizes and shapes, with NAFNet~\cite{Chu_2022_CVPR} as the base framework. A three-stage training process with varying input and crop sizes, loss functions, batch sizes, and iteration numbers, refined with a model soup~\cite{wortsman2022model}, achieved the highest PSNR in the NTIRE 2023 Image Shadow Removal Challenge on the WSRD. 

\item \textbf{Inpaint4shadow} \cite{li2023leveraging} reduces shadow remnants by pretraining on inpainting datasets, utilizing dual encoders for shadow and shadow-masked images, a weighted fusion module to merge features, and a decoder to generate shadow-free images. 

\item \textbf{LRA\&LDRA} \cite{yucel2023lra} improves shadow detection and removal by optimizing residuals in a stacked framework~\cite{wang2018stacked}. It reconstructs shadow regions using blending and color correction. It demonstrates that pre-training on a large-scale synthetic dataset containing paired shadow images, shadow-free images, and shadow masks significantly enhances performance.

\item \textbf{SHARDS} \cite{sen2023shards} removes shadows from high-resolution images using two networks: LSRNet generates a low-resolution shadow-free image from the shadow image and its mask, while DRNet refines details using the original high-resolution shadow image. This design keeps DRNet lightweight, as LSRNet handles the main shadow removal at a lower resolution. 

\item \textbf{PRNet} \cite{wang2024progressive} combines shadow feature extraction via a shallow six-block ResNet with progressive shadow removal through re-integration modules and ConvGRU-based updates~\cite{cho2014learning}. The re-integration module iteratively enhances outputs, and the update module generates shadow-attenuated features for prediction.

\end{itemize}

\vspace{1.5mm}
\hspace{-6.5mm}
\textit{(ii) GAN-based} methods adopt the generator to predict shadow-free images and the discriminator for judgement. \xwhu{Adversarial supervision encourages photorealistic relighting and is often paired with mask/matte estimation to reduce color casts.}

\begin{itemize}[leftmargin=*]
\item \textbf{ST-CGAN} \cite{wang2018stacked} uses one conditional GAN to detect shadows and leverages another conditional GAN to remove shadows by using the shadow image and shadow mask as the inputs. 

\item \textbf{AngularGAN} \cite{sidorov2019conditional} uses a GAN to predict shadow-free images end-to-end. The network is trained on synthetic paired data. \xwhu{Synthetic pretraining mitigates data scarcity.}

\item \textbf{ARGAN}~\cite{ding2019argan} first develops a shadow attention detector to generate an attention map to mark the shadows and then recurrently recovers a shadow-lighter or shadow-free image. Note that it can be trained in a \textbf{semi-supervised} manner using unlabeled data with the adversarial loss. \xwhu{Attention ties detection cues to removal.}

\item  \textbf{RIS-GAN} \cite{zhang2020ris} adopts four generators in the encoder-decoder structure and three discriminators to generate negative residual images, intermediate shadow-removal images, inverse illumination maps, and refined shadow-removal images. 

\item  \textbf{TBRNet} \cite{liu2023shadow} is a three-branch network with multitask cooperation. It consists of three specialized branches: shadow image reconstruction to preserve input image details; shadow matte estimation to identify shadow locations and adjusts illumination; and shadow removal to align the lighting of shadow areas with non-shadow ones to produce a shadow-free image.  

\end{itemize}

\vspace{1.5mm}
\hspace{-6.5mm}
\textit{(iii) Transformer-based} methods better capture global contextual information by the self-attention mechanism. \xwhu{Transformers propagate long-range illumination cues but often require mask-aware inductive biases to sharpen boundaries.}

\begin{itemize}[leftmargin=*]
\item \textbf{CRFormer} \cite{wan2022crformer} is a hybrid CNN-Transformer framework, with asymmetrical CNNs to extract features from shadow and non-shadow areas, a region-aware cross-attention mechanism to aggregate shadow region features, and a U-shaped network to refine the results. 

\item \textbf{CNSNet} \cite{yu2022cnsnet}  uses a dual approach for shadow removal, integrating shadow-oriented adaptive normalization for statistical consistency between shadow and non-shadow areas, and shadow-aware aggregation with Transformer to connect pixels across shadow and non-shadow areas. 

\item  \textbf{ShadowFormer}~\cite{guo2023shadowformer} uses a channel attention encoder-decoder framework with a shadow-interaction attention mechanism, analyzing correlations between shadow and non-shadow patches using contextual information. 

\item \textbf{SpA-Former}~\cite{10191081} consists of the Transformer layers, a series of joint Fourier transform residual blocks~\cite{mao2021deep}, and two-wheel joint spatial attentions. The two-wheel joint spatial attention is same as DSC~\cite{Hu_2018_CVPR,hu2020direction} but trained with shadow masks. 

\item \textbf{TSRFormer} \cite{Chang_2023_CVPR} is a two-stage architecture, employing distinct Transformer models for global shadow removal and content refinement, suppressing the residual shadow and refining the content information. 
SpA-Former~\cite{10191081} and ShadowFormer~\cite{guo2023shadowformer} serve as their backbones.

\item \textbf{ShadowMaskFormer} \cite{li2024shadowmaskformer} integrates the Transformer with a shadow mask in patch embedding. It uses 0/1 and -1/+1 binarization to amplify the pixels in shadow regions. 

\item  \textbf{ShadowRefiner} \cite{dong2024shadowrefiner} employs a ConvNeXt-based U-Net for extracting spatial and frequency representations to map shadow-affected to shadow-free images, and features a fast Fourier attention Transformer for color and structure consistency.

\item \textbf{HomoFormer} \cite{xiao2024homoformer} is a local window-based Transformer for shadow removal that homogenizes shadow degradation. It uses random shuffle operations and their inverse to rearrange pixels, allowing the local self-attention layer to process shadows effectively and eliminate inductive bias~\cite{hu2025demystify}. A new feed-forward network with depth-wise convolution enhances position modeling and exploits image structures.

\end{itemize}

\vspace{1.5mm}
\hspace{-6.5mm}
\textit{(iv) Diffusion-based} methods help produce even more visually-pleasant results. \xwhu{They inject strong generative priors that improve realism and color consistency, especially under complex, soft-shadow illumination.}

\begin{itemize}[leftmargin=*]
\item \textbf{ShadowDiffusion(J)} \cite{jin2022shadowdiffusion} uses classifier-driven attention for shadow detection, structure preservation loss with DINO-ViT features for reconstructions, and chromaticity consistency loss to ensure uniform colors in areas without shadows. 

\item \textbf{ShadowDiffusion(G)} \cite{guo2023shadowdiffusion} incrementally refines the output through degradation and diffusive generative priors, and enhances the accuracy of shadow mask estimation as an auxiliary aspect of the diffusion generator. 

\item \textbf{DeS3} \cite{jin2024des3} removes hard, soft, and self shadows using adaptive attention and ViT similarity mechanisms. It employs DDIM~\cite{song2020denoising} as the generative model and utilizes adaptive classifier-driven attention to emphasize shadow regions, with the DINO-ViT loss acting as the stopping criterion during inference. 

\item \textbf{Recasting} \cite{liu2024recasting} has two stages: a shadow-aware decomposition network separates reflectance and illumination using self-supervised regularizations, and a bilateral correction network adjusts lighting in shadow areas with a local lighting correction module. It then progressively restores degraded texture details with an illumination-guided texture restoration module. 

\item  \textbf{LFG-Diffusion} \cite{mei2024latent} trains a diffusion network on shadow-free images to learn shadow-free priors in a latent feature space. It then uses these pretrained weights for efficient shadow removal, minimizing the invariant loss between encoded shadow-free and shadow images with masks, while enhancing interactions between latent noise variables and the diffusion network. 

\item  \textbf{Diff-Shadow} \cite{luo2024diffshadow} is a global-guided diffusion model with parallel U-Nets: a local branch for patch noise estimation and a global branch for shadow-free image recovery. It uses the re-weight cross attention and global-guided sampling to explore global context from non-shadow regions and to determine fusion weights for patch noise, preserving illumination consistency. 

\item \textbf{DPLD}~\cite{xu2025detail} employs a two-stage Stable Diffusion adaptation, where the latent fine-tuning for mask-free shadow removal and a detail-injection module that restores high-frequency, shadow-free textures, improving generalization across datasets.

\end{itemize}

\vspace{2.5mm}
\hspace{-6.5mm} \textbf{Unsupervised Learning.} This category of methods trains the deep network without using paired shadow and shadow-free images, which are difficult to obtain. \xwhu{The core idea is to replace ground truth with cycle/contrastive constraints, physics-based regularizers, or generative priors.}

\begin{itemize}[leftmargin=*]
\item \textbf{Mask-ShadowGAN} \cite{hu2019mask} is the first unsupervised shadow removal method, which automatically learns to produce a shadow mask from the input shadow image and takes the mask to guide the shadow generation via re-formulated cycle-consistency constraints. It simultaneously learns to produce shadow masks and remove shadows. 

\item \textbf{PUL} \cite{Vasluianu_2021_CVPR} improves Mask-ShadowGAN with four additional losses: mask loss (L1 difference between sampled and generated masks), color loss (MSE between smoothed images), content loss (feature loss from VGG-16), and style loss (Gram matrix of VGG-16 features). 

\item \textbf{DC-ShadowNet} \cite{jin2021dc} handles shadow regions using a shadow/shadow-free domain classifier. It is trained with a physics-based shadow-free chromaticity loss from entropy minimization in log-chromaticity space, a shadow-robust perceptual features loss with pre-trained VGG-16, a boundary smoothness loss, and some additional losses like Mask-ShadowGAN. 

\item \textbf{LG-ShadowNet} \cite{liu2021shadow_lightness} improves Mask-ShadowGAN using a lightness-guided network. In Lab color space, a CNN first adjusts lightness in the L channel, then another CNN uses these features for shadow removal in all Lab channels. Multi-layer connections blend lightness and shadow removal features in a dual-stream architecture. 

\item \textbf{SG-GAN+DBRM} \cite{zeng2024semantic} has two networks. (i) SG-GAN, based on Mask-ShadowGAN~\cite{hu2019mask}, produces coarse shadow removal results and synthetic paired data, guided by a multi-modal semantic prompter using CLIP~\cite{radford2021learning} for text-based semantics. (ii) DBRM, a diffusion model, refines the coarse results and this model is trained on real shadow-free images and shadow-removed images, with shadows in the before-removal images synthesized by Mask-ShadowGAN. 

\item \textbf{LG-ShadowNet} \cite{liu2021shadow_lightness} improves Mask-ShadowGAN using a lightness-guided network. In Lab color space, a CNN first adjusts lightness in the L channel, then another CNN uses these features for shadow removal in all Lab channels. Multi-layer connections blend lightness and shadow removal features in a dual-stream architecture. 

\item \textbf{SG-GAN+DBRM} \cite{zeng2024semantic} has two networks. (i) SG-GAN, based on Mask-ShadowGAN~\cite{hu2019mask}, produces coarse shadow removal results and synthetic paired data, guided by a multi-modal semantic prompter using CLIP~\cite{radford2021learning} for text-based semantics. (ii) DBRM, a diffusion model, refines the coarse results and this model is trained on real shadow-free images and shadow-removed images, with shadows in the before-removal images synthesized by Mask-ShadowGAN. 

\item \textbf{LG-ShadowNet} \cite{liu2021shadow_lightness} improves Mask-ShadowGAN using a lightness-guided network. In Lab color space, a CNN first adjusts lightness in the L channel, then another CNN uses these features for shadow removal in all Lab channels. Multi-layer connections blend lightness and shadow removal features in a dual-stream architecture. 

\item \textbf{SG-GAN+DBRM} \cite{zeng2024semantic} has two networks. (i) SG-GAN, based on Mask-ShadowGAN~\cite{hu2019mask}, produces coarse shadow removal results and synthetic paired data, guided by a multi-modal semantic prompter using CLIP~\cite{radford2021learning} for text-based semantics. (ii) DBRM, a diffusion model, refines the coarse results and this model is trained on real shadow-free images and shadow-removed images, with shadows in the before-removal images synthesized by Mask-ShadowGAN. 

\end{itemize}

\vspace{2.5mm}
\hspace{-6.3mm} \textbf{Weakly Supervised Learning.} It trains the deep network only using the shadow images and shadow masks. The shadow masks can be predicted by the shadow detection methods. \xwhu{This regime operationalizes detector outputs as supervision, directly coupling detection and removal.}

\begin{itemize}[leftmargin=*]
\item \textbf{Param+M+D-Net} \cite{le2020shadow} trains on shadow images using shadow segmentation masks as supervision. It divides images into patches, learns mappings from shadow-boundary patches to non-shadow patches, and applies constraints based on a physical shadow formation model. 

\item \textbf{G2R-ShadowNet} \cite{liu2021shadow} has three sub-networks: generating, removing, and refining shadows. The shadow-generation network creates pseudo shadows in non-shadow areas, forming training pairs with non-shadow regions for the shadow-removal network. The refinement phase ensures color and illumination consistency. Shadow masks guide the entire process. \xwhu{Unifies generation and removal under mask guidance.}

\item \textbf{BCDiff} \cite{guo2023boundary} is a boundary-aware conditional diffusion model. It enhances an unconditional diffusion model by iteratively maintaining reflectance, supported by a shadow-invariant intrinsic decomposition model, to preserve structures within shadow regions. It also applies an illumination consistency constraint for uniform lighting. The base network used is Uformer~\cite{wang2022uformer}. 
\end{itemize}

\vspace{2.5mm}
\hspace{-6.3mm} \textbf{Self-Supervised Learning on a Single Image.} This task learns to remove shadows from an image by training on the image itself during testing, eliminating the need for training data. However, shadow masks are required. \xwhu{These instance-adaptive methods fit illumination statistics per image, trading speed for highly personalized correction.}

\begin{itemize}[leftmargin=*]
\item \textbf{Self-ShadowGAN} \cite{jiang2023learning} employs a shadow relighting network as the generator for shadow removal, supported by two discriminators. The relighting network uses lightweight MLPs to predict pixel-specific shadow relighting coefficients based on a physical model, with parameters determined by a fast convolutional network. It also includes a histogram-based discriminator that uses histograms from shadow-free areas as reference for restoring illumination in shadow areas, and a patch-based discriminator for improving texture quality in deshadowed regions. 
\end{itemize}

\vspace{2.5mm}
\hspace{-6.5mm} \textbf{Trends and Insights.}
(i) \emph{Mask and attenuation priors} remain the most reliable mechanisms for improving boundary fidelity and preventing over-brightening.
(ii) Cross-task conditioning (e.g., ST-CGAN, BMNet, G2R) using detection masks or generation priors yields more \emph{physically coherent relighting}, confirming shadow decomposition as a key intermediate representation.
(iii) Global illumination reasoning benefits from transformer and diffusion models, but these architectures still require \emph{shadow-aware tokenization} or frequency cues to avoid color drift.
(iv) Unpaired and self-supervised regimes narrow the performance gap by incorporating \emph{intrinsic decomposition and reflectance–shading constraints}, which are crucial for domain scalability.

\begin{table*}[tp]
\caption{Deep models for document and facial shadow removal.}
\vspace{-1mm}
\label{tab:document_facial_shadow_removal}
\centering
\resizebox{\textwidth}{!}{%
\begin{tabular}{c|c|c|c|c|c|c|c}
\hline
\xwhu{Application Type} & \xwhu{Year} & \xwhu{Refs.} & \xwhu{Method} & \xwhu{Publication} & \xwhu{Architecture Type} & \xwhu{Supervision} & \xwhu{Key Innovation / Contribution} \\
\hline

\multicolumn{8}{c}{\textbf{Document Shadow Removal}} \\
\hline
Document & 2020 & \cite{lin2020bedsr} & BEDSR-Net & CVPR & CNN (BE-Net + SR-Net) & Full & \xwhu{Background estimation with attention} \\
Document & 2023 & \cite{zhang2023document} & BGShadowNet & CVPR & Two-stage CNN & Full & \xwhu{Background-aware correction and refinement} \\
Document & 2023 & \cite{li2023high} & FSENet & ICCV & Transformer + CNN & Full & \xwhu{Frequency-based illumination restoration} \\
\hline

\multicolumn{8}{c}{\textbf{Facial Shadow Removal}} \\
\hline
Facial & 2020 & \cite{DBLP:journals/tog/ZhangBTPZNJ20} & Zhang et al. & SIGGRAPH & Dual-branch CNN & Full & \xwhu{External vs.\ self-shadow decoupling} \\
Facial & 2021 & \cite{he2021unsupervised} & He et al. & ACM MM & GAN + Decomposition & Unsupervised & \xwhu{Unsupervised identity–light disentanglement} \\
Facial & 2022 & \cite{liu2022blind} & GS+C & BMVC & Two-stage pipeline & Full & \xwhu{Grayscale deshadowing with temporal consistency} \\
Facial (Eyeglasses) & 2022 & \cite{lyu2022portrait} & Lyu et al. & CVPR & Segmentation + Restoration & Full & \xwhu{Joint shadow / eyeglass removal} \\
Facial & 2023 & \cite{zhang2023facial} & GraphFFNet & CGF (PG) & Graph + Symmetry fusion & Full & \xwhu{Graph-based global–local fusion} \\
\hline
\end{tabular}}
\vspace{-4mm}
\end{table*}

\subsubsection{Document Shadow Removal}
\label{sec:document_shadow}

Removing shadows in documents improves the visual quality and readability of digital copies. General shadow removal methods face challenges in handling documents, due to the need for a large paired dataset and the lack of considering specific document image properties.
Table~\ref{tab:document_facial_shadow_removal} summarizes deep models for this task.

\begin{itemize}[leftmargin=*]
  \item \textbf{BEDSR-Net}~\cite{lin2020bedsr}. It is the first deep network designed for document image shadow removal. It consists of two sub-networks: BE-Net estimates the global background color and generates an attention map. These, along with the input shadow image, are used by SR-Net to produce the shadow-free image.

  \item \textbf{BGShadowNet}~\cite{zhang2023document}. It leverages backgrounds from a color-aware background extraction network for shadow removal in a two-stage process. First, it fuses background and image features to generate realistic initial results. Second, it corrects illumination and color inconsistencies using a background-based attention module and enhances low-level details with a detail enhancement module, inspired by image histogram equalization.

  \item \textbf{FSENet}~\cite{li2023high}. It aims for high-resolution document shadow removal by first splitting images into low- and high-frequency components. The low-frequency part uses a Transformer for illumination adjustments, while the high-frequency part uses cascaded aggregations and dilated convolutions to enhance pixels and recover textures.
\end{itemize}

\vspace{2.5mm}
\hspace{-6.5mm} \textbf{Trends and Insights.}
(i) Methods increasingly adopt \emph{background-aware and layout-aware} conditioning to preserve page color uniformity and text integrity.
(ii) Hybrid frequency and global-context designs help suppress illumination artifacts while maintaining crisp edges.
(iii) Future progress depends on \emph{OCR-consistency or layout priors} to enable weak and self-supervised training without requiring clean paired scans.


\subsubsection{Facial Shadow Removal}
\label{sec:subsub_facial_shadow}

Facial shadow removal involves eliminating external shadows, softening facial shadows, and balancing lighting. Table~\ref{tab:document_facial_shadow_removal} summarizes the deep models. This topic is also related to face relighting~\cite{hou2021towards}, as accurate shadow manipulation is crucial for photo-realistic results. Additionally, removing shadows improves the robustness of facial landmark detection~\cite{fu2021benchmarking}.

\begin{itemize}[leftmargin=*]
  \item \textbf{Zhang et al.}~\cite{DBLP:journals/tog/ZhangBTPZNJ20}. They present the first deep-learning-based method tailored for facial image shadow removal. It uses two separate deep models: one for removing foreign shadows cast by external objects and another for softening the facial shadows. Both models are based on the modified GridNet~\cite{fourure2017residual,niklaus2018context}.

  \item \textbf{He et al.}~\cite{he2021unsupervised}. They present the first unsupervised facial shadow removal method by framing it as an image decomposition task. It processes a single shadowed portrait to produce a shadow-free image, a full-shadow image, and a shadow mask, using the pretrained face generators like StyleGAN2 and the face segmentation masks.

  \item \textbf{GS+C}~\cite{liu2022blind}. It removes shadows by splitting it into grayscale processing and colorization. Shadows are identified and removed in grayscale, then colors are restored through inpainting. To maintain consistency across video frames, it includes a temporal sharing module that addresses pose and expression variations.

  \item \textbf{Lyu et al.}~\cite{lyu2022portrait}. They present a two-stage model to remove eyeglasses together with their shadows. The first stage predicts masks using a cross-domain segmentation module, while the second stage uses these masks to guide a deshadow and deglass network. The model is trained on synthetic data and uses a domain adaptation network for real images.

  \item \textbf{GraphFFNet}~\cite{zhang2023facial}. It is a graph-based feature fusion network for removing shadows from facial images. It employs a multi-scale encoder to extract local features, an image flipper to leverage facial symmetry for a coarse shadow-less image, and a graph-based convolution encoder to identify global relationships. A feature modulation module combines these global and local features, and a fusion decoder generates the shadow-free image.

\end{itemize}

\vspace{2.5mm}
\hspace{-6.5mm} \textbf{Trends and Insights.} 
(i) Facial deshadowing increasingly relies on \emph{illumination–identity disentanglement}, leveraging symmetry, intrinsic decomposition, or generative priors to preserve identity.
(ii) Video settings highlight the need for \emph{temporal coherence and semantic stability}, which raw pixel models often struggle to maintain.
(iii) Unified relighting-deshadowing frameworks and diffusion-based priors offer a promising path toward high-fidelity, identity-preserving facial illumination correction.

\subsection{Deep Models for Video Shadow Removal}

\begin{itemize}[leftmargin=*]
\item \textbf{PSTNet} \cite{chen2024learning} is a video shadow removal method, combining physical, spatial, and temporal features, supervised by shadow-free images and masks. It uses a physical branch for adaptive exposure and supervised attention, and spatial and temporal branches for resolution and coherence. A feature fusion module refines outputs, and an S2R strategy adapts the synthetically trained model for real-world use without retraining.


\item \textbf{GS+C} \cite{liu2022blind} performs facial shadow removal in videos. See Section~\ref{sec:subsub_facial_shadow} for details.

\end{itemize}

\subsection{Shadow Removal Datasets}
\label{sec:shadow_removal_data}

\subsubsection{General Image Shadow Removal Datasets}
\begin{itemize}[leftmargin=*] 

\item  \textbf{SRD}
~\cite{qu2017deshadownet} is the first large-scale shadow removal dataset with 3,088 shadow and shadow-free image pairs.
The dataset's diversity spans four dimensions: illumination (hard and soft shadows), a wide range of scenes (parks to beaches), varying reflectance by casting shadows on different objects, and diverse silhouettes and penumbra widths using occluders of different shapes.
The shadow masks of SRD are newly labeled by~\cite{liu2024recasting}.

\item  \textbf{ISTD}~\cite{wang2018stacked} \& \textbf{ISTD+}~\cite{le2019shadow}:
Both consist of shadow images, shadow-free images, and shadow masks, with 1,330 training images and 540 testing images from 135 unique background scenes. ISTD suffers from color and luminosity inconsistencies between shadow and shadow-free images~\cite{hu2020direction, le2019shadow}, which ISTD+ corrects with a color compensation mechanism to ensure uniform pixel colors across the ground-truth images.

\item  \textbf{GTAV}
~\cite{sidorov2019conditional} is a synthetic dataset of 5,723 shadow and shadow-free image pairs. The scenes are rendered from the video game GTAV by Rockstar, depicting real-world-like scenes in two editions: with and without shadows. It includes 5,110 standard daylight scenes and an additional 613 indoor and night scenes.

\item  \textbf{USR}
~\cite{hu2019mask} is designed for unpaired shadow removal tasks, containing 2,511 images featuring shadows and 1,772 images without shadows. This dataset encompasses a wide array of scenes, showcasing shadows cast by a diverse range of objects. It spans over a thousand unique scenes, offering a substantial variety for research in shadow removal technologies.

\item \textbf{SFHQ} \cite{sen2023shards}, Shadow Food-HQ, consists of 14,520 high-resolution food images (12MP) with annotated shadow masks. It includes diverse scenes under various lighting and perspectives, divided into 14,000 training and 520 testing triplets.

\item \textbf{WSRD} 
~\cite{Vasluianu_2023_CVPR} was created in a controlled indoor setting with directional and diffuse lighting. It features 1,200 high-resolution (1920x1440) image pairs: 1,000 for training, 100 for validation, and 100 for testing. The dataset includes surfaces of various colors, textures, and geometries, and objects of different thicknesses, heights, depths, and materials, including opaque, translucent, and transparent types. It was used by 19 teams in the NTIRE23 challenge for image shadow removal \cite{Vasluianu_2023_CVPR_report}.

\end{itemize}

\subsubsection{General Video Shadow Removal Datasets}
\begin{itemize}[leftmargin=*] 

\item \textbf{SBU-Timelapse}~\cite{le2021physics} is a video shadow removal dataset with 50 videos of static scenes, featuring only shifting shadows and no-moving objects. A pseudo shadow-free frame is derived from each video using the ``max-min'' technique.


\item 
\textbf{SVSRD-85} \cite{chen2024learning}  is a synthetic video shadow removal dataset from GTAV, containing 85 videos with 4,250 frames, collected by toggling the shadow renderer. It covers various object categories and motion/illumination conditions, with each frame paired with shadow-free images.

\end{itemize}

\subsubsection{Document Shadow Removal Datasets}

\begin{itemize}[leftmargin=*]

\item \textbf{SDSRD} \cite{lin2020bedsr} is a synthetic dataset created with Blender, containing 970 document images and 8,309 synthesized shadow images under different lighting and occluders. It includes 7,533 training triplets and 776 testing triplets.


\item \textbf{RDSRD} \cite{lin2020bedsr} is a real dataset captured by cameras. The dataset comprises 540 images featuring 25 documents with shadow images, shadow-free images, and shadow masks. This dataset is used only for evaluation.

\item \textbf{RDD} \cite{zhang2023document} uses document backgrounds such as papers, books, and pamphlets. It consists of 4,916 image pairs, each captured with and without shadows by positioning and then removing an occluder. 4,371 pairs are for training and 545 for testing.

\item \textbf{SD7K} \cite{li2023high} contains 7,620 pairs of high-resolution real-world document images with and without shadows, along with annotated shadow masks. It includes various document types (manga, papers, figures), 30+ occluders, and 350+ documents captured under three lighting conditions (cool, warm, and sunlight).

\end{itemize}

\subsubsection{Facial Shadow Removal Datasets}

\begin{itemize}[leftmargin=*] 

\item \textbf{UCB} \cite{DBLP:journals/tog/ZhangBTPZNJ20} comprises synthesized foreign and facial shadows. Foreign shadows are created by blending lit and shadowed images using shadow masks on a dataset of 5,000 faces without foreign shadows; however, eyeglass shadows are considered inherent. Facial shadows are generated from Light Stage~\cite{debevec2000acquiring} scans of 85 subjects across various expressions and poses, using the weighted one-light-at-a-time combinations method. 
 
\item \textbf{SFW} \cite{liu2022blind} is assembled for facial shadow removal in real-world conditions, consisting of 280 videos from 20 subjects, with most videos recorded in 1080p resolution. Labels are provided for various shadow masks, such as cast shadows, self-shadows, bright or saturated face regions, and eyeglasses, across 440 frames.




\item \textbf{PSE} \cite{lyu2022portrait}, Portrait Synthesis with Eyeglasses, is a synthetic dataset by 3D rendering. It simulates 3D eyeglasses on face scans using node-based registration, rendering them under various illuminations to produce four image types with masks. From 438 identities, 73 are chosen, each with 20 expression scans, paired with five eyeglass styles and four HDR lighting conditions, generating 29,200 training samples.

\end{itemize}

\begin{figure}[tp]
    \centering
    \begin{subfigure}[b]{0.9\linewidth}
        \centering
        \includegraphics[width=\linewidth]{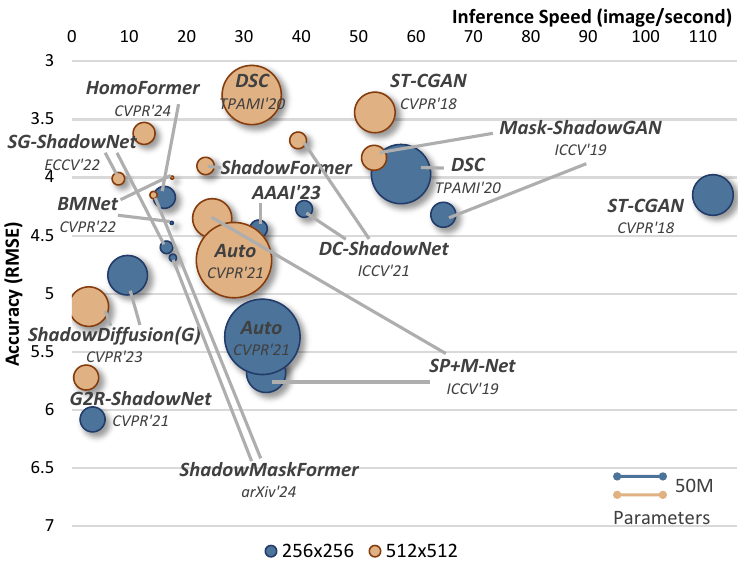}
        \vspace{-4mm}
    \end{subfigure}
    \\[2mm]
    \begin{subfigure}[b]{0.9\linewidth}
        \centering
        \includegraphics[width=\linewidth]{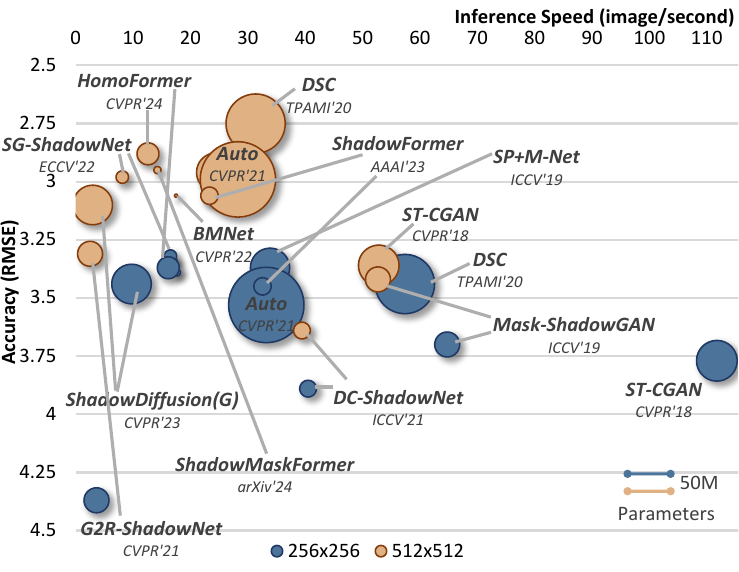}
        \vspace{-4mm}
    \end{subfigure}
    \vspace{-3mm}
    \caption{Shadow removal methods on the SRD (top) and ISTD+ (bottom) datasets: accuracy, parameters (indicated by the area of the bubbles), and speed.}
    \label{fig:shadow_removal}
    \vspace{-4mm}
\end{figure}

\begin{table*}[tp]
\centering
\caption{Comparing image shadow removal methods on an NVIDIA GeForce RTX 4090 GPU. LPIPS uses the VGG as the extractor. Please note that for the results shown in the rightmost columns, we report the cross-dataset generalization evaluation, where the models were trained on SRD and tested on DESOBA. Note that \textit{Mask-ShadowGAN} and \textit{DC-ShadowNet} are unsupervised methods, and \textit{G2R-ShadowNet} is a weakly-supervised method.}
\vspace{-1mm}
\label{table:SRD_ISTD_removal_evaluation}
\resizebox{1.0\linewidth}{!}{%
\setlength{\tabcolsep}{3pt}
\begin{tabular}{c|c|c|c|c|c|c|c|c|c|c|c||c|c}
\hline
\multirow{2}{*}{Input Size} & \multirow{2}{*}{Methods} & \multicolumn{4}{c|}{SRD} & \multicolumn{4}{c|}{ISTD+} & \multirow{2}{*}{Param.(M)} & \multirow{2}{*}{Infer.(images/s)} & \multicolumn{2}{c}{\textit{DESOBA (cross)}} \\ \cline{3-10} \cline{13-14} 
& & RMSE$\downarrow$ & PSNR$\uparrow$ & SSIM$\uparrow$ & LPIPS$\downarrow$ & RMSE$\downarrow$ & PSNR$\uparrow$ & SSIM$\uparrow$ & LPIPS$\downarrow$ & & & RMSE$\downarrow$ & PSNR$\uparrow$ \\ \hline
\multirow{13}{*}{256 $\times$ 256} 
& ST-CGAN \cite{wang2018stacked} & 4.15 & 25.08 & 0.637 & 0.443 & 3.77 & 25.74 & 0.691 & 0.408 & 58.49 & \textbf{111.79} & {7.07} & {20.23} \\
& SP+M-Net \cite{le2019shadow} & 5.68 & 22.25 & 0.636 & 0.444 & 3.37 & 26.58 & 0.717 & 0.373 & 54.42 & 33.88 & 5.10 & 23.35 \\
& Mask-ShadowGAN \cite{hu2019mask} & 4.32 & 24.67 & 0.662 & 0.427 & 3.70 & 25.50 & 0.720 & 0.377 & 22.76 & 64.77 & 6.94 & 20.47 \\
& DSC \cite{hu2020direction} & \textbf{3.97} & \textbf{25.46} & 0.678 & 0.412 & 3.44 & 26.53 & \textbf{0.738} & \textbf{0.347} & 122.49 & 57.40 & 6.66 & 20.71 \\
& Auto \cite{fu2021auto} & 5.37 & 23.20 & 0.694 & 0.370 & 3.53 & 26.10 & 0.718 & 0.365 & 196.76 & 33.23 & 5.88 & 22.62 \\
& G2R-ShadowNet \cite{liu2021shadow} & 6.08 & 21.72 & 0.619 & 0.460 & 4.37 & 24.23 & 0.696 & 0.396 & 22.76 & 3.62 & 5.13 & 23.14 \\
& DC-ShadowNet \cite{jin2021dc} & 4.27 & 24.72 & 0.670 & 0.383 & 3.89 & 25.18 & 0.693 & 0.406 & 10.59 & 40.51 & 6.88 & 20.58 \\
& BMNet \cite{zhu2022bijective} & 4.39 & 24.24 & 0.721 & 0.327 & 3.34 & 26.62 & 0.731 & 0.354 & \textbf{0.58} & 17.42 & 5.37 & 22.75 \\
& SG-ShadowNet \cite{wan2022style} & 4.60 & 24.10 & 0.636 & 0.443 & \textbf{3.32} & \textbf{26.80} & 0.717 & 0.369 & 6.17 & 16.51 & \textbf{4.92} & 23.36 \\
& ShadowDiffusion(G) \cite{guo2023shadowdiffusion} & 4.84 & 23.26 & 0.684 & 0.363 & 3.44 & 26.51 & 0.688 & 0.404 & 55.52 & 9.73 & 5.59 & 22.08 \\
& ShadowFormer \cite{guo2023shadowformer} & 4.44 & 24.28 & 0.715 & 0.348 & 3.45 & 26.55 & 0.728 & 0.350 & 11.37 & 32.57 & 5.01 & \textbf{23.49} \\
& ShadowMaskFormer \cite{li2024shadowmaskformer} & 4.69 & 23.85 & 0.671 & 0.386 & 3.39 & 26.57 & 0.698 & 0.395 & 2.28 & 17.63 & 5.82 & 22.14 \\
& HomoFormer \cite{xiao2024homoformer} & 4.17 & 24.64 & \textbf{0.723} & \textbf{0.325} & 3.37 & 26.72 & 0.732 & 0.348 & 17.81 & 16.14 & 5.02 & 23.41 \\ \hline
\multirow{13}{*}{512 $\times$ 512} 
& ST-CGAN \cite{wang2018stacked} & 3.44 & 26.95 & 0.786 & 0.282 & 3.36 & 27.32 & 0.829 & 0.252 & 58.49 & \textbf{52.84} & {6.65} & {20.98} \\
& SP+M-Net \cite{le2019shadow} & 4.35 & 24.89 & 0.792 & 0.269 & 2.96 & 28.31 & \textbf{0.866} & \textbf{0.183} & 54.42 & 24.48 & 4.57 & 24.80 \\
& Mask-ShadowGAN \cite{hu2019mask} & 3.83 & 25.98 & 0.803 & 0.270 & 3.42 & 26.51 & 0.865 & 0.196 & 22.76 & 52.70 & 6.74 & 20.96 \\
& DSC \cite{hu2020direction} & \textbf{3.29} & \textbf{27.39} & 0.802 & 0.263 & \textbf{2.75} & \textbf{28.85} & 0.861 & 0.196 & 122.49 & 31.37 & 5.58 & 22.61 \\
& Auto \cite{fu2021auto} & 4.71 & 24.32 & 0.800 & 0.247 & 2.99 & 28.07 & 0.853 & 0.189 & 196.76 & 28.28 & 5.05 & 24.16 \\
& G2R-ShadowNet \cite{liu2021shadow} & 5.72 & 22.44 & 0.765 & 0.302 & 3.31 & 27.13 & 0.841 & 0.221 & 22.76 & 2.50 & 4.60 & 24.56 \\
& DC-ShadowNet \cite{jin2021dc} & 3.68 & 26.47 & 0.808 & 0.255 & 3.64 & 26.06 & 0.835 & 0.234 & 10.59 & 39.45 & 6.62 & 21.25 \\
& BMNet \cite{zhu2022bijective} & 4.00 & 25.39 & 0.820 & 0.225 & 3.06 & 27.74 & 0.848 & 0.212 & \textbf{0.58} & 17.49 & 5.06 & 23.65 \\
& SG-ShadowNet \cite{wan2022style} & 4.01 & 25.56 & 0.786 & 0.279 & 2.98 & 28.25 & 0.849 & 0.205 & 6.17 & 8.12 & 4.47 & 24.53 \\
& ShadowDiffusion(G) \cite{guo2023shadowdiffusion} & 5.11 & 23.09 & 0.804 & 0.240 & 3.10 & 27.87 & 0.839 & 0.222 & 55.52 & 2.96 & 5.50 & 22.34 \\
& ShadowFormer \cite{guo2023shadowformer} & 3.90 & 25.60 & 0.819 & 0.228 & 3.06 & 28.07 & 0.847 & 0.204 & 11.37 & 23.32 & 4.55 & 24.81 \\
& ShadowMaskFormer \cite{li2024shadowmaskformer} & 4.15 & 25.13 & 0.798 & 0.249 & 2.95 & 28.34 & 0.849 & 0.211 & 2.28 & 14.25 & 5.51 & 23.11 \\
& HomoFormer \cite{xiao2024homoformer} & 3.62 & 26.21 & \textbf{0.827} & \textbf{0.219} & 2.88 & 28.53 & 0.857 & 0.196 & 17.81 & 12.60 & \textbf{4.42} & \textbf{24.89} \\ \hline
\end{tabular} }
\vspace{-2.5mm}
\end{table*}

\subsection{Evaluation Metrics}

\begin{itemize}[leftmargin=*] 

\item \textbf{RMSE}\footnote{Some previous works use code that mistakenly computes the MAE (mean absolute error). This paper corrects that issue.} \cite{guo2013paired} calculates the root-mean-square error in the LAB color space between the ground-truth shadow-free image and the recovered image, ensuring the local perceptual uniformity.

\item \textbf{LPIPS} \cite{zhang2018unreasonable} (Learned Perceptual Image Patch Similarity) assesses the perceptual distance between image patches, where a higher score indicates lower similarity and vice versa. This paper adopts VGG~\cite{simonyan2015very} as the feature extractor in LPIPS.
 
\end{itemize}

\hspace{-5mm} \textbf{SSIM}~\cite{wang2004image} and \textbf{PSNR} are sometimes used in evaluation.

\subsection{Experimental Results}

\subsubsection{General Image Shadow Removal}
%

\hspace{-1.5mm} \textbf{Overall Performance Benchmark Results.}
Two widely-used datasets, SRD~\cite{qu2017deshadownet} and ISTD+~\cite{le2019shadow}, are adopted to assess the performance of shadow removal methods.
Methods compared are listed in Table \ref{table:SRD_ISTD_removal_evaluation}, and we excluded those for which the code is not available. 
We re-trained the compared methods using their original code, setting the input sizes to $256 \times 256$ and $512 \times 512$ to report results at two resolutions. 
For DSC~\cite{hu2020direction}, we transferred the code from Caffe to PyTorch and used a ResNeXt101 backbone.
ShadowDiffusion(G)~\cite{guo2023shadowdiffusion} uses pretrained Uformer~\cite{wang2022uformer} weights for ISTD+ inference.
For methods requiring shadow masks as inputs, \emph{unlike several previous methods using predicted shadow masks during the training, we adopt well-labeled masks in both SRD and ISTD+.
Unlike certain previous methods that rely on ground-truth masks during inference (may lead to data leakage), we employ shadow masks generated by the SDDNet detector~\cite{cong2023sddnet}.}
The detector is trained on the SBU dataset at a $512 \times 512$ resolution, which shows superior generalization,  as shown in Table \ref{table:sbu_CUHK-Shadow_evaluation}.
The employed evaluation metrics include RMSE, PSNR, SSIM, and LPIPS. \emph{Results are resized to match the ground-truth resolution in evaluation for a fair comparison. Some papers that resize the ground truth are incorrect, as this distorts details and leads to biased, less accurate evaluations of image quality.}

Table \ref{table:SRD_ISTD_removal_evaluation} and Fig. \ref{fig:shadow_removal}  summarize the accuracy\footnote{Some results differ significantly from the original reports due to our use of the consistent input size, evaluation code, and safeguards against data leakage.}, runtime, and model complexity of each method. Key insights include: (i) early methods such as DSC and ST-CGAN outperform later approaches across several evaluation metrics; (ii) unsupervised methods demonstrate performance comparable to supervised ones on SRD and ISTD+, likely due to the similar background textures in the training and test sets, with Mask-ShadowGAN offering the best trade-off between effectiveness and efficiency; (iii) smaller models like BMNet (0.58M) provide competitive performance without significant increases in model size; and (iv) most methods show improved results at higher resolutions, such as $512 \times 512$.
\emph{See the visual comparisons in the appendix.} 

\vspace{2.5mm}
\hspace{-6.5mm} \textbf{Cross-Dataset Generalization Evaluation.}
To assess the generalization capability of shadow removal methods, we conduct cross-dataset evaluations using models trained on the SRD training set to detect shadows on the combination of DESOBA (see Sec.~\ref{subsec:generation_datasets}) training and testing sets. 
Both datasets contain outdoor scenes, but SRD lacks occluders casting shadows, while DESOBA presents more complex environments. This marks the first large-scale evaluation of generalization on such a challenging dataset.
Note that DESOBA only labels  \textit{cast shadows} and we set the self shadows on objects as ``don't care'' in evaluation. 
SSIM and LPIPS are excluded, as SSIM depends on image windows and LPIPS uses network activations, both conflicting with the ``don't care'' policy.

The two rightmost columns in Table~\ref{table:SRD_ISTD_removal_evaluation} show that models performing well on controlled datasets like SRD and ISTD+ struggle in the more complex environments of DESOBA.
This is because SRD mainly features cast shadows in simpler, localized scenes with softer shadows and no occluders, whereas DESOBA presents more intricate scenes with harder shadows and occlusions.
This highlights the need for diverse training data and more adaptable models capable of handling real-world shadow scenarios.

\vspace{2.5mm}
\hspace{-6.5mm} \textbf{Summary.} As demonstrated by the experimental results, \emph{how to develop a robust model and prepare a representative dataset that delivers high performance for image shadow removal in complex scenarios}, remains a challenging problem.

\begin{table}[tp]
\centering
\caption{Comparing document shadow removal methods on an NVIDIA GeForce RTX 4090 GPU. VGG is used in LPIPS.}
\vspace{-1mm}
\label{table:SD7K_SFW_removal_evaluation}
\resizebox{1.0\linewidth}{!}{%
\setlength{\tabcolsep}{2.5pt}
\begin{tabular}{c|c|c|c|c|c|c}
\hline
 {Methods} & RMSE$\downarrow$ & PSNR$\uparrow$ & SSIM$\uparrow$ & LPIPS$\downarrow$ & Param.(M) & Infer.(images/s) \\ \hline
 BEDSR-Net \cite{lin2020bedsr} & 3.13 & 28.480 & 0.912 & 0.171 & 32.21 & 10.41 \\  \hline
 FSENet \cite{li2023high} & \textbf{2.46} & \textbf{31.251} & \textbf{0.948} & \textbf{0.161} & \textbf{29.40} & \textbf{19.37} \\ \hline
\end{tabular} }
\vspace{-2.5mm}
\end{table}

\subsubsection{Document Shadow Removal}
The RDD \cite{zhang2023document}  dataset is used to train and evaluate the document shadow removal methods and the input size is $512 \times 512$. 
The results are shown in Table \ref{table:SD7K_SFW_removal_evaluation}, where we observe that FSENet significantly outperforms BEDSR-Net in both accuracy and efficiency, making it the superior method across all metrics.

%

	\begin{table*}[tp]
\centering
\caption{Deep models for image shadow generation.}
\vspace{-2mm}
\label{tab:shadow-generation}
\resizebox{\textwidth}{!}{%
\begin{tabular}{c|c|c|c|c|c|c|c}
\hline
\xwhu{Year} & \xwhu{Refs.} & \xwhu{Method} & \xwhu{Publication} & \xwhu{Architecture Type} & \xwhu{Supervision} & \xwhu{Application Domain} & \xwhu{Key Innovation / Contribution} \\
\hline

\multicolumn{8}{c}{\textbf{GAN-based Methods}} \\ 
\hline
2019 & \cite{zhang2019shadowgan} & ShadowGAN & CVM & GAN (Dual Discriminators) & Full & Composition & \xwhu{Adversarial learning for realistic shadow synthesis} \\
2020 & \cite{liu2020arshadowgan} & ARShadowGAN & CVPR & Attention-GAN & Full & AR Composition & \xwhu{Attention-guided shadow transfer in AR scenes} \\
2021 & \cite{sheng2021ssn} & SSN & CVPR & Interactive CNN & Full & Composition & \xwhu{User-controllable soft shadow generation} \\
2022 & \cite{hong2022shadow} & SGRNet & AAAI & Two-stage GAN & Full & Composition & \xwhu{Joint shadow mask and illumination prediction} \\
2023 & \cite{valencca2023shadow} & Valença et al. & SIGGRAPH Asia & GAN + Physics-guided & Full & Composition & \xwhu{Physics-based illumination–shadow consistency} \\
\hline

\multicolumn{8}{c}{\textbf{Geometry / Feature-aware Methods}} \\ 
\hline
2022 & \cite{sheng2022controllable} & SSG & ECCV & Geometry-aware CNN & Full & Composition & \xwhu{Pixel-height representation for directional control} \\
2023 & \cite{Sheng_2023_CVPR} & PixHt-Lab & CVPR & Neural Renderer + 3D Buffer & Full & Composition & \xwhu{3D-aware projection for realistic soft shadows} \\
2023 & \cite{meng2023automatic} & HAU-Net \& IFNet & TMM & Attention U-Net + Fusion Net & Full & Composition & {Hierarchical illumination inference and fusion} \\
2024 & \cite{tao2024shadow} & DMASNet & AAAI & Two-stage CNN & Full & Composition & \xwhu{Decomposed mask–fill shadow generation pipeline} \\
\hline

\multicolumn{8}{c}{\textbf{Diffusion-based and Hybrid Models}} \\ 
\hline
2024 & \cite{Liu2024SG} & SGDiffusion & CVPR & Diffusion + ControlNet & Full & Composition & \xwhu{Diffusion priors for intensity–shape realism} \\
\hline

\multicolumn{8}{c}{\textbf{Unified or Multi-task Shadow Generation / Editing Models}} \\
\hline
2025 & \cite{wang2025metashadow} & MetaShadow & CVPR & Unified CNN–Transformer Framework & Full & Detection / Removal / Generation & Object-centered tri-task framework \\
\hline

\multicolumn{8}{c}{\textbf{Sketch-based and Cross-task Generation}} \\ 
\hline
2020 & \cite{zheng2020learning} & Zheng et al. & CVPR & CNN + Latent 3D Decoder & Full & Sketch & \xwhu{Latent 3D reconstruction for artistic shadows} \\
2021 & \cite{zhang2021smartshadow} & SmartShadow & ICCV & CNN (Interactive Tools) & Full & Sketch & \xwhu{User-guided shadow refinement via tools} \\
2019 & \cite{hu2019mask} & Mask-ShadowGAN & ICCV & CycleGAN + Mask Generator & Unsupervised & Shadow Removal & \xwhu{Mask-guided cycle-consistent} \\
2020 & \cite{cun2020towards} & Shadow Matting GAN & AAAI & Conditional GAN & Full & Shadow Removal & \xwhu{Shadow matting for region-based synthesis} \\
2021 & \cite{liu2021shadow} & G2R-ShadowNet & CVPR & Multi-branch Network & Weak & Shadow Removal & \xwhu{Generative-to-removal cross-task supervision} \\
\hline
\end{tabular}}
\vspace{-3mm}
\end{table*}

\section{Shadow Generation}
\label{sec:generation}

Shadow generation serves three main purposes: (i) image composition, which involves generating cast shadows for objects in photos such that one can insert or reposition objects realistically; (ii) data augmentation, which creates cast shadows in images to produce photo-realistic samples for deep model training; and (iii) sketching, which focuses on generating shadows for hand-drawn sketches to accelerate the artistic process.  
\xwhu{Unlike shadow removal, shadow generation requires explicit modeling of illumination geometry and occlusion consistency to maintain visual plausibility.}

\subsection{Deep Models for Image Shadow Generation}

\subsubsection{Shadow Generation for Image Composition}

\begin{itemize}[leftmargin=*]

\item \textbf{ShadowGAN}~\cite{zhang2019shadowgan} uses a GAN framework with dual discriminators to generate realistic shadows for virtual objects in natural scenes, ensuring geometric alignment and illumination consistency.

\item \textbf{ARShadowGAN}~\cite{liu2020arshadowgan} adds shadows to virtual objects in augmented reality under single-light settings using an attention-based generator.  
\xwhu{It learns the correspondence between real and virtual shadows without explicit 3D geometry or light estimation, enabling lightweight deployment.}

\item \textbf{SSN}~\cite{sheng2021ssn} provides an interactive system for controllable soft-shadow creation using 2D object masks.  
\xwhu{Its dynamic light-map conditioning allows real-time manipulation of shadow softness and direction.}

\item \textbf{SSG}~\cite{sheng2022controllable} introduces pixel height as a differentiable geometry proxy for shadow direction and shape control.  
\xwhu{This bridges the gap between geometric projection and neural softness modeling.}

\item \textbf{SGRNet}~\cite{hong2022shadow} adopts a two-stage generator to produce both shadow masks and corresponding shadow regions, achieving high realism by parameterizing global lighting.  

\item \textbf{Liu \textit{et al.}}~\cite{liu2022shadow} employ multi-scale feature enhancement and multi-level fusion to refine mask accuracy and illumination prediction.  
\xwhu{This improves shape coherence and brightness realism in composited scenes.}

\item \textbf{PixHt-Lab}~\cite{Sheng_2023_CVPR} reconstructs pixel heights into 3D space and employs a neural renderer to synthesize shadows and reflections.  
\xwhu{This marks a transition from purely 2D generation to hybrid 3D-aware shadow modeling.}

\item \textbf{HAU-Net \& IFNet}~\cite{meng2023automatic} jointly infer global illumination and shadow fusion using a hierarchical attention U-Net and an illumination-aware blending network.  

\item \textbf{Valença \textit{et al.}}~\cite{valencca2023shadow} enhance photo compositing via a generator that estimates a shadow gain map and mask, followed by physics-guided post-processing using lighting and camera priors.  

\item \textbf{DMASNet}~\cite{tao2024shadow} performs two-stage shadow synthesis: mask generation (box + shape) followed by illumination-adaptive refinement.  
\xwhu{This modular design yields controllable shadow geometry and tone balance.}

\item \textbf{SGDiffusion}~\cite{Liu2024SG} leverages a diffusion model enriched with real shadow priors and ControlNet-based~\cite{Zhang_2023_ICCV} modulation.  
\xwhu{It achieves high-fidelity shape–intensity coupling through semantic conditioning, extending diffusion models to physical lighting tasks.}

\item \textbf{MetaShadow}~\cite{wang2025metashadow} introduces an object-centered unified framework that jointly performs shadow detection, removal, and generation within a single model. 
By integrating relational shadow–object reasoning with a shared illumination-aware representation, it achieves consistent geometry, attenuation, and relighting behavior across tasks, enabling controllable shadow manipulation.

\end{itemize}

\subsubsection{Shadow Generation for Shadow Removal}
See \textbf{Mask-ShadowGAN}~\cite{hu2019mask}, \textbf{Shadow Matting GAN}~\cite{cun2020towards}, and \textbf{G2R-ShadowNet}~\cite{liu2021shadow} in Section~\ref{sec:general_shadow_removal}.
\xwhu{These works use synthetic shadow generation to create pseudo pairs for deshadowing supervision, bridging shadow generation and shadow removal tasks.}

\subsubsection{Shadow Generation for Sketch}

\begin{itemize}[leftmargin=*]
\item \textbf{Zheng \textit{et al.}}~\cite{zheng2020learning} generate artistic shadows from sketches under specified light directions by mapping strokes into a latent 3D space.  
\xwhu{Their network jointly models geometry and style, allowing self-shadowing and rim-light simulation.}

\item \textbf{SmartShadow}~\cite{zhang2021smartshadow} assists artists through three tools, i.e., shadow brush, boundary brush, and global generator, using CNNs to infer global direction and local shadow maps from user-guided sketches.  
\xwhu{It emphasizes controllability and interactive generation.}
\end{itemize}

\subsection{Shadow Generation Datasets}
\label{subsec:generation_datasets}

\subsubsection{For Image Composition}
\begin{itemize}[leftmargin=*]
\item \textbf{Shadow-AR}~\cite{liu2020arshadowgan}: 3,000 synthesized quintuples containing paired synthetic and real-world shadows, occluder masks, and matting labels for AR scenarios.  
\item \textbf{DESOBA}~\cite{hong2022shadow}: derived from SOBA~\cite{wang2020instance}, providing 840 training and 160 testing images with shadow-object associations.  
\item \textbf{RdSOBA}~\cite{tao2024shadow}: Unity-based dataset with 30 scenes and 800 objects, totaling 114k images and 28k pairs.  
\item \textbf{DESOBAv2}~\cite{Liu2024SG}: 21,575 images, 28,573 pairs, built via instance shadow detection and inpainting.  
\end{itemize}

\subsubsection{For Sketch}
\begin{itemize}[leftmargin=*]
\item \textbf{SmartShadow}~\cite{zhang2021smartshadow}: includes 1,670 artist-drawn, 25k synthetic, and 292k Internet-extracted shadow–sketch pairs, offering large stylistic diversity.
\end{itemize}

\subsection{Discussion and Trends}
\label{subsec:generation_discussion}

\xwhu{Different shadow generation paradigms require distinct data supervision and task formulations, driven by their intended applications.}  
For example, SGRNet requires a foreground shadow mask and a target shadow image for image composition. In contrast, Mask-ShadowGAN only needs unpaired shadow and shadow-free images for shadow removal. ARShadowGAN uses binary maps of real shadows and their occluders for training, generating shadows for virtual objects in augmented reality. SmartShadow utilizes line drawings and shadow pairs provided by artists to train the deep network to generate shadows on line drawings.
\xwhu{This diversity in supervision reflects a broader trend, which is from pixel-level synthesis toward semantically grounded, geometry-aware generation.}

\xwhu{Despite these advances, several open challenges remain.}  
Most existing shadow generation methods focus on single objects in static images, limiting their generalization to complex, dynamic environments.  
A key unresolved issue is \emph{how to generate temporally consistent and geometrically coherent shadows for multiple objects in video scenes}.  
Moreover, beyond generating shadows for missing regions, future work should explore \emph{interactive shadow editing}, adjusting shadow direction, softness, or intensity under user-defined or estimated lighting conditions.  
\xwhu{Such controllable, physically grounded shadow manipulation could enable unified frameworks for generation, removal, and relighting, bridging artistic and scientific illumination understanding.}

	\section{A Unified Perspective on Shadow Analysis}
\label{sec:discussion_unified}

\paragraph{Analytical linkage grounded in physics.}
\xwhu{Shadow \emph{detection}, \emph{removal}, and \emph{generation} are stages of one physically grounded process governed by illumination, geometry, and surface reflectance. Concretely: (i) detection provides spatial supports and boundary/penumbra transitions that parameterize removal via \emph{attenuation fields} and \emph{shadow mattes}; (ii) removal yields physically meaningful priors, such as attenuation ratios and intrinsic-like shading, which regularize and condition generation to respect light direction, softness, and color constancy; and (iii) generation functions as a stress test for physical plausibility (e.g., penumbra width, cast-direction consistency), exposing diagnostics that feed back to improve detection and removal objectives. This explains why specific architectural choices (e.g., direction-aware context, boundary refinement) and losses recur across tasks.}

\paragraph{Literature-backed evidence of reuse (as reflected in our tables).}
\xwhu{Our taxonomy \emph{intentionally cross-lists} methods whose modules or objectives are reused across tasks. In {Table~1} (image shadow detection) and {Table~8}/{Table~10} (image shadow removal), the following appear in both categories:
\emph{Detection $\rightarrow$ Removal}: direction-aware spatial context (DSC) first used in detection~\cite{Hu_2018_CVPR} and extended to removal in TPAMI~\cite{hu2020direction}; the stacked adversarial pipeline ST-CGAN couples a detector and a remover~\cite{wang2018stacked}; ARGAN/ARGAN+SS performs attentive detection followed by removal within one framework~\cite{ding2019argan}. These are included in detection (Table~1) \emph{and} removal (Table~8/\!10) because their detection outputs (masks/boundaries) directly condition de-shadowing and their context modules (e.g., direction-aware cues~\cite{Hu_2018_CVPR,hu2020direction}) benefit both tasks.
\emph{Removal $\leftrightarrow$ Generation}: in {Table~12} (image shadow generation), several removal models reappear as generators or supervision engines: Mask-ShadowGAN~\cite{hu2019mask} (unsupervised removal) is listed under removal (Table~8/\!10) and generation (Table~12) because its learned matte/attenuation priors also drive synthesis; Shadow Matting GAN~\cite{DBLP:conf/aaai/CunPS20} and G2R-ShadowNet~\cite{liu2021shadow} explicitly transfer matting/reflectance–shading cues between generation and removal, hence they are cross-referenced across those tables.
More broadly, composition-oriented generators, such as {ShadowGAN}~\cite{zhang2019shadowgan}, {ARShadowGAN}~\cite{liu2020arshadowgan}, {SSN} (Shadow Synthesis Network)~\cite{sheng2021ssn}, {SSG} (Shadow Style Generator)~\cite{sheng2022controllable}, {SGRNet} (Shadow Generation and Removal Network)~\cite{hong2022shadow}, {PixHt-Lab}~\cite{Sheng_2023_CVPR}, and diffusion-based pipelines such as \emph{DiffuShadow}~\cite{guo2023shadowdiffusion}, \emph{Latent-Shadow Diffusion}~\cite{mei2024latent}, \emph{Shadow-Aware Diffusion Transformer (SADT)}~\cite{tao2024shadow}, and \emph{SG-Diffusion}~\cite{Liu2024SG}, benefit when conditioned on removal or detection signals. These frameworks demonstrate that physically informed priors (e.g., attenuation ratios, shadow mattes, and light-direction embeddings) consistently improve the realism, direction consistency, and controllability of synthesized shadows, justifying their cross-references across removal and generation sections.}

\paragraph{Empirical evidence under a unified protocol.}
Under our standardized benchmark, where all models are re-trained on common splits for images and videos using unified resolutions, hardware, metrics, and released code and dataset refinements, several consistent trends emerge across tasks.
(i) Detectors with stronger boundary fidelity (Table 6/7 for instance detection) lead to lower removal error when their masks are used as conditioning signals.
(ii) Removal methods that explicitly estimate attenuation or matting (e.g., DSC, ST-CGAN, Mask-ShadowGAN) generalize better across datasets and exhibit fewer color shifts, as shown in Table 10.
(iii) Shadow generation improves in penumbra quality and light-direction consistency when guided by detection or removal priors (Table 12).
These findings provide empirical support for the cross-task analytical connections discussed above.

\paragraph{On cross-listing and citation hygiene.}
\xwhu{Because some systems are \emph{architecturally cross-task}, we adopt a clear policy: each cross-task method is \emph{introduced and fully described} in the section of its \emph{primary} contribution (e.g., DSC in detection/removal~\cite{Hu_2018_CVPR,hu2020direction}; ST-CGAN in detection+removal~\cite{wang2018stacked}; Mask-ShadowGAN and G2R-ShadowNet in removal plus generation~\cite{hu2019mask,liu2021shadow}), and \emph{referenced succinctly} in other sections where its outputs/principles are reused. This keeps the narrative non-redundant while making reuse explicit.}

\paragraph{Positioning against prior surveys.}
\xwhu{Earlier overviews often treat subproblems in isolation and lack a single re-training or evaluation protocol for different methods. In contrast, our survey (i) reviews \emph{detection, instance detection, removal, and generation} across \emph{image and video}; (ii) employs a \emph{unified benchmarking protocol} with re-trained implementations for comparability; (iii) reports \emph{size–speed–accuracy trade-offs} and \emph{cross-dataset generalization}; and (iv) releases \emph{models, code, and refined datasets/masks}. This combination provides concrete, reproducible evidence for reuse and synergy that prior surveys do not supply.}

\paragraph{Implications and guidance.}
A unified perspective highlights several reusable components across tasks. Edge and penumbra encoders, as well as direction-aware context modules developed for detection, provide strong spatial and illumination cues. Attenuation and matte estimators, together with intrinsic-like decompositions from removal models, supply physically grounded representations of shading and reflectance. Physically guided conditioning signals used in generation further reinforce consistency in light direction and penumbra structure. In practice, accurate boundary detection improves removal quality, and attenuation- or matte-based removal produces more controllable and physically plausible shadow generation. Aligning these shared modules and objectives suggests a viable path toward a unified model that handles detection, removal, and generation within a single framework.

\paragraph{Scope and limitations.}
\xwhu{Our synthesis focuses on ground-level images and videos, with remote sensing briefly discussed in Sec.~\ref{sec:1.2} for contrast.
To ensure fair evaluation, we adopt a unified protocol that standardizes resolution, metrics, and training settings; however, residual dataset bias and implementation variance may still influence results.
To mitigate these factors, we conduct cross-dataset evaluations, provide visual comparisons in the supplementary material, and release all results online for transparency. 
Finally, we note that traditional metrics such as BER and F-measure mainly reflect pixel-level accuracy, offering limited insight into perceptual or illumination realism.}

\section{Future Directions}
\label{sec:future}

We outline five open challenges and research directions, emphasizing how geometry-, semantics-, and foundation-model-based reasoning can reshape the landscape of shadow understanding and manipulation.

\vspace{1.5mm}
\noindent\textbf{\emph{(1) Toward unified, all-in-one frameworks for shadow understanding.}} 
Most current approaches address only detection, removal, or generation individually, even though all three tasks are driven by shared illumination and occlusion physics.
Developing a multi-task, all-in-one framework that jointly models shadow-object relations across detection, removal, and generation can leverage shared geometric priors and reduce redundant supervision.  
Recent progress in unified vision–language architectures and transformer backbones suggests that transferable representations across image, video, and sketch domains could generalize better to complex illumination scenarios.

\vspace{1.5mm}
\noindent\textbf{\emph{(2) Semantics- and geometry-aware shadow reasoning.}} 
The semantics and geometries of objects remain underexplored in shadow analysis.  
\xwhu{While early CNN-based models relied mainly on low-level cues (edges, intensity, or texture), large vision and vision–language models (e.g., SAM~\cite{kirillov2023segment,ravi2024sam}, Depth Anything~\cite{yang2024depth,depth_anything_v2}, and InternVL~\cite{chen2024internvl,wang2023internimage}) now provide dense segmentation, depth estimation, and semantic grounding that can be exploited for illumination reasoning.  
Recent 3D-aware methods, such as LERF~\cite{Kerr2023LERF} and NeRFactor~\cite{Zhang2021NeRFactor}, explicitly disentangle geometry, reflectance, and shadow visibility, enabling physically interpretable scene understanding.  
Furthermore, relightable radiance-field models including NeRF-OSR~\cite{Rudnev2022NeRFOSR}, ReNeRF~\cite{Xu2023ReNeRF}, and diffusion-based relighting~\cite{PoirierGinter2024DiffRelight} illustrate how lighting and viewpoint control can be unified while maintaining shadow coherence.  
Challenges remain in building large-scale 3D datasets with accurate shadow labels, disentangling illumination from material effects, and scaling differentiable rendering for realistic supervision.}

\vspace{1.5mm}
\noindent\textbf{\emph{(3) Shadow–object relationships for intelligent editing and scene manipulation.}} 
Instance-level shadow detection and association directly support editing tasks such as inpainting, relighting, and composition~\cite{wang2020instance,wang2021single,wang2023instance,xing2024video}.  
\xwhu{We clarify that technologies like multi-camera systems, HDR imaging, and neural radiance fields can be exploited to reconstruct coherent shadow geometry across multiple views and lighting conditions.  
For instance, NeRF-OSR~\cite{Rudnev2022NeRFOSR} and ReLight-My-NeRF~\cite{Toschi2023ReLightMyNeRF} enable relightable view synthesis.  
Future research may focus on dynamic shadow editing and temporal consistency, integrating instance shadow detection with inverse rendering and photometric calibration to support immersive AR/VR or cinematic applications.}

\vspace{1.5mm}
\noindent\textbf{\emph{(4) Shadow-based evaluation of AI-generated content (AIGC).}}  
AI-generated imagery often exhibits geometric inconsistencies in shadows.  
\xwhu{Analyzing illumination alignment between objects and their shadows can serve as a robust cue for authenticity verification and forensics~\cite{bhaumik2023exploiting}.  
Conversely, shadows can act as stealthy adversarial perturbations that degrade model predictions~\cite{zhong2022shadows}.  
Exploring shadow-consistency metrics and geometry-aware discriminators may improve AIGC trustworthiness and robustness.}

\vspace{1.5mm}
\noindent\textbf{\emph{(5) Integration with multimodal foundation models.}}  
\xwhu{Embedding shadow reasoning into large multimodal foundation models, such as InternVL~\cite{chen2024internvl,wang2023internimage} or Kosmos-2~\cite{peng2023kosmos}, represents an emerging frontier.  
Incorporating physics-informed illumination priors and geometric constraints into these large backbones could unify perception, reasoning, and generation within one agent framework, capable of not only detecting or removing shadows but also understanding their semantic and spatial roles in context.}

\vspace{1.5mm}
\noindent\textbf{\emph{Applications.}}
Beyond academic benchmarks, shadow understanding is increasingly vital in applied domains.
In remote sensing, shadow removal enhances land-cover classification and urban mapping by correcting illumination imbalance in aerial and satellite imagery~\cite{liu2022shadowremotesensing}.
In 3D reconstruction, shadow-aware neural radiance fields improve photometric consistency and surface geometry recovery under varying light conditions~\cite{Derksen2021ShadowNeRF}.
In autonomous driving and robotics, integrating shadow cues strengthens perception in complex illumination, reducing visual ambiguity and improving depth estimation~\cite{Li2024ShadowAwareNeRF}.
These applications demonstrate how advances in shadow modeling directly support high-level visual understanding and real-world decision making.

\vspace{1.5mm}
\noindent\textbf{\emph{Outlook.}}  
We foresee future shadow understanding systems evolving from task-specific pipelines into holistic, physically grounded, and semantically aware models that bridge geometry, illumination, and multimodal reasoning.  
This convergence will advance both scientific understanding and various real-world applications, from robust perception and visual authenticity verification to intelligent, controllable image and video editing.

\section{Conclusion}
\label{sec:conclusion}

This survey provides the first unified and comprehensive review of deep-learning-based shadow detection, removal, and generation across images and videos. By analyzing more than one hundred methods, we establish consistent architectural and supervisory taxonomies, clarify the evolution of technical paradigms, and standardize experimental protocols for fair comparison. 
Our benchmarking under unified training and evaluation settings reveals the influence of model design, resolution, and supervision on performance, and exposes substantial discrepancies in previously reported results. 
Cross-dataset experiments further highlight the limited generalization of existing approaches and the impact of dataset bias. 
We also synthesize common principles shared across detection, removal, and generation, showing how they interact through illumination priors, semantic cues, and scene geometry.
Overall, this work consolidates fragmented developments in shadow analysis, provides reproducible baselines and corrected datasets, and offers an integrated perspective that supports both newcomers and experienced researchers in understanding the current landscape of the field.

	\begin{acknowledgements}
		This work was supported by the Research Start-up Fund for Prof. Xiaowei Hu at the Guangzhou International Campus, South China University of Technology (Grant No. K3250310). X. Hu and Z. Xing are joint first authors.
	\end{acknowledgements}
	
	{\small
		\bibliography{reference} 
	}
	
	\appendix

\section{Visual Comparisons}\label{app}
This appendix consists of five parts. 
Parts 1 to 5 provide visual comparisons of various methods applied to image shadow detection, video shadow detection, instance shadow detection, general image shadow removal, and document shadow removal, respectively.
The images selected for comparison are chosen according to the criterion of significant differences among the results of the compared methods and between each method's results and the ground-truth images.

\clearpage
\onecolumn
\vspace*{2mm}
\large
\noindent
\section*{Part 1: Visual Comparisons on Image Shadow Detection}\label{part1}
\vspace{5mm}
\begin{figure*}[hbp]
	\centering
	\includegraphics[width=1\linewidth]{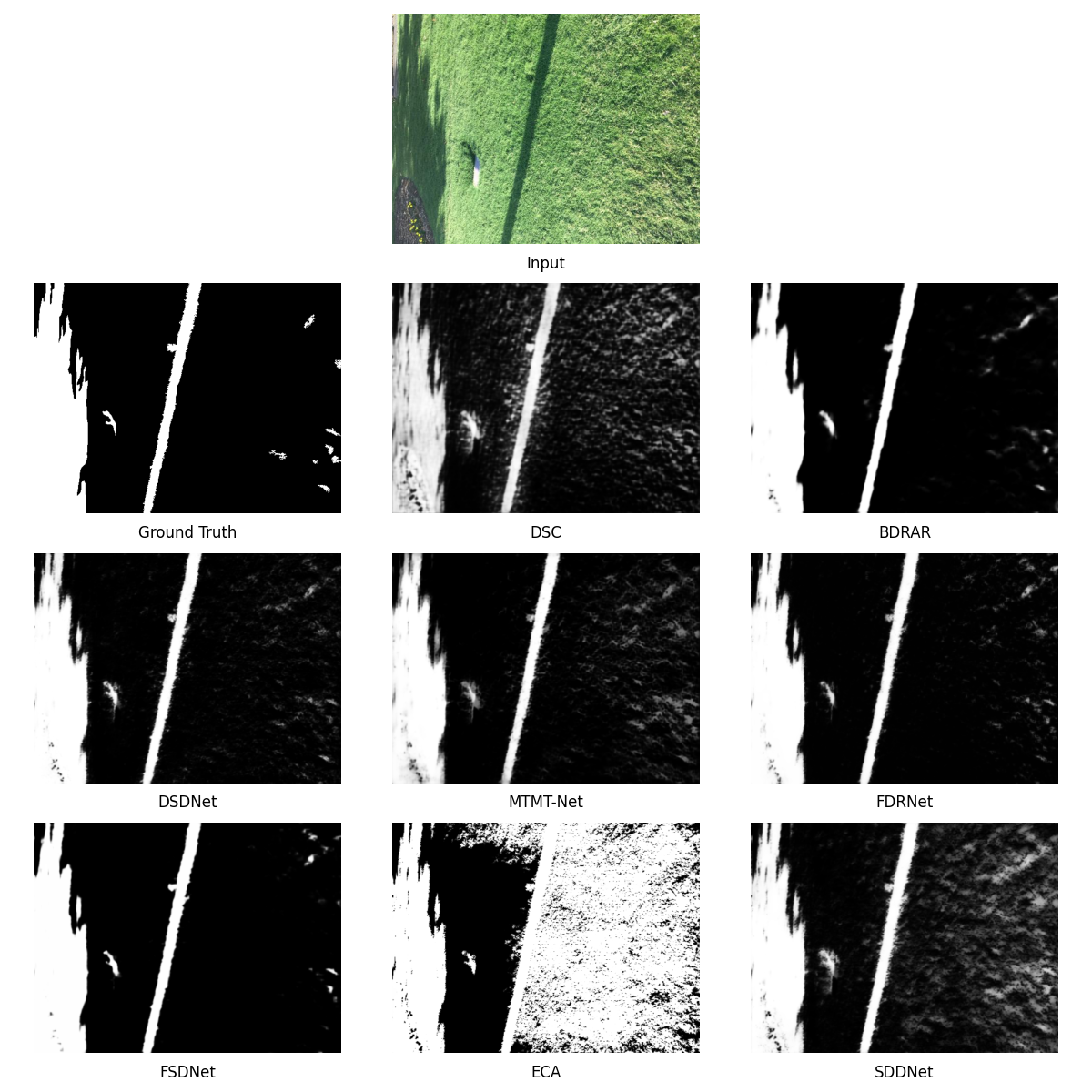}
	\caption{Visual comparison result \#1 on the CUHK-Shadow dataset (white indicates shadows and black indicates non-shadows).}
	\label{fig:supp_sd1}

\end{figure*}

\begin{figure*}[hbp]
	\centering
	\includegraphics[width=1\linewidth]{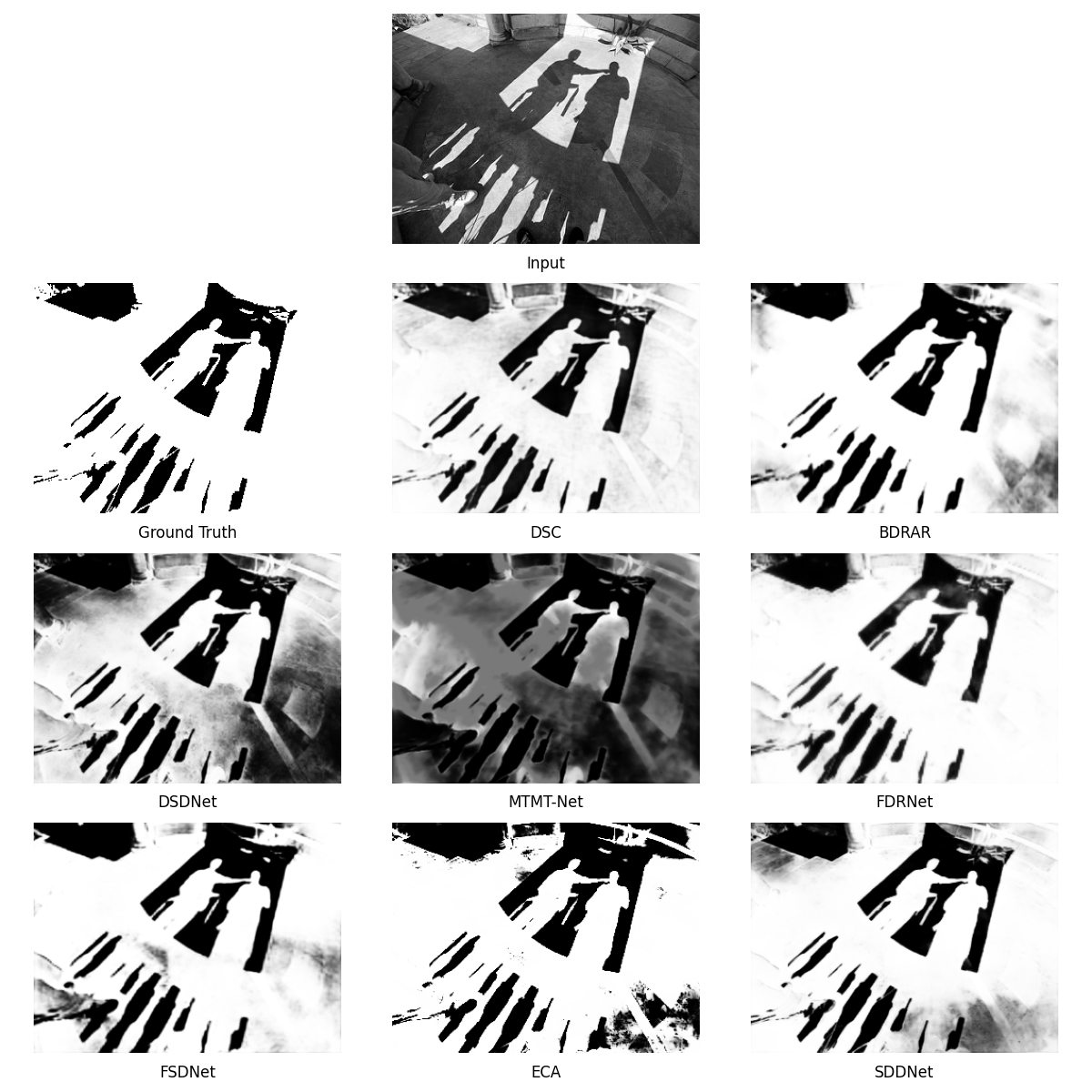}
	\caption{Visual comparison result \#2 on the CUHK-Shadow dataset (white indicates shadows and black indicates non-shadows).}
	\label{fig:supp_sd2}

\end{figure*}

\begin{figure*}[hbp]
	\centering
	\includegraphics[width=1\linewidth]{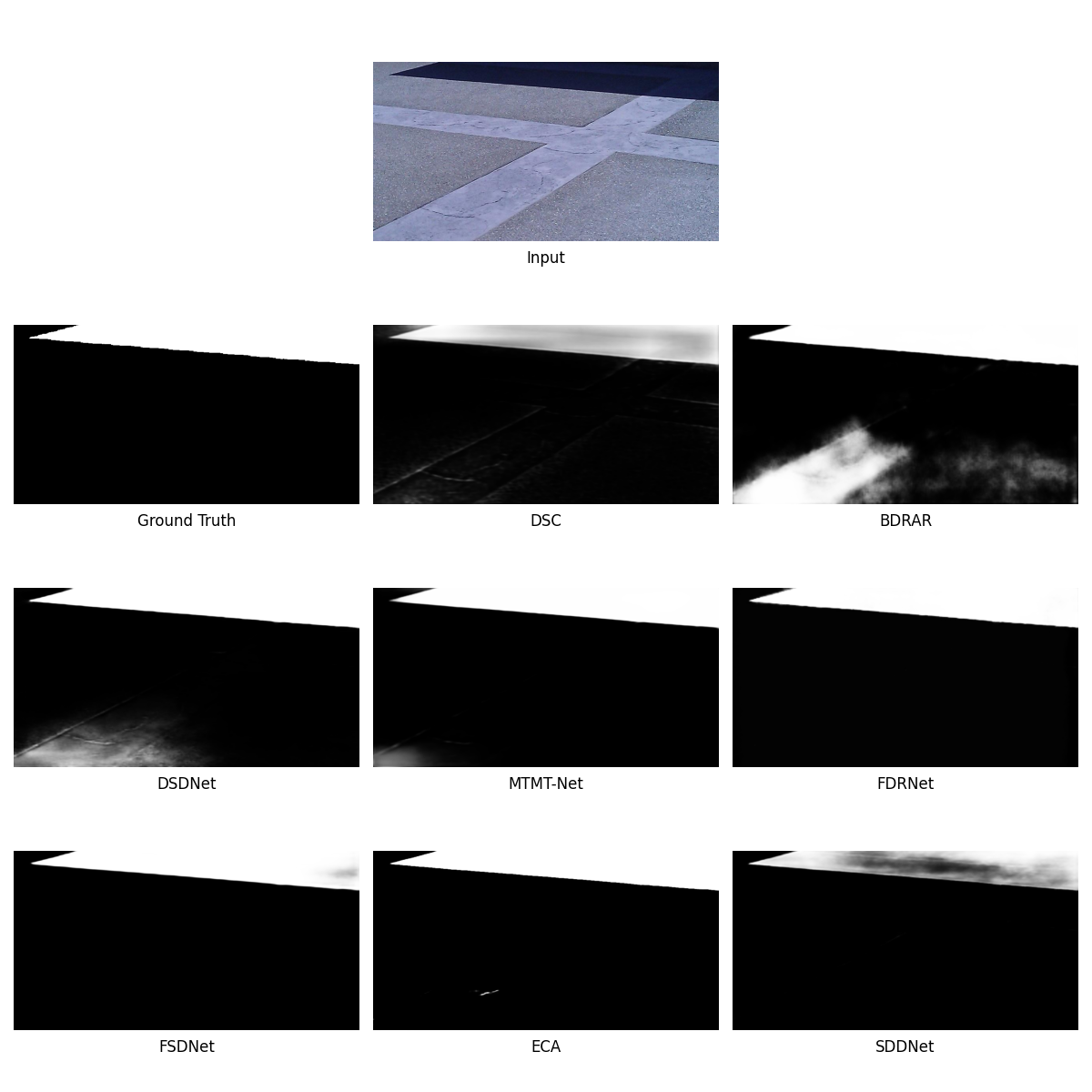}
	\caption{Visual comparison result \#3 on the SBU-Refined dataset (white indicates shadows and black indicates non-shadows).}
	\label{fig:supp_sd3}

\end{figure*}

\begin{figure*}[hbp]
	\centering
	\includegraphics[width=1\linewidth]{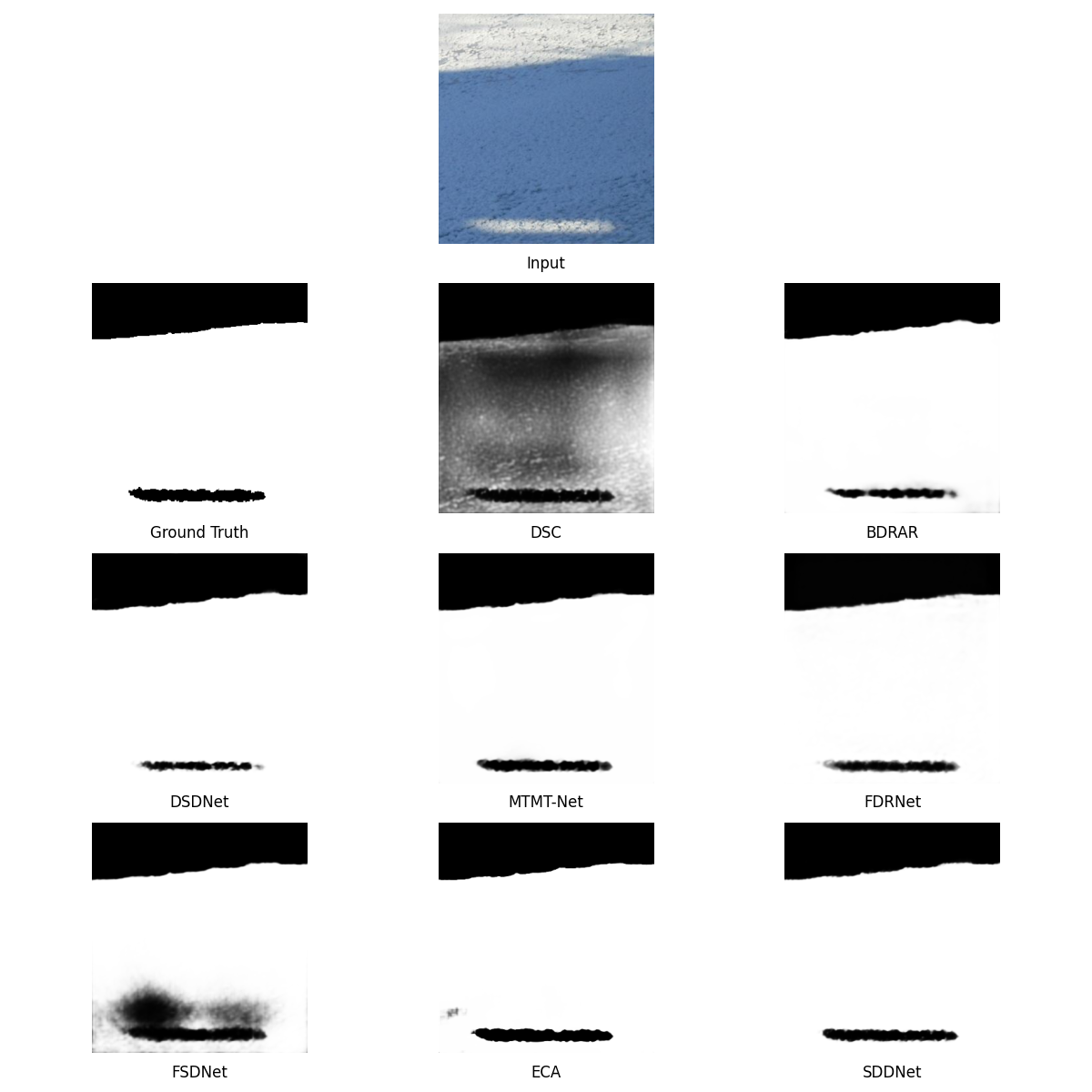}
	\caption{Visual comparison result \#4 on the SBU-Refined dataset (white indicates shadows and black indicates non-shadows).}
	\label{fig:supp_sd4}

\end{figure*}

\begin{figure*}[hbp]
	\centering
	\includegraphics[width=1\linewidth]{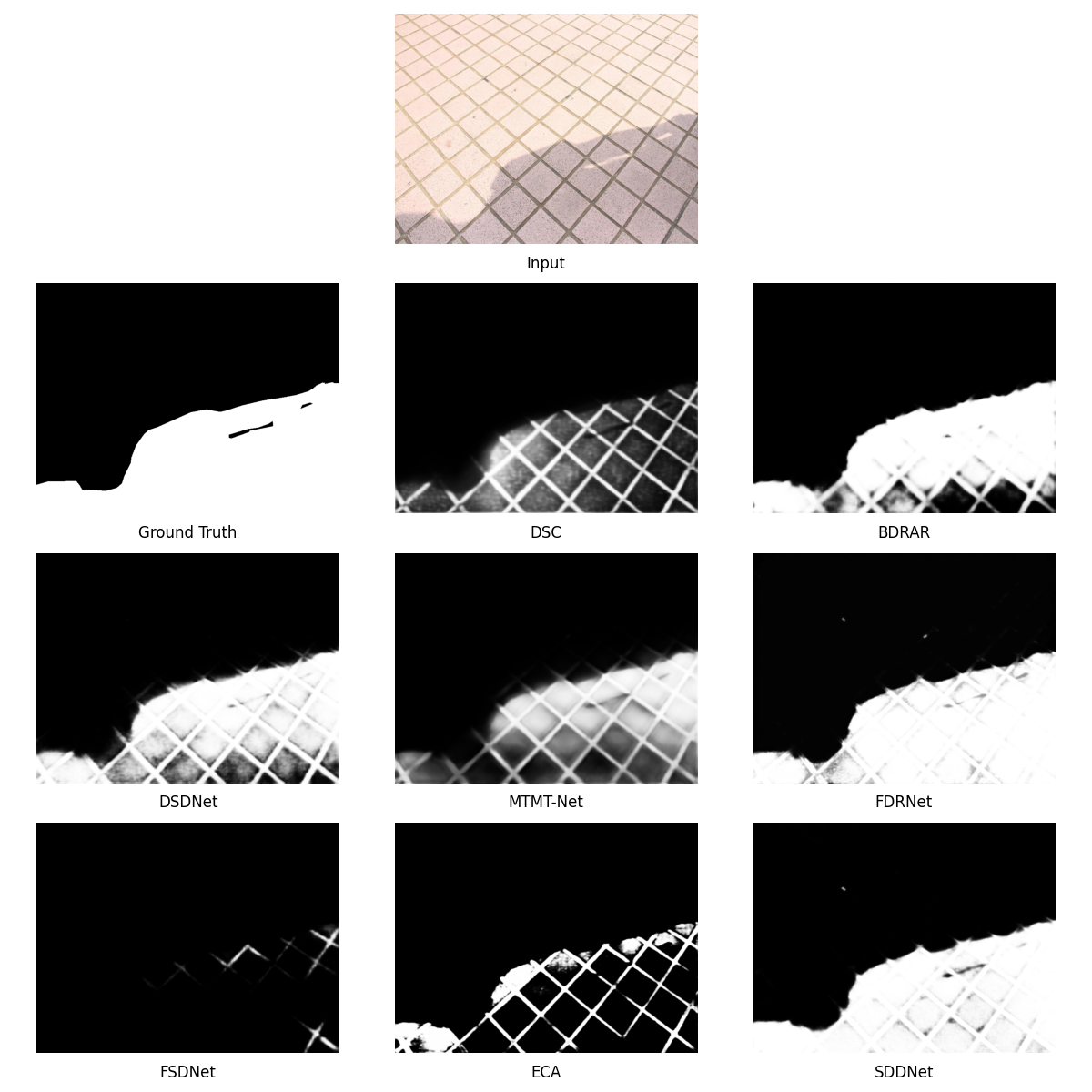}
	\caption{Visual comparison result \#5 on the SRD  dataset (cross-dataset generalization evaluation; white indicates shadows and black indicates non-shadows).}
	\label{fig:supp_sd5}

\end{figure*}

\begin{figure*}[hbp]
	\centering
	\includegraphics[width=1\linewidth]{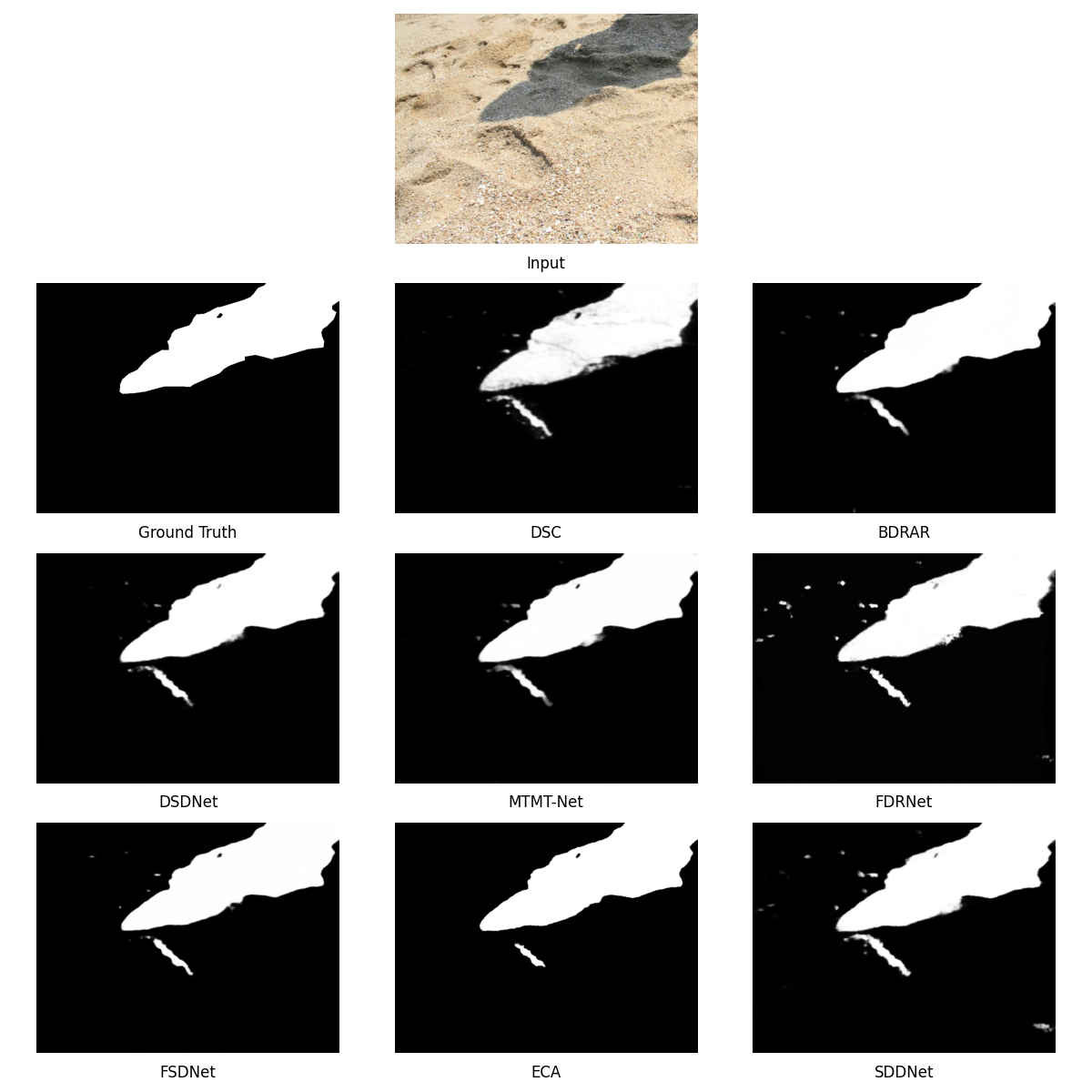}
	\caption{Visual comparison result \#6 on the SRD  dataset (cross-dataset generalization evaluation; white indicates shadows and black indicates non-shadows).}
	\label{fig:supp_sd6}

\end{figure*}

\clearpage
\onecolumn
\vspace*{2mm}
\large
\noindent
\section*{Part 2: Visual Comparisons on Video Shadow Detection}\label{part2}
\vspace{5mm}
\begin{figure*}[hbp]
	\centering
	\includegraphics[width=1\linewidth]{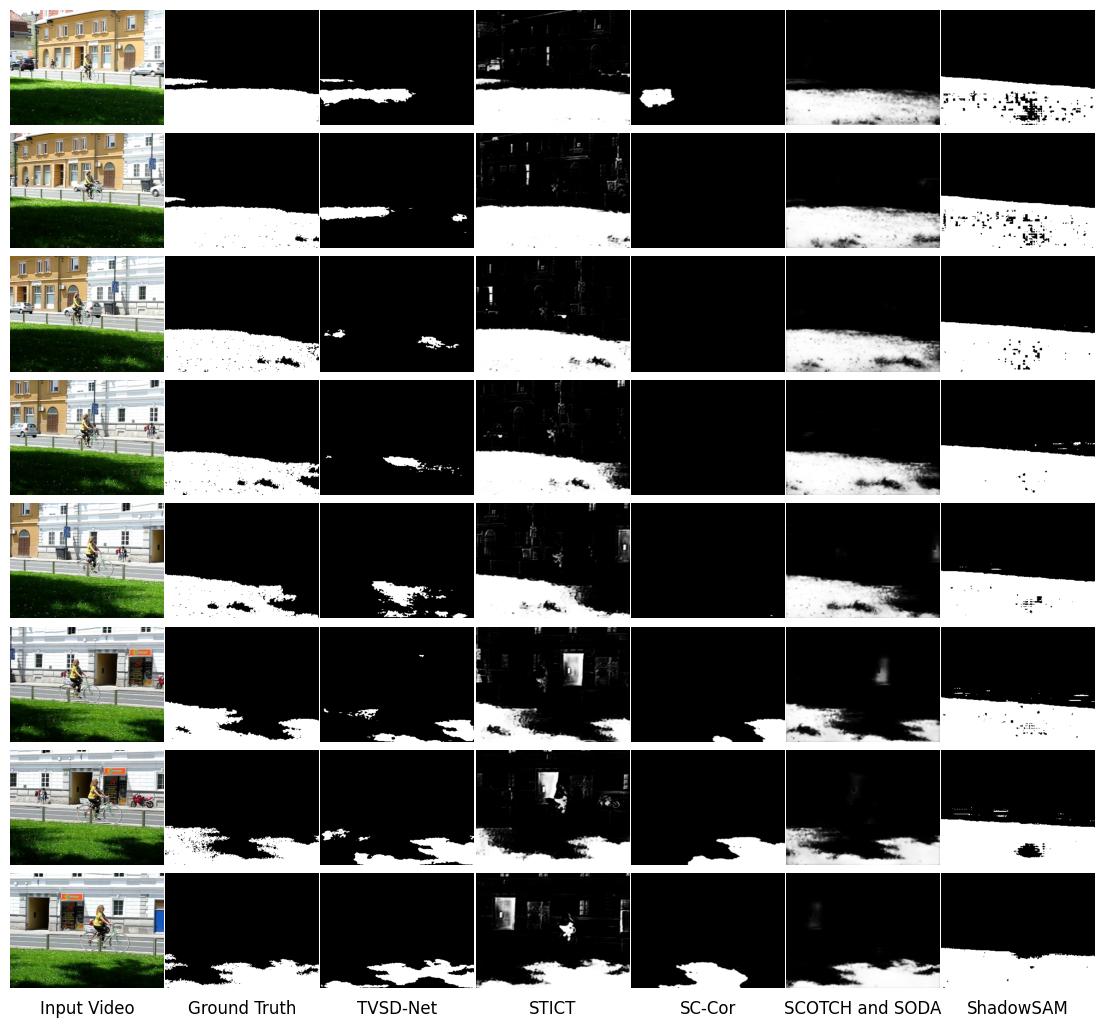}
	\caption{Visual comparison result \#1 on the ViSha dataset (white indicates shadows and black indicates non-shadows).}
	\label{fig:supp_vsd1}
\end{figure*}

\begin{figure*}[hbp]
	\centering
	\includegraphics[width=1\linewidth]{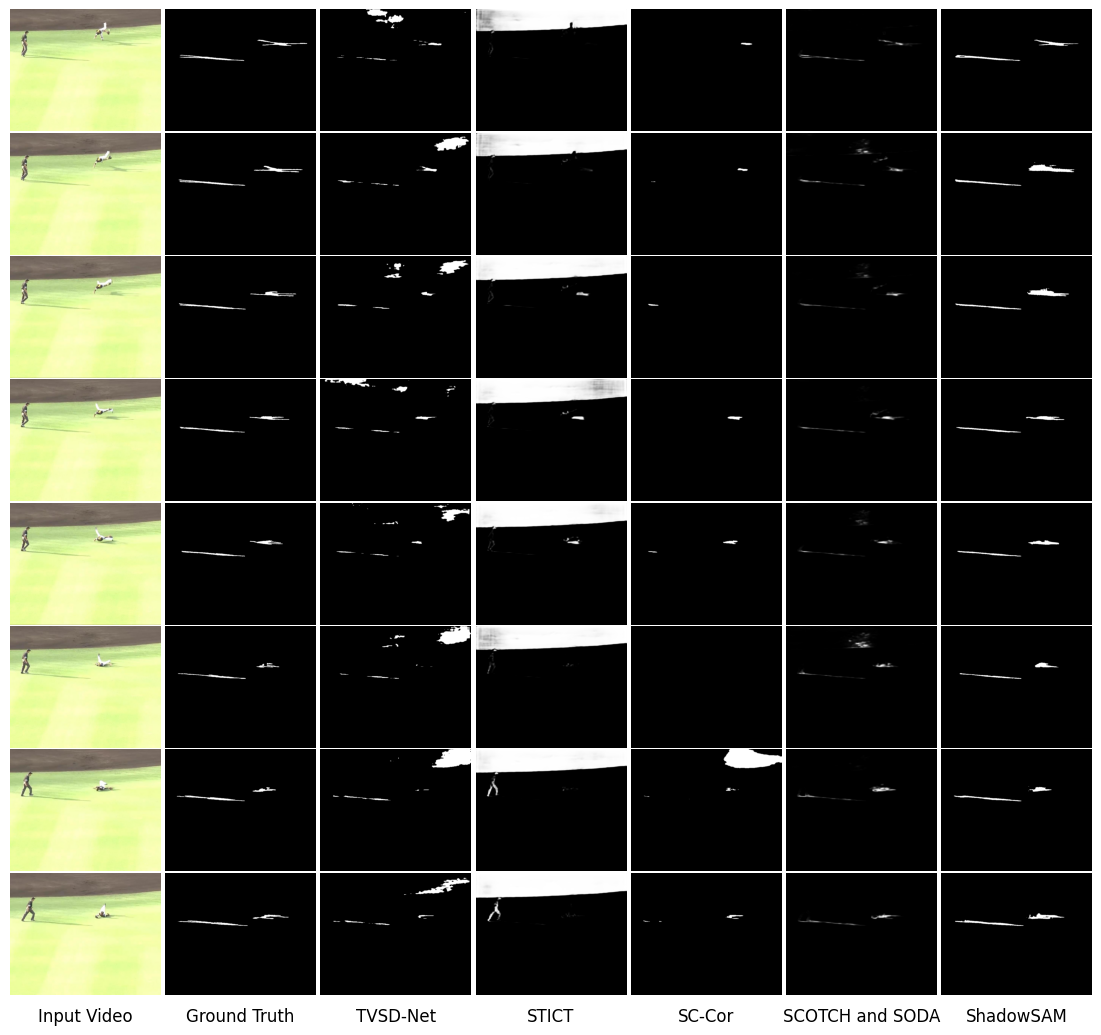}
	\caption{Visual comparison result \#2 on the ViSha dataset (white indicates shadows and black indicates non-shadows).}
	\label{fig:supp_vsd2}
\end{figure*}

\begin{figure*}[hbp]
	\centering
	\includegraphics[width=1\linewidth]{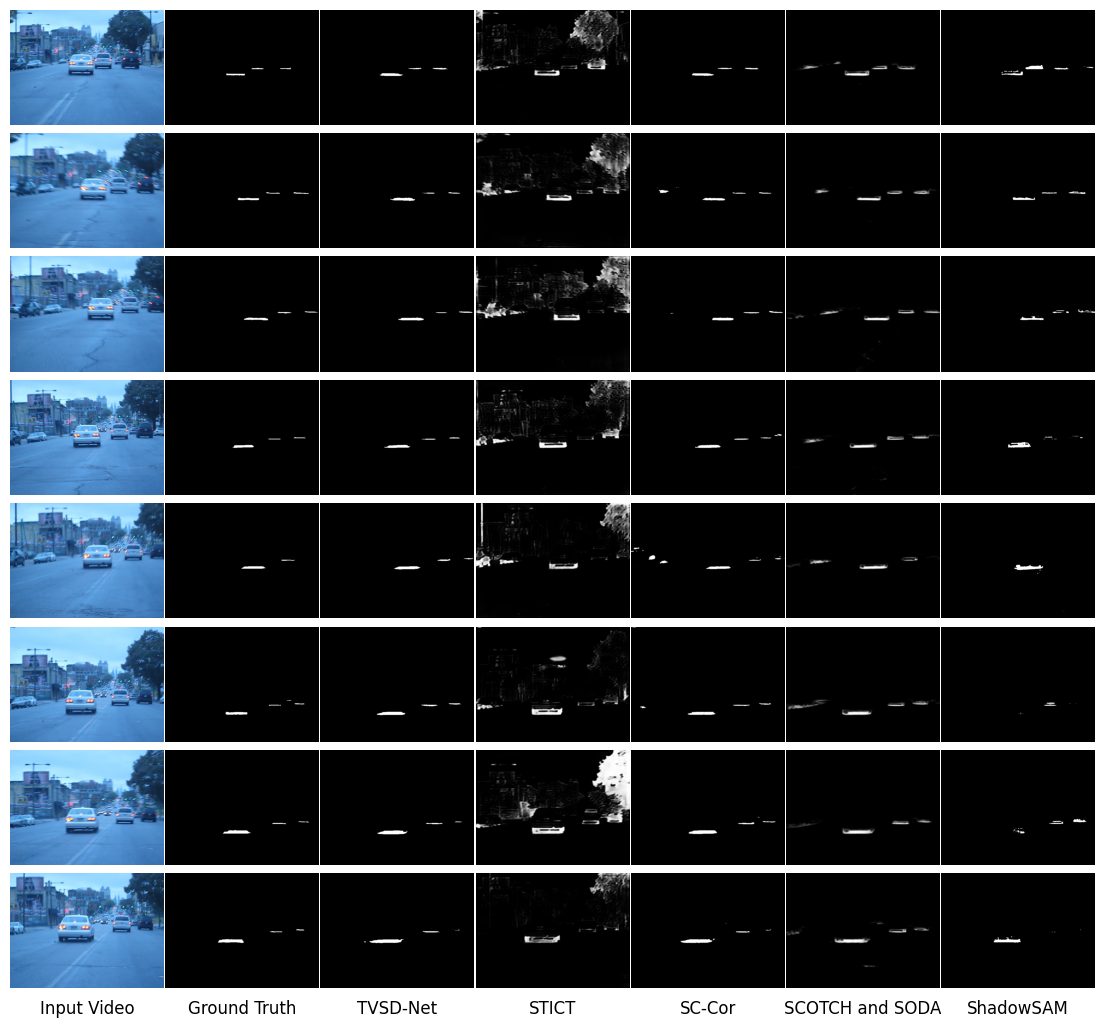}
	\caption{Visual comparison result \#3 on the ViSha dataset (white indicates shadows and black indicates non-shadows).}
	\label{fig:supp_vsd3}
\end{figure*}

\begin{figure*}[hbp]
	\centering
	\includegraphics[width=1\linewidth]{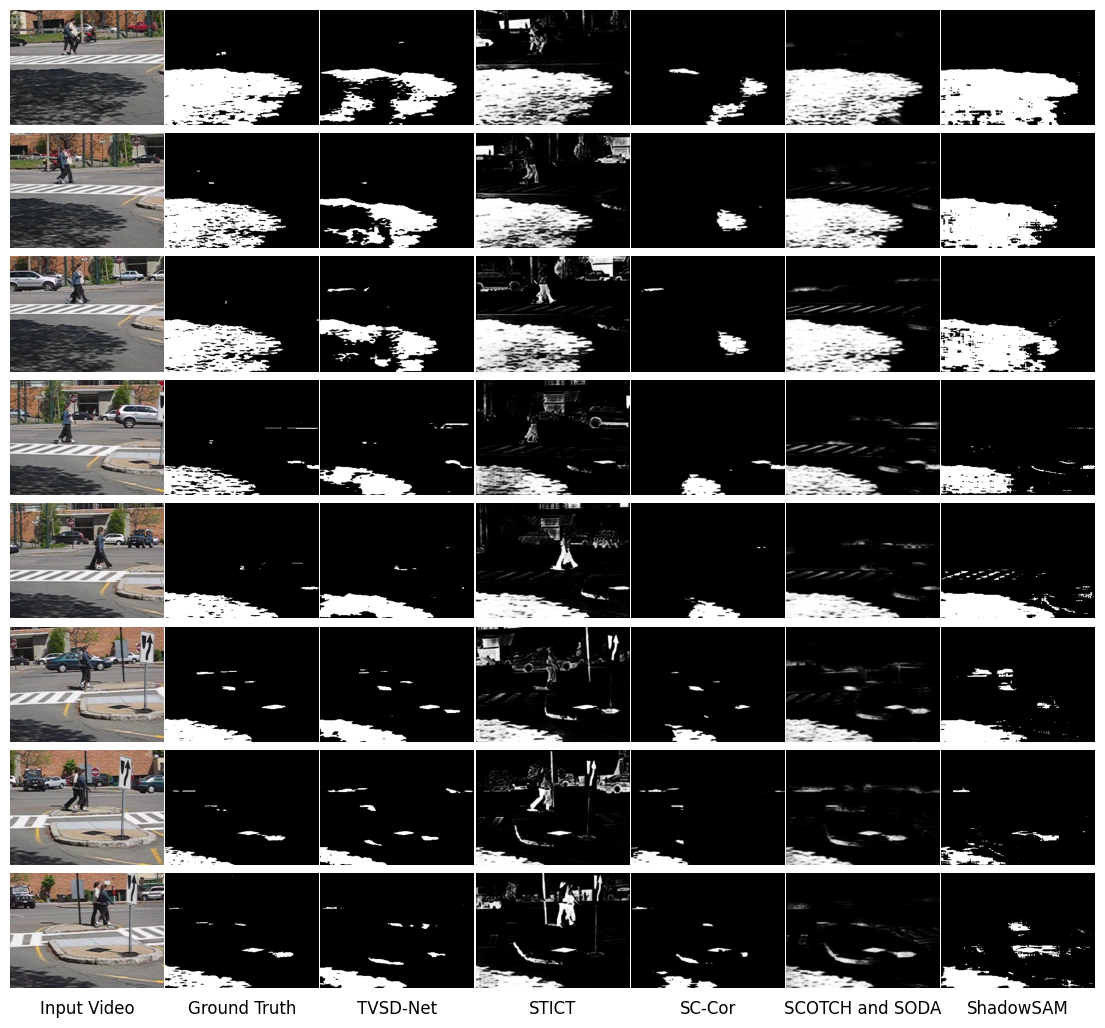}
	\caption{Visual comparison result \#4 on the ViSha dataset (white indicates shadows and black indicates non-shadows).}
	\label{fig:supp_vsd4}
\end{figure*}

\clearpage
\onecolumn
\vspace*{2mm}
\large
\noindent
\section*{Part 3: Visual Comparisons on Instance Shadow Detection}\label{part3}
\vspace{5mm}
\begin{figure*}[hbp]
	\centering
	\includegraphics[width=.75\linewidth]{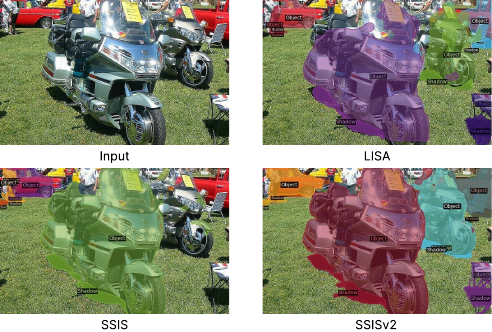}
	\caption{Visual comparison result \#1 on the SOBA dataset (paired shadow and object instances are indicated in the same color).}
	\label{fig:supp_isd1}

\end{figure*}

\begin{figure*}[hbp]
	\centering
	\includegraphics[width=.75\linewidth]{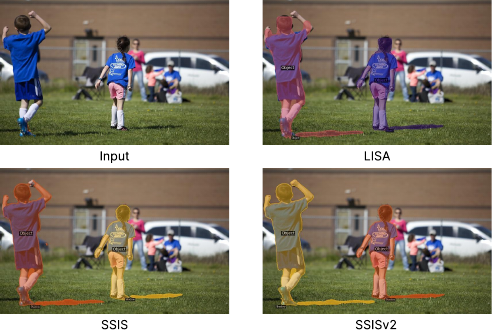}
	\caption{Visual comparison result \#2 on the SOBA dataset (paired shadow and object instances are indicated in the same color).}
	\label{fig:supp_isd2}

\end{figure*}

\begin{figure*}[hbp]
	\centering
	\includegraphics[width=.75\linewidth]{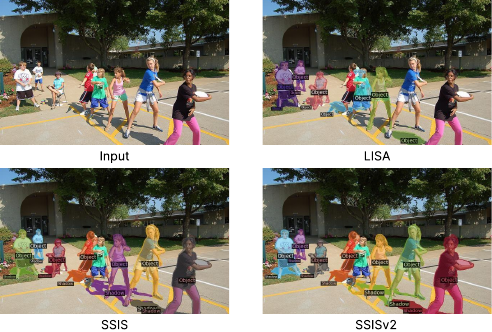}
	\caption{Visual comparison result \#3 on the SOBA dataset (paired shadow and object instances are indicated in the same color).}
	\label{fig:supp_isd3}

\end{figure*}

\begin{figure*}[hbp]
	\centering
	\includegraphics[width=.75\linewidth]{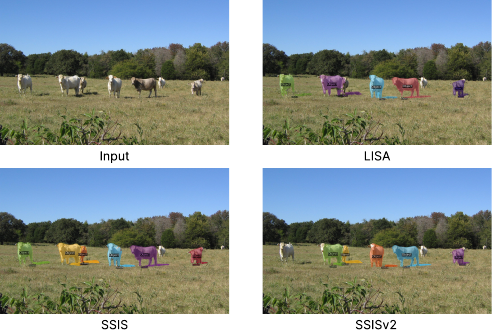}
	\caption{Visual comparison result \#4 on the SOBA dataset (paired shadow and object instances are indicated in the same color).}
	\label{fig:supp_isd4}

\end{figure*}

\begin{figure*}[hbp]
	\centering
	\includegraphics[width=.75\linewidth]{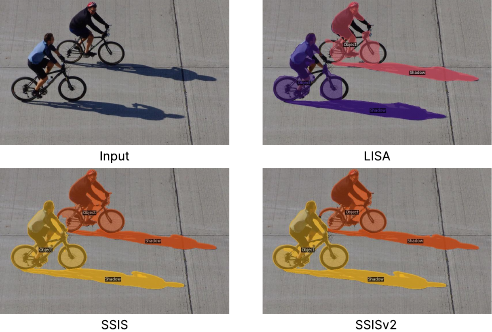}
	\caption{Visual comparison result \#5 on the SOBA dataset (paired shadow and object instances are indicated in the same color).}
	\label{fig:supp_isd5}

\end{figure*}

\begin{figure*}[hbp]
	\centering
	\includegraphics[width=.75\linewidth]{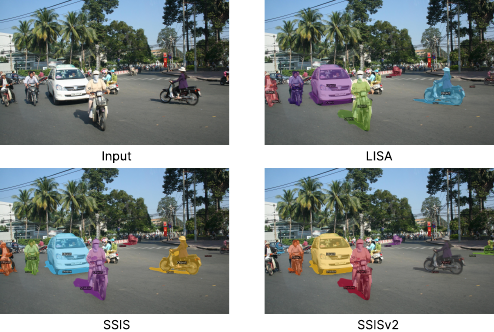}
	\caption{Visual comparison result \#6 on the SOBA dataset (paired shadow and object instances are indicated in the same color).}
	\label{fig:supp_isd6}

\end{figure*}

\clearpage
\onecolumn
\vspace*{2mm}
\large
\noindent
\section*{Part 4: Visual Comparisons on Image Shadow Removal}\label{part4}
\vspace{5mm}

\begin{figure*}[hbp]
	\centering
	\includegraphics[width=1\linewidth]{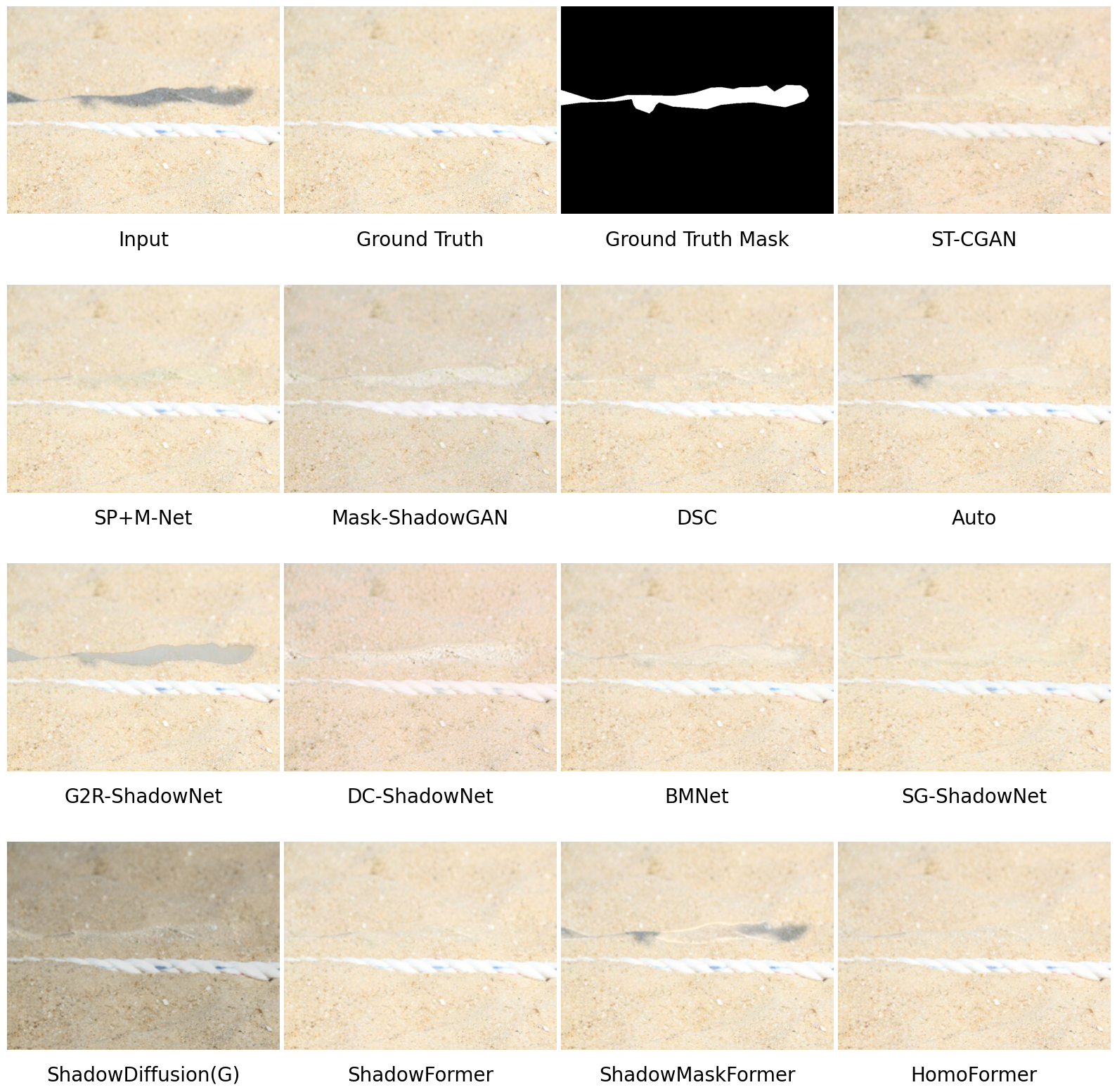}
	\caption{Visual comparison result \#1 on the SRD dataset.}
	\label{fig:supp_sr1}

\end{figure*}

\begin{figure*}[hbp]
	\centering
	\includegraphics[width=1\linewidth]{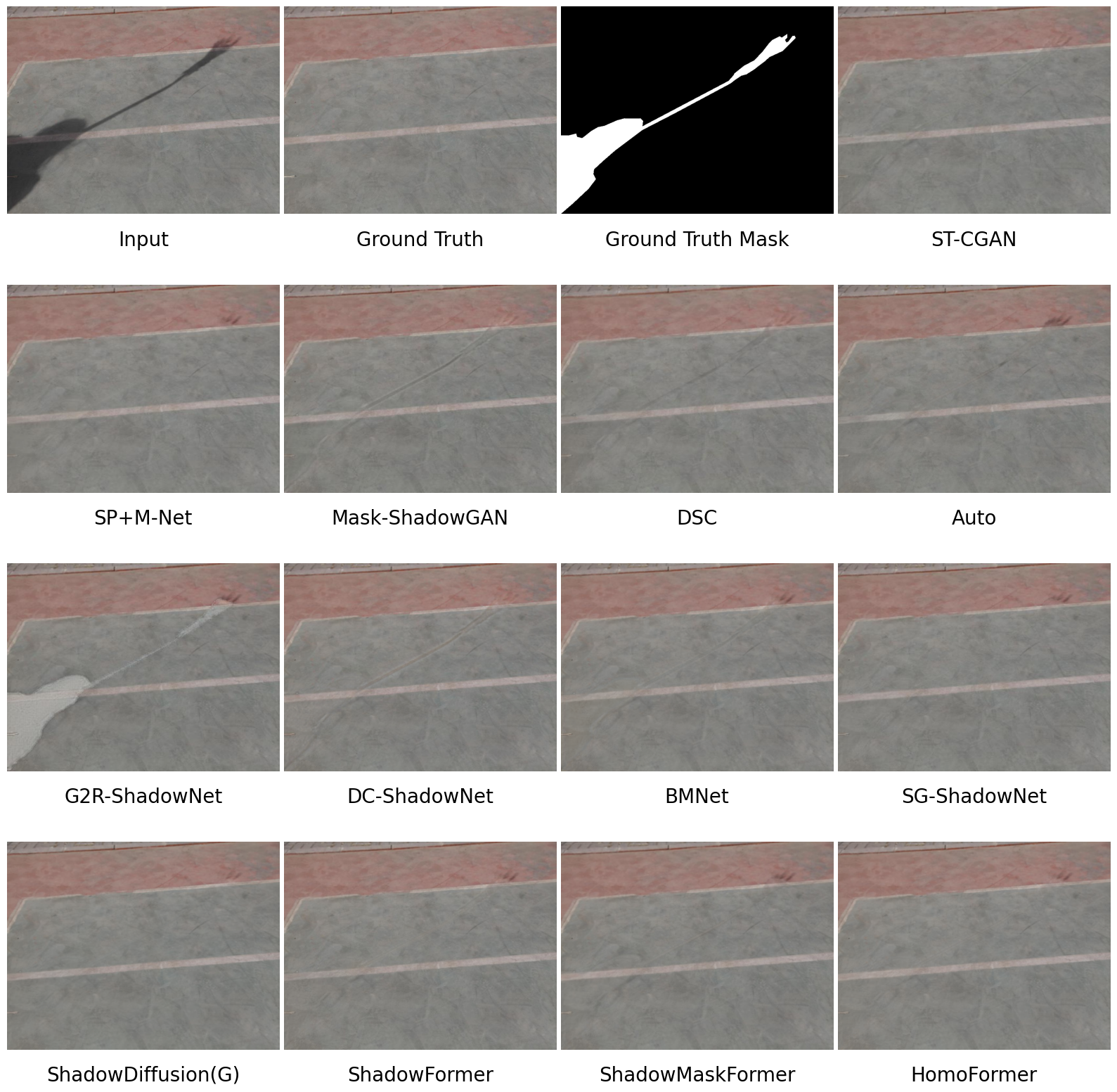}
	\caption{Visual comparison result \#2 on the SRD dataset.}
	\label{fig:supp_sr2}

\end{figure*}

\begin{figure*}[hbp]
	\centering
	\includegraphics[width=1\linewidth]{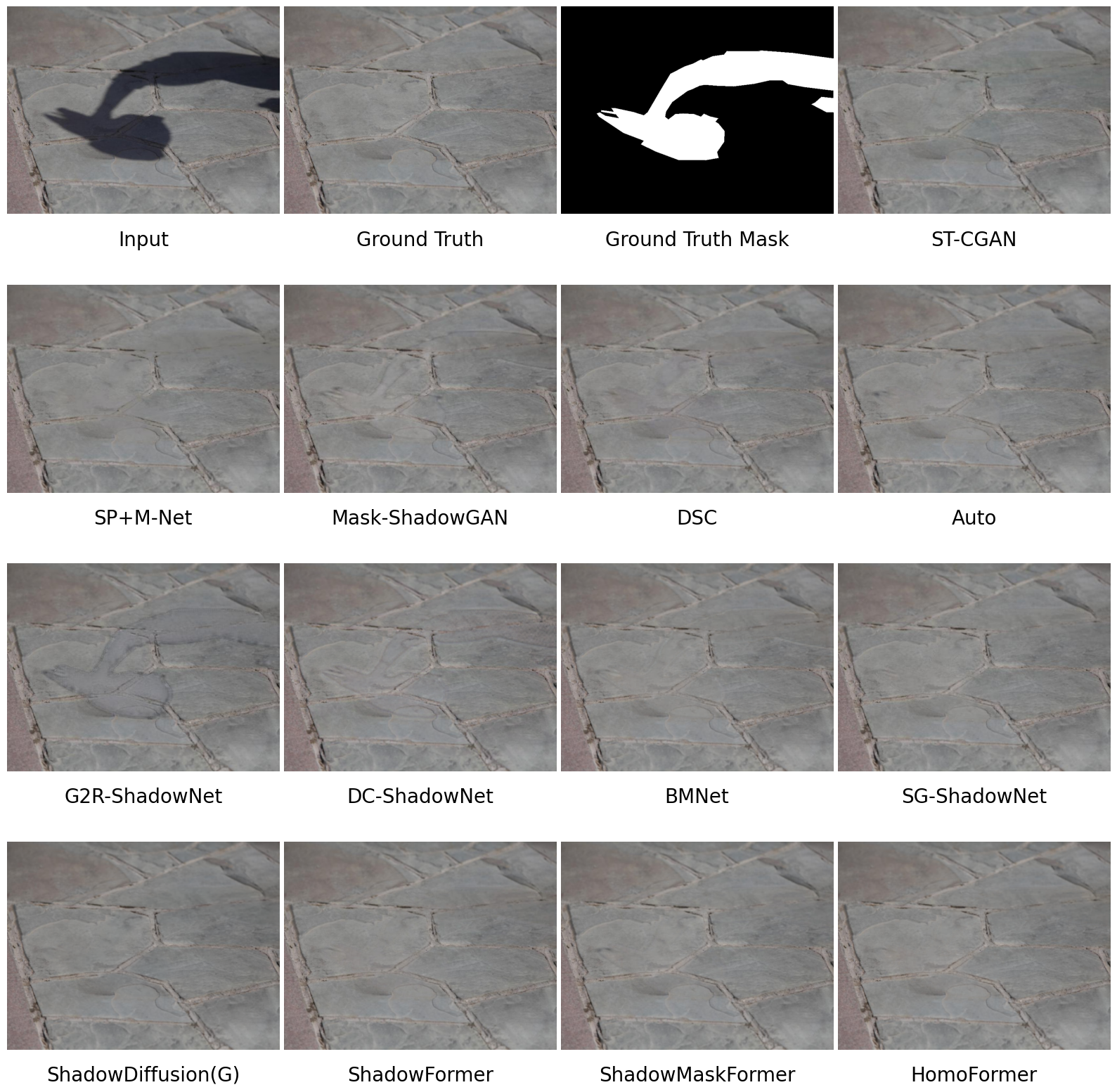}
	\caption{Visual comparison result \#3 on the SRD dataset.}
	\label{fig:supp_sr3}
\end{figure*}

\begin{figure*}[hbp]
	\centering
	\includegraphics[width=1\linewidth]{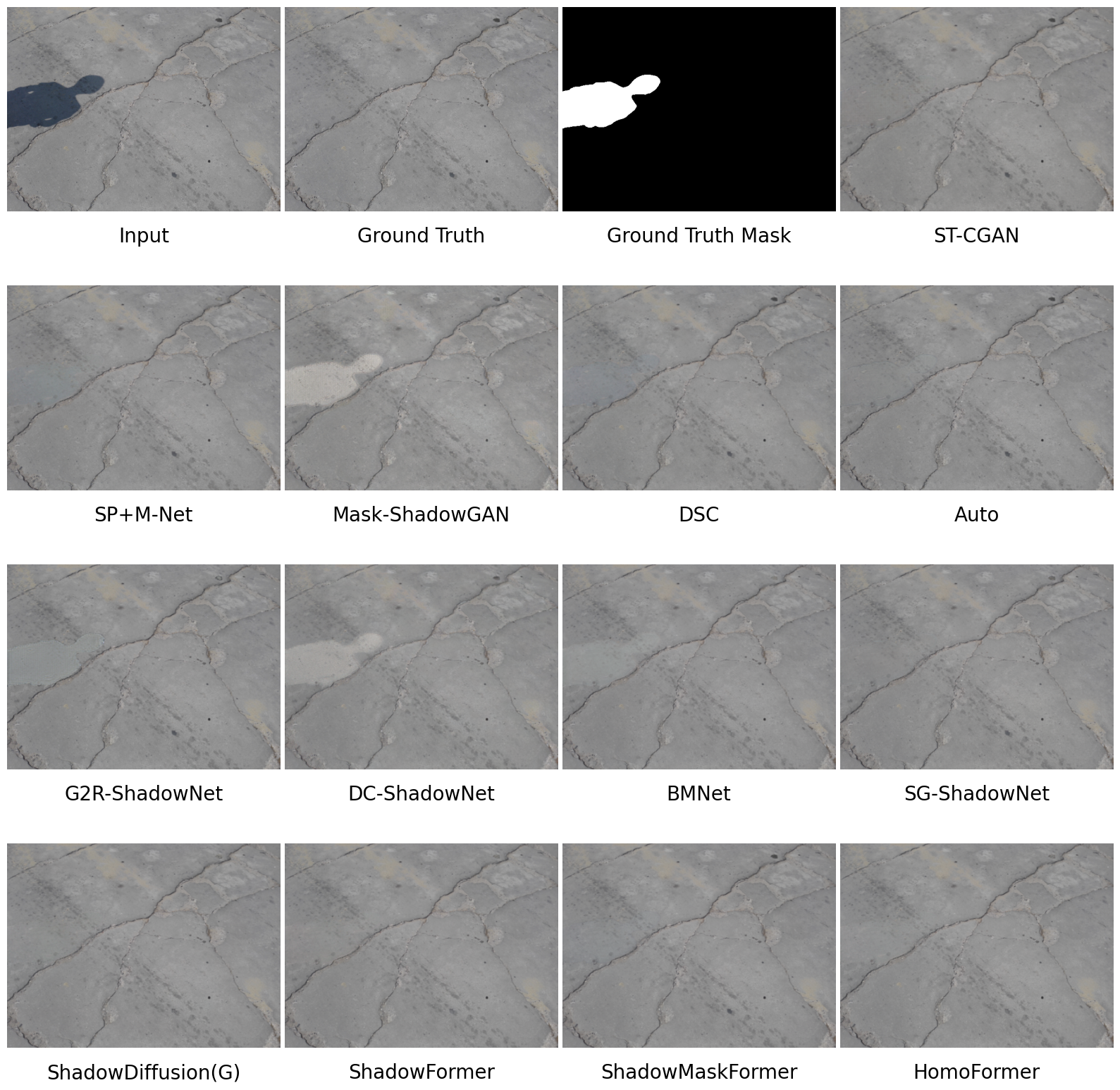}
	\caption{Visual comparison result \#4 on the ISTD+ dataset.}
	\label{fig:supp_sr4}
\end{figure*}

\begin{figure*}[hbp]
	\centering
	\includegraphics[width=1\linewidth]{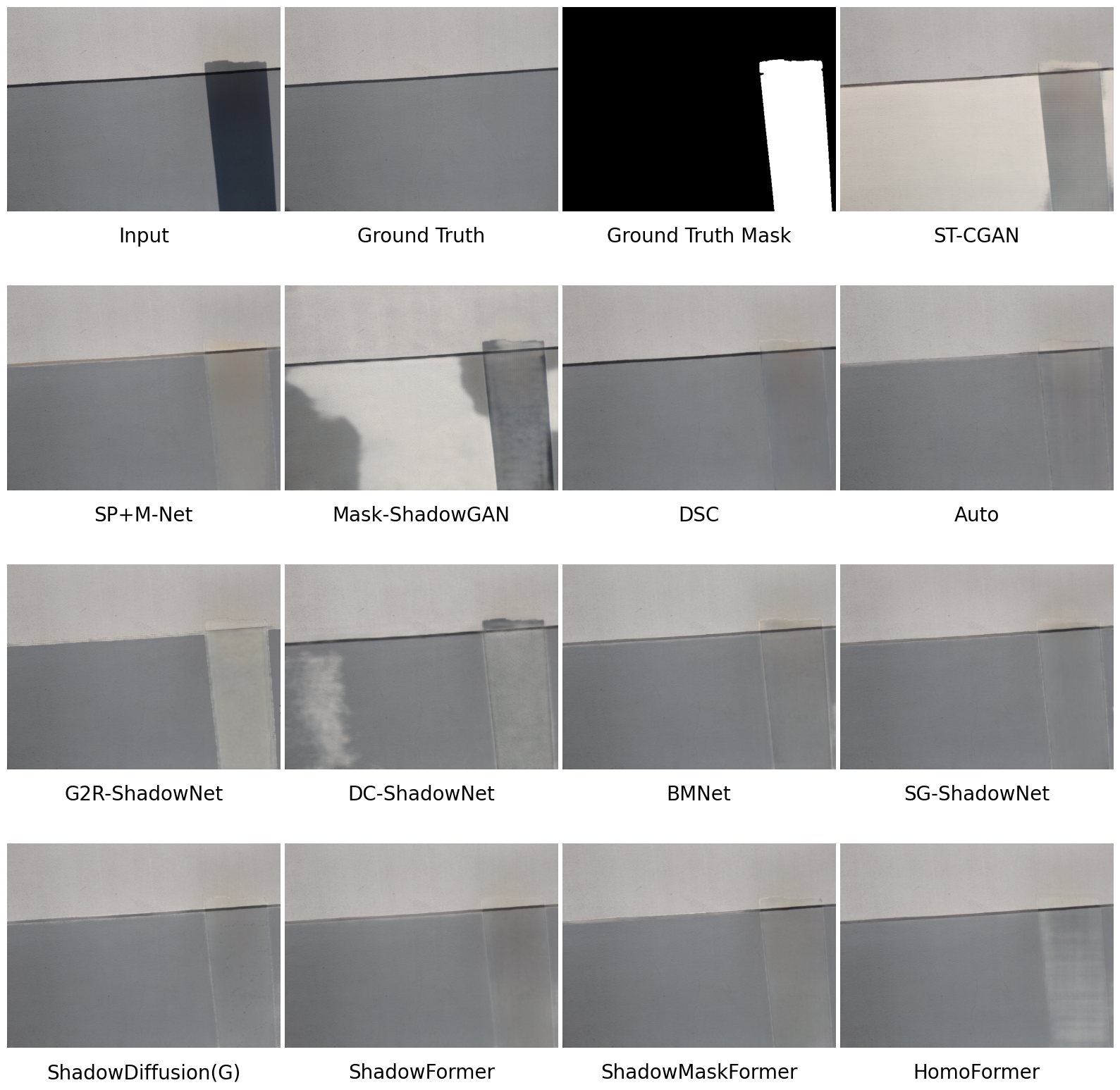}
	\caption{Visual comparison result \#5 on the ISTD+ dataset.}
	\label{fig:supp_sr5}
\end{figure*}

\begin{figure*}[hbp]
	\centering
	\includegraphics[width=1\linewidth]{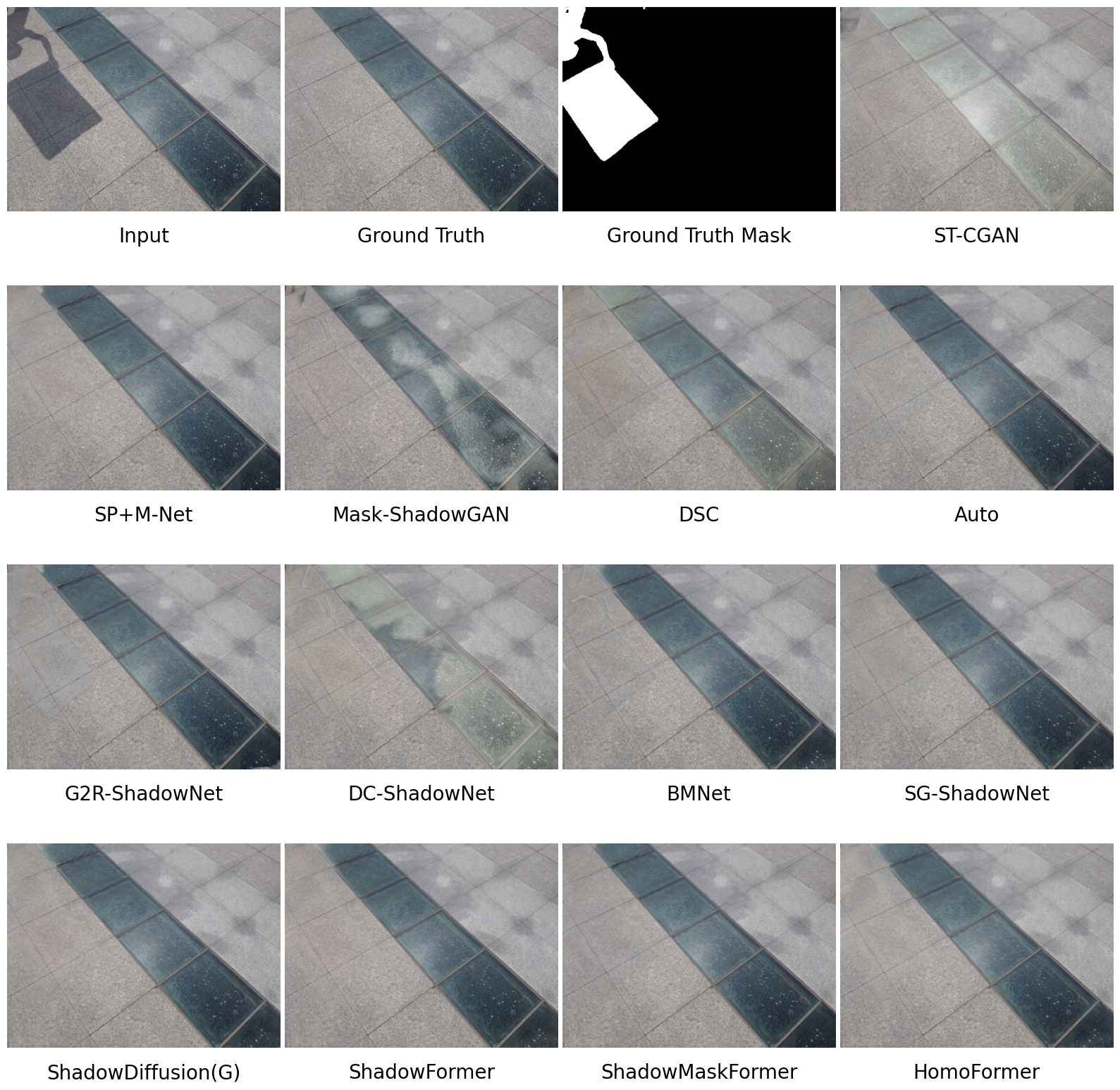}
	\caption{Visual comparison result \#6 on the ISTD+ dataset.}
	\label{fig:supp_sr6}
\end{figure*}

\begin{figure*}[hbp]
	\centering
	\includegraphics[width=1\linewidth]{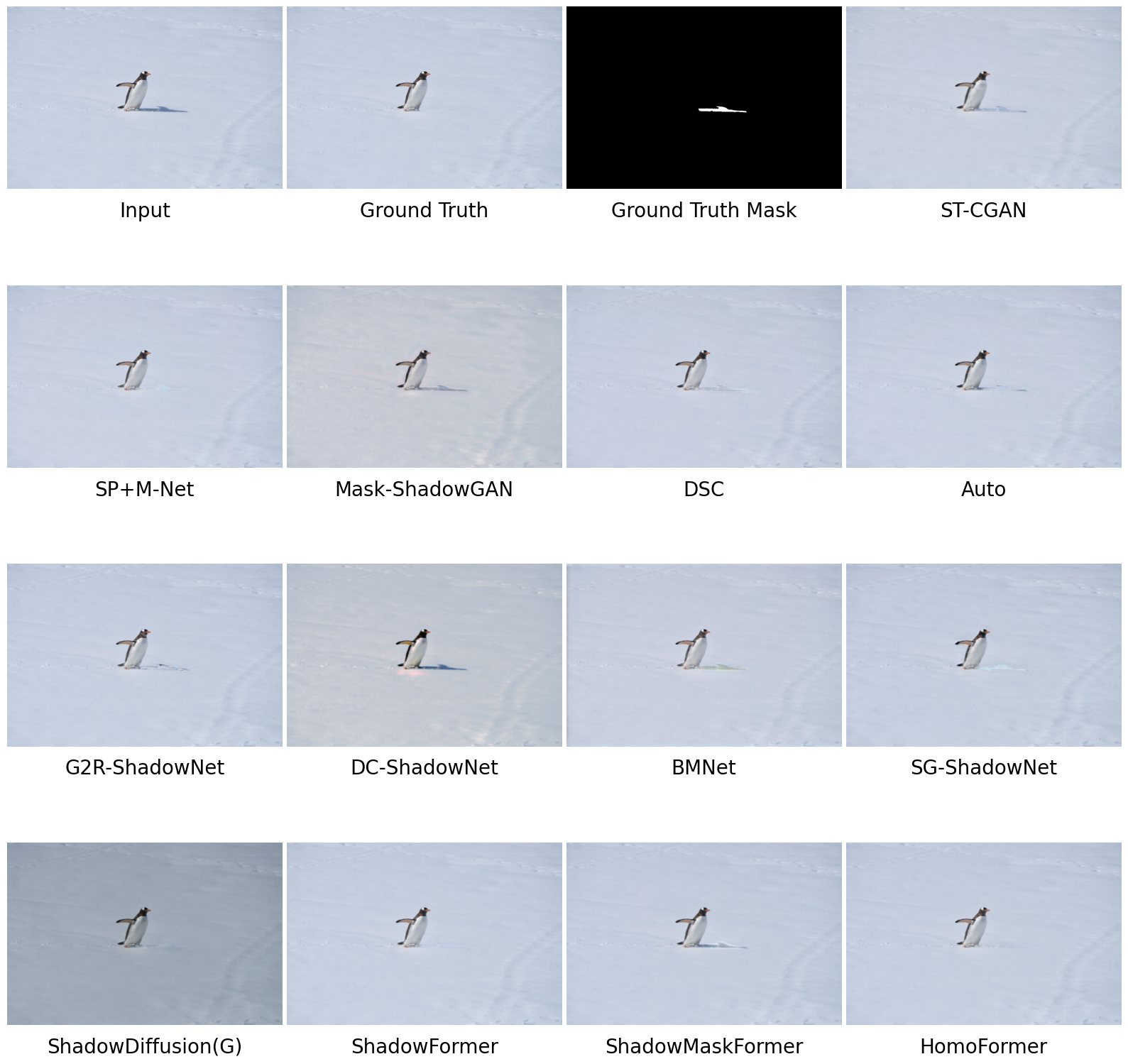}
	\caption{Visual comparison result \#7 on the DESOBA dataset (cross-dataset generalization evaluation).}
	\label{fig:supp_sr7}
\end{figure*}

\begin{figure*}[hbp]
	\centering
	\includegraphics[width=1\linewidth]{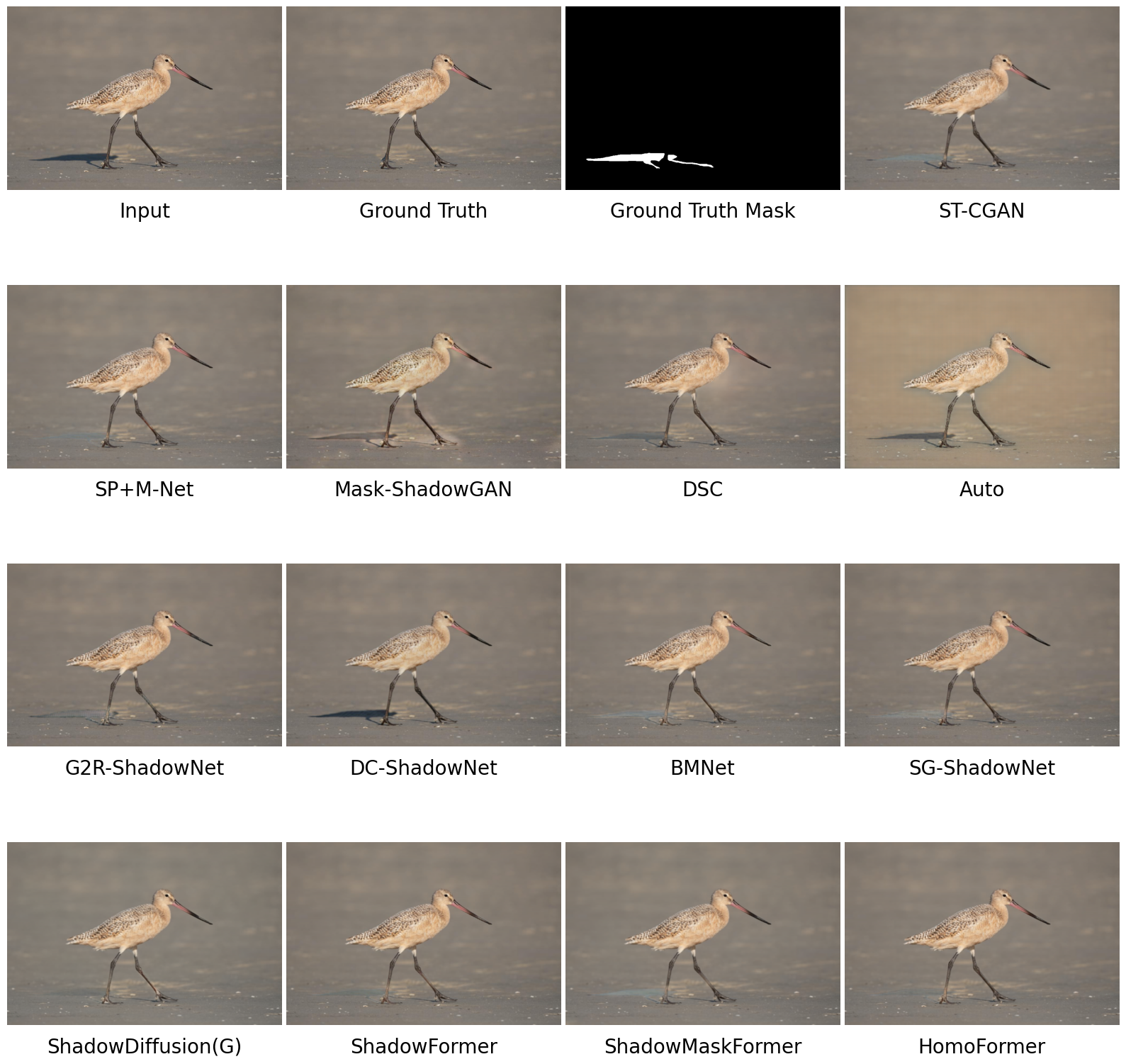}
	\caption{Visual comparison result \#8 on the DESOBA dataset (cross-dataset generalization evaluation).}
	\label{fig:supp_sr8}
\end{figure*}

\begin{figure*}[hbp]
	\centering
	\includegraphics[width=1\linewidth]{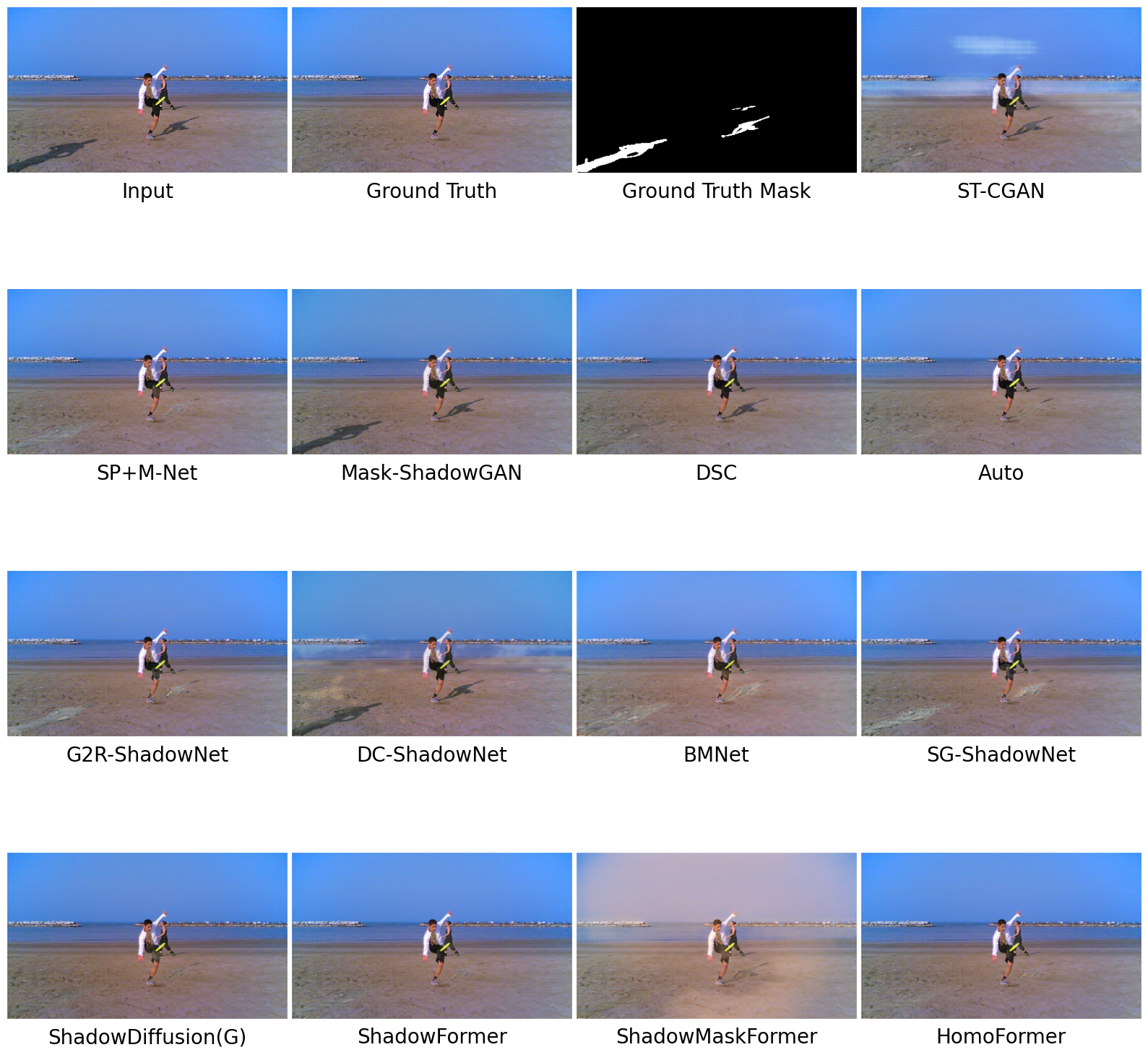}
	\caption{Visual comparison result \#9 on the DESOBA dataset (cross-dataset generalization evaluation).}
	\label{fig:supp_sr9}
\end{figure*}

\begin{figure*}[hbp]
	\centering
	\includegraphics[width=1\linewidth]{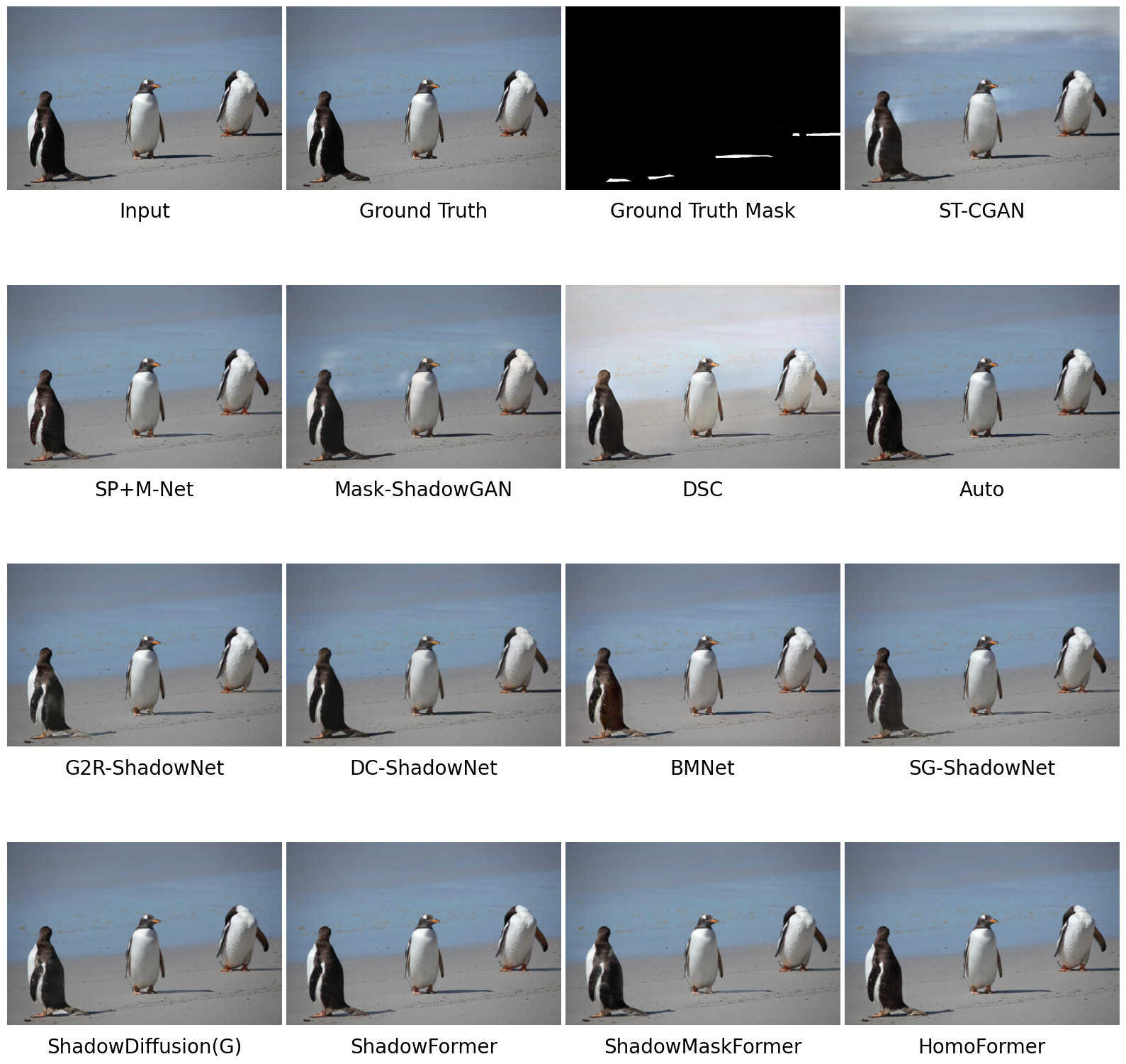}
	\caption{Visual comparison result \#10 on the DESOBA dataset (cross-dataset generalization evaluation).}
	\label{fig:supp_sr10}
\end{figure*}

\clearpage
\onecolumn
\vspace*{2mm}
\large
\noindent
\section*{Part 5: Visual Comparisons on Image Shadow Removal}\label{part5}
\vspace{5mm}

\begin{figure*}[hbp]
	\centering
	\includegraphics[width=.85\linewidth]{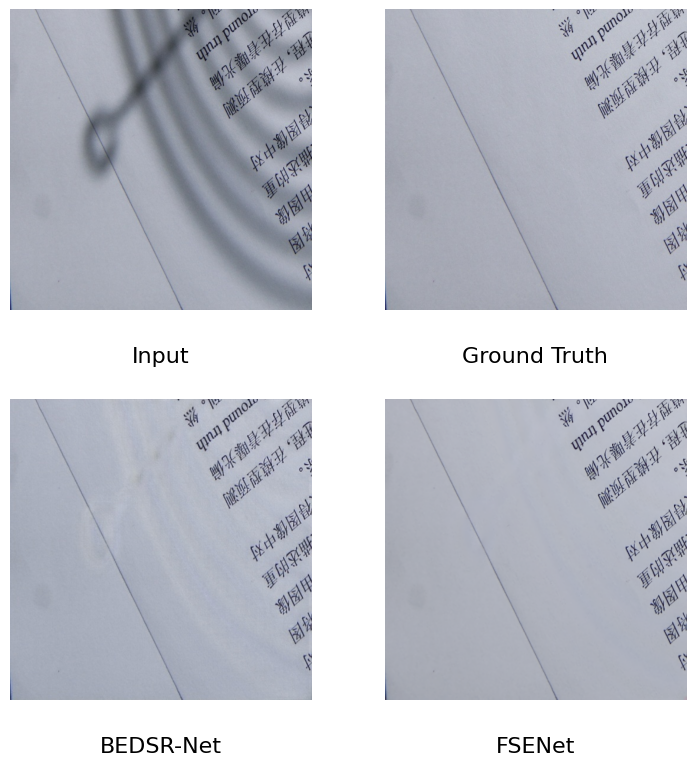}
	\caption{Visual comparison result \#1 on the RDD dataset.}
	\label{fig:supp_sr_doc1}
\end{figure*}

\begin{figure*}[hbp]
	\centering
	\includegraphics[width=.85\linewidth]{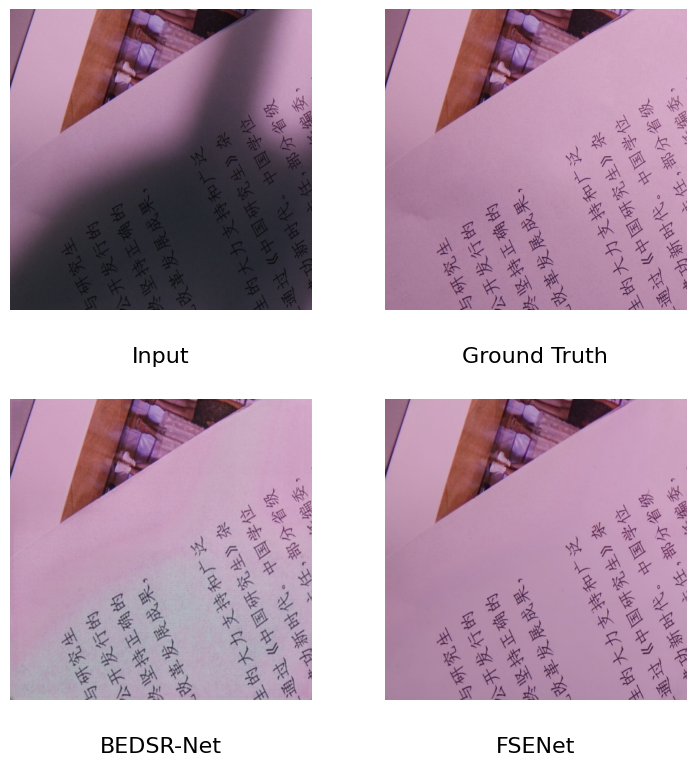}
	\caption{Visual comparison result \#2 on the RDD dataset.}
	\label{fig:supp_sr_doc2}
\end{figure*}

\begin{figure*}[hbp]
	\centering
	\includegraphics[width=.85\linewidth]{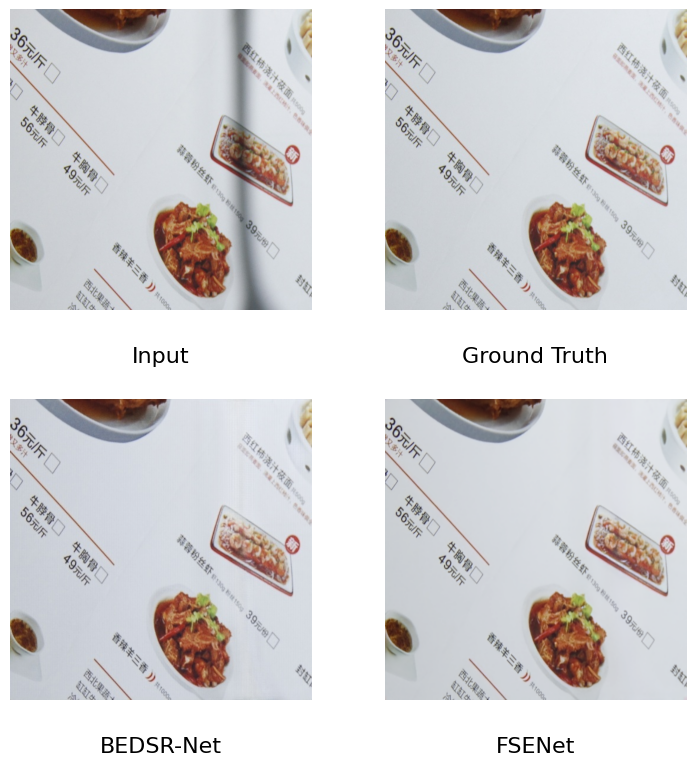}
	\caption{Visual comparison result \#3 on the RDD dataset.}
	\label{fig:supp_sr_doc3}
\end{figure*}

\begin{figure*}[hbp]
	\centering
	\includegraphics[width=.85\linewidth]{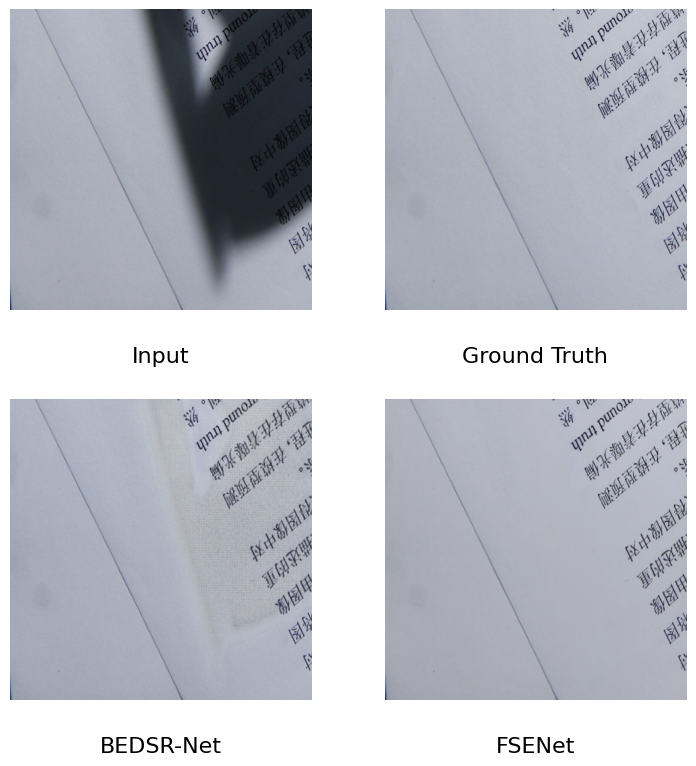}
	\caption{Visual comparison result \#4 on the RDD dataset.}
	\label{fig:supp_sr_doc4}
\end{figure*}

\end{document}